\documentclass{article}

\PassOptionsToPackage{numbers,compress}{natbib}
\usepackage[main,final]{neurips_2026}

\usepackage[utf8]{inputenc}
\usepackage[T1]{fontenc}
\usepackage{microtype,tabularx}
\usepackage{xspace}

\usepackage{amsmath,amssymb,mathtools,amsthm}

\usepackage{xcolor}
\usepackage{pifont}

\usepackage{graphicx}
\usepackage{subcaption}
\usepackage{wrapfig}
\usepackage{float}
\usepackage{tikz}
\usetikzlibrary{shapes,arrows,positioning}

\usepackage{array}
\usepackage{booktabs}
\usepackage{makecell}
\usepackage{multirow}
\usepackage{threeparttable}

\usepackage{enumitem}
\usepackage[linesnumbered,ruled,vlined]{algorithm2e}

\usepackage[textsize=tiny]{todonotes}

\usepackage[
  colorlinks=true,
  linkcolor=purple,
  citecolor=blue
]{hyperref}
\usepackage[acronym]{glossaries}
\usepackage[capitalize,noabbrev]{cleveref}

\makeglossaries

\theoremstyle{plain}
\newtheorem{theorem}{Theorem}[section]
\newtheorem{proposition}[theorem]{Proposition}

\theoremstyle{definition}

\theoremstyle{remark}

\newcommand{\xhdr}[1]{\noindent\textbf{#1.}}

\newcommand{\base}[2]{#1 {\scriptsize (#2)}}

\newcommand{\cmark}{\ding{51}}
\newcommand{\xmark}{\ding{55}}

\newcommand{\ie}{\textit{i.e.,}~}

\newcommand{\method}{\textsc{TopoGED}\xspace}
\definecolor{revised}{rgb}{0,0,1}

\newcommand{\tozero}{\ensuremath{\rightarrow 0}\xspace}
\newcommand{\higherbetter}{\ensuremath{\uparrow}\xspace}
\newcommand{\lowerbetter}{\ensuremath{\downarrow}\xspace}

\newcommand{\C}{\mathcal{C}}

\newcommand{\E}{\mathcal{E}}
\newcommand{\F}{\mathcal{F}}
\newcommand{\G}{\mathcal{G}}

\newcommand{\mS}{\mathcal{S}}

\newcommand{\V}{\mathcal{V}}

\newcommand{\e}{\epsilon}

\title{A Topological Encoder Decoder Framework for Temporal Graph Learning}

\author{%
  Ronan Buck \\
  University of Central Florida \\
  Orlando, FL, USA \\
  \And
  Kiarash Shamsi \\
  University of Manitoba \\
  Winnipeg, Canada \\
  \And
  Tran Gia Bao Ngo \\
  University of Manitoba \\
  Winnipeg, Canada \\
  \AND
  Astrit Tola \\
  Florida State University \\
  Tallahassee, FL, USA \\
  \And
  Baris Coskunuzer \\
  The University of Texas at Dallas \\
  Richardson, TX, USA \\
  \And
  Cuneyt Gurcan Akcora \\
  University of Central Florida \\
  Orlando, FL, USA
}

\begin{document}

\maketitle

\begin{abstract}
Most temporal graph learning methods reduce future prediction to discriminative inference over past interactions, focusing on edges or node labels for a largely persistent set of nodes. This view breaks down when new nodes and edges appear, since the future graph becomes a new object with changing size, composition, and structure. We address this gap by framing temporal graph prediction as an \textit{inverse topology} problem. Instead of predicting edges directly, we first predict a multiscale topological descriptor of the future graph and then reconstruct a plausible future snapshot that realizes this descriptor under inductive constraints. This approach makes global structure and node churn explicit prediction targets and produces forecasted graphs that downstream tasks can evaluate without retraining. Across 14 temporal graph datasets, we evaluate our method, \method, on node, link, and graph property prediction and compare it against state-of-the-art temporal graph models. \method significantly improves node-level forecasting accuracy over the strongest baseline, with a macro average of $0.49$ versus $0.05$. It also outperforms baselines in $60\%$ of graph-structure metric evaluations and substantially increases macro-average edge prediction metrics, from near-zero to $0.12$. Our results show that topology-guided graph forecasting can predict inductive future snapshots whose structure supports multiple downstream evaluations.
\end{abstract}

\section{Introduction}
\label{sec:intro}

Dynamic graphs, in which nodes and edges evolve over time, are central to modeling real systems such as social networks, citation graphs, online communities, and blockchain transaction networks~\cite{DBLP:journals/fcsc/ZhengYW25,DBLP:journals/tkde/FengWWSWH26}. In these settings, the object of interest is often not a single future edge or label, but the next network state itself. A risk signal for an ERC20 token\cite{DBLP:journals/fbloc/ZhuKA24}, a stress indicator for a transaction ecosystem, or a moderation statistic for an online forum depends on the structure of the future graph: which nodes remain active, which new nodes arrive, and how connectivity reorganizes.

Most temporal graph learning methods approach this problem through discriminative prediction. They learn temporal representations from past interactions and use them to score candidate edges or node labels~\citep{DBLP:conf/kdd/YouDL22,DBLP:conf/iclr/XuRKKA20,DBLP:journals/corr/abs-2006-10637}. This paradigm is effective when the future graph is assumed to be an extension of the past over a largely persistent node set. However, this assumption breaks down under node churn. In many real temporal graphs, new nodes appear, old nodes disappear and reappear, and edges form between both existing and newly arriving nodes. In this regime, future prediction is no longer only a ranking problem over known node pairs; it becomes a graph reconstruction problem with changing size, composition, and topology.

We argue that this setting is more naturally addressed through \emph{inductive snapshot forecasting}. Given a sequence of graph snapshots up to time $t$, the goal is to predict the node set and structure for the next snapshot $\widehat{\G}_{t+1}$. This forecast-then-predict view aligns with deployment: downstream decisions are made before the next graph is observed, and many downstream quantities can be computed only after constructing a future graph. Thus, a temporal graph model should not only rank candidate edges, but also produce a coherent future graph on which new tasks can be evaluated.

Topology~\cite{DBLP:journals/frai/ChazalM21} is useful in this setting because it compactly describes graph structure across scales while abstracting away exact node identities. Two snapshots may differ in size and node labels, yet still share similar structural organization, such as heavy-tailed degree patterns, hub-centered motifs, or changes in edge density. Rather than predicting edges directly, we summarize each snapshot by a multiscale filtration descriptor that records how node and edge mass accumulate across filtration thresholds~\cite{DBLP:journals/frai/ChazalM21}. This descriptor captures the graph's structural state, including its final node and edge budgets, while remaining lightweight enough to forecast over time.

This leads to our central formulation: temporal graph forecasting as an \emph{inverse topology problem}. Instead of predicting individual edges, we first predict the future topological descriptor $\widehat{\Phi}(\G_{t+1})$. We then reconstruct a plausible future graph whose structure realizes this predicted descriptor while respecting inductive constraints such as node arrivals, node reappearances, and edge types involving new nodes. Because the descriptor is not injective, many non-isomorphic graphs may share the same topological signature. Our decoder resolves this ambiguity by combining global topological constraints with learned edge-formation scores, seeking a reconstruction that has low descriptor error and matches empirical interaction patterns.

We introduce \textbf{\method} (\textbf{Topo}logical \textbf{G}raph \textbf{E}ncoder \textbf{D}ecoder), a fully inductive encoder--decoder framework for discrete-time temporal graph forecasting. \method encodes each snapshot through a degree-based filtration descriptor and forecasts the next descriptor together with node and edge budgets. A memory module estimates which historical vertices reappear; new unlabeled vertices are instantiated according to the predicted arrival budget, and a multi-phase decoder generates edges by inductive type: old--old recurring, old--old newly formed, old--new, and new--new. The resulting forecasts are complete future snapshots and can be used directly for downstream prediction without task-specific retraining. Our contributions are as follows:

\begin{itemize}
\item We formulate temporal graph learning under node churn as an inductive snapshot forecasting problem and define arrival-aware node and edge types, enabling explicit modeling of reappearing old nodes, newly arriving nodes, and edges involving new nodes.
\item We introduce a topology-guided inverse formulation that forecasts a multiscale filtration descriptor and reconstructs a graph realization from the descriptor.
\item We propose \method, a topological encoder--decoder that combines descriptor prediction, memory-based node reappearance, and multi-phase edge decoding.
\item We evaluate \method across social, web and transaction networks, showing consistent gains in structural fidelity, node-level forecasting, edge-level prediction, and downstream graph-property prediction while reducing the time costs by 95\% against the best competitor.
\end{itemize}

\section{Related Work}
\label{sec:background}

Temporal graph models learn evolving node and edge representations, and are commonly modeled as discrete-time dynamic graphs (DTDGs) or continuous-time dynamic graphs (CTDGs).  CTDGs methods~\cite{DBLP:conf/iclr/YuLZF25,kazemi2022dynamic,DBLP:conf/iclr/TianQG24,DBLP:conf/icml/LuoYHJQJZS23, DBLP:conf/kdd/YouDL22,DBLP:conf/kdd/KumarZL19,DBLP:conf/iclr/TrivediFBZ19,DBLP:conf/iclr/XuRKKA20,DBLP:journals/corr/abs-2006-10637} model the network as a stream of time-stamped events. DTDGs~\cite{DBLP:conf/iclr/ShamsiPHNCA24,DBLP:conf/wsdm/SankarWGZY20, DBLP:conf/iclr/SuZ024,DBLP:conf/kdd/0001ZKHK21} represent the network as a sequence of snapshot graphs collected at discrete timestamps.
These approaches provide strong predictors, but most assume a fixed node universe or persistent node identities, limiting their ability to inductively generate unseen nodes and their interactions (Table~\ref{tab:temporal_graph_comparison}). We focus on DTDGs because snapshot-level modeling offers a tractable framework for forecasting, enabling batch processing over time and explicit control of structural targets (e.g., node and edge budgets under churn). In contrast, CTDG generation relies on event-level modeling and point-process assumptions that are unnecessary for snapshot forecasting.

\textbf{Dynamic graph generation.} Dynamic graph generation has been studied in both discrete-time~\citep{DBLP:conf/sdm/0002ZP021,DBLP:conf/icde/LiWCCZL25} and continuous-time~\citep{DBLP:conf/icml/GravinaLGBG24,DBLP:conf/kdd/ZhouZ0H20,DBLP:conf/aaai/GuptaMBR22,DBLP:conf/aaai/HosseiniSVH25} regimes.
While effective for edge-event synthesis, they do not directly yield coherent {snapshot} forecasts in the presence of churn (node disappearance) and are often tied to training node IDs. This leaves a gap for discrete-time snapshot-forecasting methods that can inductively predict future graphs with new nodes while preserving global structural fidelity.
\begin{wraptable}{l}{0.6\textwidth}

\caption{Temporal graph forecasting capabilities. $o$ and $n$ denote previously observed and newly arriving nodes; the $n$ column denotes the forecast of new nodes (i.e., their count). Unseen $(o$--$o)$ denotes previously unseen edges between observed nodes. Full snapshot denotes joint forecasting of the future node and edge sets.}
\label{tab:temporal_graph_comparison}
\centering
\resizebox{\linewidth}{!}{%
\begin{tabular}{lccccc}
\toprule
Method(s)
& n
& \makecell{unseen\\$(o$--$o)$}
& \makecell{$(o$--$n),$ \\$(n$--$n)$}
& \makecell{Full\\snapshot}
& Objective \\
\midrule
JODIE~\citep{DBLP:conf/kdd/KumarZL19}
& \xmark & \cmark & ? & \xmark & IS \\

DyRep~\citep{DBLP:conf/iclr/TrivediFBZ19}
& \xmark & \cmark & \cmark & \xmark & LT \\

TGAT, TGN, DyGPrompt~\citep{DBLP:conf/iclr/XuRKKA20,
DBLP:journals/corr/abs-2006-10637,DBLP:conf/iclr/YuLZF25}
& \xmark & \cmark & \cmark & \xmark & LN \\

DySAT~\citep{DBLP:conf/wsdm/SankarWGZY20}
& \xmark & \cmark & \cmark & \xmark & LP \\

EvolveGCN~\citep{DBLP:conf/aaai/ParejaDCMSKKSL20}
& \xmark & \cmark & \cmark & \xmark & MP \\

HTGN~\citep{DBLP:conf/kdd/0001ZKHK21}
& \xmark & \cmark & \cmark & \xmark & LP \\

CTAN~\citep{DBLP:conf/icml/GravinaLGBG24}
& \xmark & \cmark & ? & \xmark & LN \\

HOPE~\citep{DBLP:conf/icml/LuoYHJQJZS23}
& \xmark & \xmark & \xmark & \xmark & NT \\

SLATE~\citep{DBLP:conf/nips/KarmimLFT24}
& \xmark & \cmark & \xmark & \xmark & LP \\

TIGGER-I, DG-Gen~\citep{DBLP:conf/aaai/GuptaMBR22,
DBLP:conf/aaai/HosseiniSVH25}
& \xmark & \cmark & \cmark & \xmark & GG \\

GraphPulse, MiNT~\citep{DBLP:conf/iclr/ShamsiPHNCA24,
DBLP:conf/nips/ShamsiNSHPARCRA25}
& \xmark & \xmark & \xmark & \xmark & GP \\
\midrule
\method\ (ours)
& \cmark & \cmark & \cmark & \cmark & SF \\
\bottomrule
\end{tabular}}

\end{wraptable} 

\textbf{Topological and spectral methods.}
Topological and spectral methods incorporate structural priors into graph learning to capture patterns beyond local message passing, such as cycles and global connectivity changes. SLATE~\citep{DBLP:conf/nips/KarmimLFT24} uses temporal Laplacian features but does not address snapshot generation or inductive churn. Persistent homology supports graph-level readouts through differentiable vectorizations and layers~\citep{DBLP:conf/icml/HoferGRNK20,DBLP:conf/aistats/CarriereCILRU20}, while topology-aware GNNs use rewiring and topological augmentations~\citep{DBLP:conf/nips/ChenCG21,DBLP:conf/iclr/HornBMMRB22}. Related methods use topology to study structural evolution and regularize graph generation~\citep{DBLP:journals/kbs/YeJJL23,DBLP:conf/nips/ParkLSWK25}. Our method builds directly on TopER~\citep{DBLP:conf/nips/TolaTAC25}, which summarizes a graph using the slope and intercept of a linear fit between node and edge counts across filtration thresholds. We omit the line-fitting step but extend its filtration-based approach to temporal graph learning through compact multiscale signatures that support snapshot forecasting and generation.

\section{Preliminaries}

\begin{figure}  
    \centering
    \includegraphics[width=\linewidth]{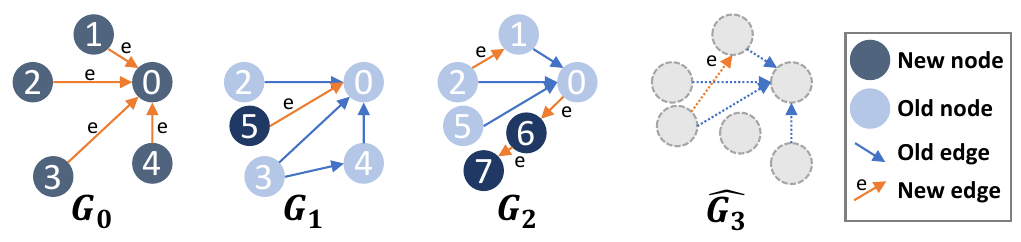}
   \caption{Temporal graph evolution and next-snapshot forecasting. Old nodes and edges have appeared in at least one earlier snapshot; new nodes and edges appear for the first time in the current snapshot. Topological encodings obtained through sublevel filtrations of the observed snapshots $G_0$, $G_1$, and $G_2$ are used to forecast $\widehat{G}_3$.}
    \label{fig:newandold} 
    \vspace{-1.5em}
\end{figure}

\xhdr{Dynamic graph concepts}\label{sec:prelim} We study discrete-time dynamic graphs represented as a sequence of snapshots
$\mathcal{G} = \{\mathcal{G}_1, \ldots, \mathcal{G}_T\}$ with $\mathcal{G}_t = (\mathcal{V}_t, \mathcal{E}_t)$. We consider the inductive setting where the node set changes over time. For each node $v$, we define its first appearance time as
$\tau(v) \coloneqq \min\{t : v \in \mathcal{V}_t\}$. At time $t$, we partition the node set as $\mathcal{V}_t = \mathcal{V}^{\mathrm{new}}_t \cup \mathcal{V}^{\mathrm{old}}_t$, where
$\mathcal{V}^{\mathrm{new}}_t = \{ v \in \mathcal{V}_t : \tau(v) = t \}$ and
$\mathcal{V}^{\mathrm{old}}_t = \{ v \in \mathcal{V}_t : \tau(v) < t \}$.
Nodes that disappear and later reappear are treated as \emph{old} upon reappearance (see Figure~\ref{fig:newandold}).
We decompose edges in $\mathcal{E}_t$ into four disjoint types based on the appearance of edge endpoints:
old--old (oo), old--new (on), new--new (nn). Among old--old edges, we distinguish between previously observed pairs and newly formed ones. Following EdgeBank~\citep{DBLP:conf/nips/PoursafaeiHPR22}, we refer to these as \emph{oo-bank} (\ie previously seen) and \emph{oo-nobank} (\ie not seen) edges. Formally, we define:

 {\footnotesize
\setlength{\abovedisplayskip}{3pt}
\setlength{\belowdisplayskip}{3pt}
$
\begin{aligned}
\mathcal{E}_t^{\mathrm{oo-bank}}
&= \{(u,v)\in\mathcal{E}_t : u,v\in\mathcal{V}_{\mathrm{old}}(t),
(u,v)\in\cup_{j=1}^{t-1}\mathcal{E}_j\}, \quad
\mathcal{E}_t^{\mathrm{on}}
&= \{(u,v)\in\mathcal{E}_t : u\in\mathcal{V}_{\mathrm{old}}(t),
v\in\mathcal{V}_{\mathrm{new}}(t)\},\\
\mathcal{E}_t^{\mathrm{oo-nobank}}
&= \{(u,v)\in\mathcal{E}_t : u,v\in\mathcal{V}_{\mathrm{old}}(t),
(u,v)\notin\cup_{j=1}^{t-1}\mathcal{E}_j\},\quad
\mathcal{E}_t^{\mathrm{nn}}
&= \{(u,v)\in\mathcal{E}_t : u,v\in\mathcal{V}_{\mathrm{new}}(t)\}.~~~~~~~~~~~~~~~~~\\
\end{aligned}
$
}

\xhdr{Topological filtration}\label{sec:filtration_background} Topological data analysis~\cite{dey2022computational} introduces the notion of a filtration to describe how the structure of a graph emerges across scales. By tracking the birth and merging of components and higher-order features as a scale parameter varies, filtrations provide a principled multiscale view of graph organization~\cite{DBLP:conf/nips/ParkLSWK25,DBLP:conf/nips/TolaTAC25}. However, computing full persistent homology on large or evolving graphs is often prohibitively expensive~\cite{DBLP:conf/nips/AkcoraKGC22}, both computationally and in memory, which limits its direct use in temporal graph learning. This motivates the use of lightweight filtration-based summaries that retain multiscale structural information while remaining scalable.

Given $\G$ and a filtration function, such as node degree, $f:\V\to\mathbb{R}$, an induced sublevel filtration using thresholds can be formed as $\e_1<\cdots<\e_n$: $\V_i=\{v\in \V: f(v)\le \e_i\}, \E_i=\{(u,v)\in \E: u,v\in \V_i\}, \G_{\e_1}\subseteq\cdots\subseteq \G_{\e_n}=\G.$
When the filtration level is clear, we omit the subscript $i$.

\section{\method: Topological Graph Encoder-Decoder Framework}
\label{sec:method}

\textbf{Philosophy.} We propose the view that the evolution of a graph can be understood through its multiscale structural shape, and that topology provides a principled way to make this shape comparable across time. In temporal graphs, snapshots may differ widely in size, node identity, and local connectivity due to churn, yet they often follow coherent structural patterns as they evolve. For example, in blockchain transaction networks~\cite{DBLP:conf/nips/ShamsiVKGA22}, new nodes must transact with exchange nodes to buy coins, which creates star-shaped subgraphs around exchange nodes.  
\begin{wrapfigure}{r}{0.28\linewidth}
    \centering
    \vspace{-0.2in}
    \includegraphics[width=\linewidth]{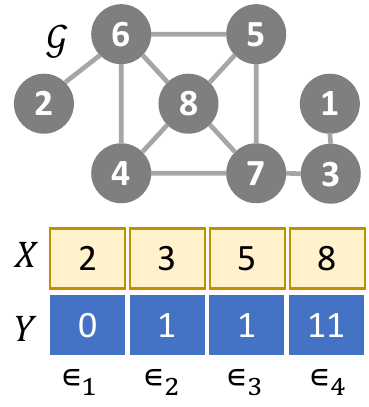}
    \caption{Filtering nodes/edges with $1\text{ to }4$ degree filtration. $X$ and $Y$ track the number of nodes/edges in a given filtration bucket.}
    \label{fig:encoded_graph}
    \vspace{-2.5em}
\end{wrapfigure}
Topological representations capture these patterns by summarizing how connectivity accumulates across scales, abstracting away from specific node labels and exact graph size. In this sense, topology can provide a structural blueprint across snapshots: it preserves the graph's organization while allowing scale- and node-specific effects to be learned downstream.

\begin{figure*}[t]
\vspace{-7px}
    \centering
    \includegraphics[width=.85\linewidth]{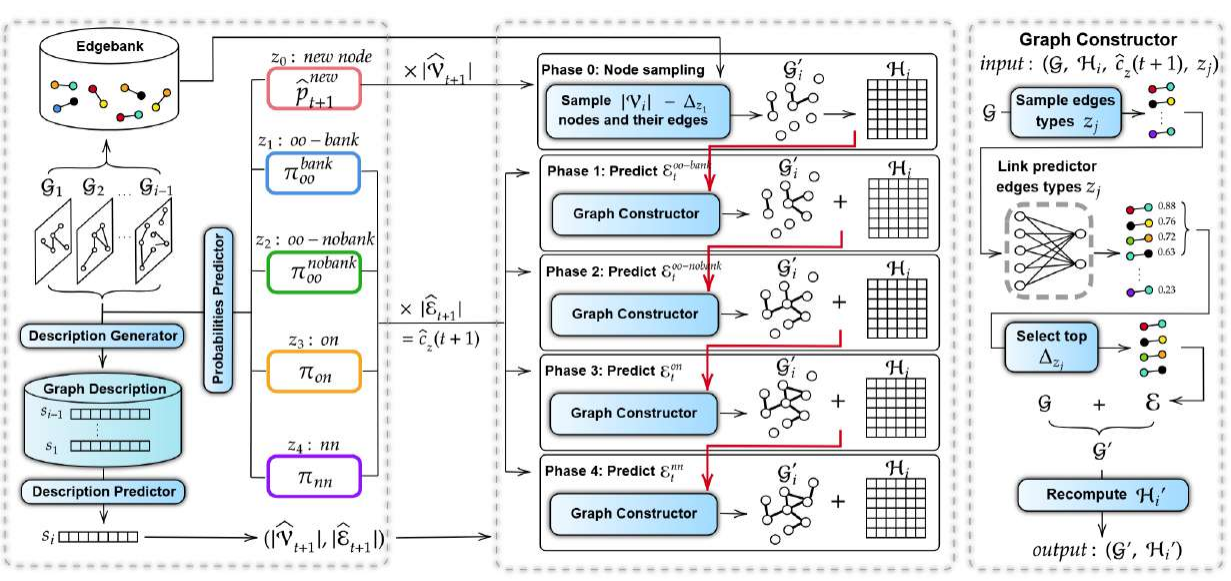}
    \caption{ \textbf{\method overview.}
Each observed snapshot $\G_t$ is encoded into a multiscale topological descriptor $\Phi(\G_t)$ via filtration. A lightweight topology predictor forecasts the next-step descriptor $\Phi(\G_{t+1})$ together with coarse structural targets (node and edge budgets). Conditioned on these targets and a node-memory state, the phased decoder reconstructs the next snapshot $\widehat{\G}_{t+1}$.}
    \label{fig:topoged}
    \vspace{-.2in}
\end{figure*}

\textbf{Goal and high-level design.}
We consider discrete-time dynamic graphs with node churn. 
 Given snapshots up to time $t$, \method forecasts the next snapshot
$\widehat{\G}_{t+1}=(\widehat{\V}_{t+1},\widehat{\E}_{t+1})$ in an inductive setting where new nodes may appear. 
\method consists of three components: (i) a scalable multiscale \emph{topological descriptor} $\Phi(\G_t)$ for each snapshot, computed via TDA-style filtration sequences;
(ii) a \emph{topology predictor} that forecasts the next-snapshot descriptor $\Phi(\G_{t+1})$ together with coarse structural targets, including the edge budget $|\E_{t+1}|$, node budget $|\V_{t+1}|$, and the probability of new nodes $p^{\text{new}}_{t+1}$; and
(iii) a \emph{decoder} that reconstructs $\widehat{\G}_{t+1}$ by generating edges in four stages aligned with the four edge types defined in Section~\ref{sec:prelim}. Figure~\ref{fig:topoged} provides an overview of our design.

\subsection{Graph Encoder}
\label{sec:method_encoder}

For a snapshot $\G_t=(\V_t,\E_t)$, the encoder computes a filtration-based multiscale topological descriptor. Given the degree thresholds $\epsilon_1<\cdots<\epsilon_n$ (by default $n=10$), we form the sublevel filtration $\G_{t,\epsilon_1}\subseteq\cdots\subseteq \G_{t,\epsilon_n}=\G_t$. At each filtration level $i$, we record node and induced-edge counts, $x_{t,i}=|\V_{t,\epsilon_i}|$ and $y_{t,i}=|\E_{t,\epsilon_i}|$.
This yields monotone sequences $X_t=(x_{t,1},\ldots,x_{t,n})$ and $Y_t=(y_{t,1},\ldots,y_{t,n})$. We denote the resulting topological descriptor of $\G_t$ by $\Phi(\G_t)\coloneqq (X_t,Y_t).$ By definition,  at the final threshold $\epsilon_n$, these sequences recover the snapshot sizes: $X_{t,n}=|\V_t|$ and $Y_{t,n}=|\E_t|$ (Fig.~\ref{fig:encoded_graph}). See Appendix Figure~\ref{fig:example_graphs_desxcriptors} for example graphs and their descriptors.

We use the degree-based node filtration $f(v)=\deg(v)$, which is computationally inexpensive. Importantly, a degree provides a direct and efficient handle on the \emph{inverse topology} problem: predicting a graph, given its topological description. Degree thresholds in the descriptor control node inclusion and induced edge counts, which we can use to construct a graph that realizes (\ie decodes) the predicted topological descriptor.  For example, a star graph with four nodes has the descriptor \(((3,0),(3,0),(4,3))\) for the degree thresholds $\epsilon_1,\epsilon_2,\epsilon_3$.

The following proposition supports using $\Phi(\G_{t+1})$ as a stable forecasting target by showing that, for a fixed node set, bounded edge noise induces bounded descriptor error; the churn model handles node-set changes separately. We defer the proof to Appendix~\ref{sec:proof}.

\begin{proposition}[Stability under edge edits]
\label{prop:stability_edge_edits_main}
Let $\G=(\V,\E)$ and $\G'=(\V,\E')$ be graphs on the same node set with 
$|\E\triangle \E'|\le \delta$. Let $X,Y$ and $X',Y'$ be the 
degree-sublevel filtration node and induced-edge count sequences computed 
using thresholds $\epsilon_1<\cdots<\epsilon_n$.
\begin{enumerate}[label=(\roman*)]
    \item If the thresholds are consecutive degrees $\{0,1,\ldots,D\}$, 
    then $\|X-X'\|_1 \le 2\delta$.
    \item For any thresholds, 
    $\|Y-Y'\|_1 \le \delta\sum_{i=1}^n (1+2\epsilon_i)$.
\end{enumerate}
\end{proposition}

\subsection{Predictor for an Encoded Graph}
\label{sec:method_predictor}

Given the most recent $k$ descriptors
$\Phi(\G_{t-k+1}),\ldots,\Phi(\G_t)$, we use a lightweight temporal predictor to forecast the next-step descriptor
$\widehat{\Phi}(\G_{t+1})$. Since the final entries of the descriptor recover snapshot size, this prediction also determines the node and edge budgets
$|\widehat{\V}_{t+1}|$ and $|\widehat{\E}_{t+1}|$. The snapshot node count is learned from $|\hat{\V}_{t+1}|$, but decoding the next snapshot additionally requires anticipating node churn. We therefore jointly predict the fraction of newly appearing nodes at $t+1$, defined as $\hat{p}^{\text{new}}_{t+1} \coloneqq \frac{|\V^{\mathrm{new}}_{t+1}|}{|\V_{t+1}|}$. 
Given the predicted node budget $|\widehat{\V}_{t+1}|$ and the predicted fraction of new nodes $\widehat{p}^{\mathrm{new}}_{t+1}$, we determine the node counts
$\widehat{n}^{\mathrm{old}}_{t+1}=\big\lfloor |\widehat{\V}_{t+1}|(1-\widehat{p}^{\mathrm{new}}_{t+1})\big\rfloor$ and
$\widehat{n}^{\mathrm{new}}_{t+1}=|\widehat{\V}_{t+1}|-\widehat{n}^{\mathrm{old}}_{t+1}$.
This separation isolates the reappearance problem from the arrival problem.

\subsection{Memory Module for Node Reappearance}
\label{sec:memorymodule}

To identify $\widehat{n}^{\mathrm{old}}_{t+1}$ reappearing nodes, we restrict candidates to the set of nodes observed in the last $k$ snapshots memory window: $\mathcal{V}_{t-k:t}=\bigcup_{\tau=t-k}^{t} \V(\G_\tau)$. This window limits variance and reflects the empirical fact that reappearance likelihood decays with time. For each node  $v\in\mathcal{V}_{t-k:t}$, we assign:
\begin{equation}
P(v,t)=\textcolor{teal}{\exp\!\left(-\frac{t-\ell_v}{\lambda}\right)}
\textcolor{purple}{\left(\frac{\deg_v}{\deg_{\max}}\right)^\alpha} \textcolor{blue}{f_v^\beta },
\end{equation}
where $\ell_v$ is the last observed time of $v$, $\deg_v$ is its degree aggregated over the window, $\deg_{\max}$ is the maximum degree in the window, and $f_v$ is the number of snapshots in which $v$ appears within the window. The \textcolor{teal}{exponential term} enforces recency, the \textcolor{purple}{normalized degree} captures structural prominence, and the \textcolor{blue}{frequency term} captures persistence across snapshots. The normalization by $\deg_{\max}$ keeps scores on a comparable scale across datasets. We keep these factors separate to avoid conflating instantaneous connectivity with temporal activity. 

We normalize $\{P(v,t)\}_{v\in\mathcal{V}_{t-k:t}}$ to a distribution and sample $\widehat{n}^{\mathrm{old}}_{t+1}$ nodes without replacement. Sampling introduces diversity and prevents concentration on a small set of high-degree nodes, which improves stability under heavy-tailed degree distributions. We then introduce $|\widehat{\V}^{\mathrm{new}}_{t+1}|$  nodes to complete $\widehat{\V}_{t+1}$. This step treats arrivals as a count prediction problem and avoids imposing identities where none exist. 
 
\subsection{Filtration-Guided Edge Budgets}
\label{sec:filtrationmodule}

\method reconstructs the predicted  snapshot by generating edges in four stages aligned with the edge types in Section~\ref{sec:prelim}:
old--old bank ($\E^{\text{oo-bank}}$), old--old nobank ($\E^{\text{oo-nobank}}$), old--new ($\E^{\text{on}}$), and new--new ($\E^{\text{nn}}$).
Decoding is driven by three sources of information: (i) node churn targets produced by the predictor and memory module, (ii) edge-type budgets, and (iii) optional re-encoding between stages to update node representations as structure (through edges) accumulates.

\xhdr{Edge-type budgets}
Let $|\widehat{\E}_{t+1}|$ be the predicted next snapshot edge count. We predict the edge-type proportions from their recent empirical history. For each snapshot, we compute
$\boldsymbol{\pi}_t=(\pi^{\mathrm{oo-bank}}_{t},\pi^{\mathrm{oo-nobank}}_{t},\pi^{\mathrm{on}}_{t},\pi^{\mathrm{nn}}_{t})$,
where $\pi^z_t=|\E^z_t|/|\E_t|$ for each edge type
$z\in\{\mathrm{oo\text{-}bank},\mathrm{oo\text{-}nobank},\mathrm{on},\mathrm{nn}\}$.
Given the last $k$ proportion vectors, we forecast $\widehat{\boldsymbol{\pi}}_{t+1}$ with the same lightweight temporal predictor (e.g., a weighted average model such as V-EWMA\cite{brown1959statistical})  used for the descriptor variables and renormalize it to lie in the simplex.
The predicted total edge budget $|\widehat{\E}_{t+1}|$ is then divided across edge types as
$\hat{c}^{~z}_{t+1}=\lfloor \widehat{\pi}^z_{t+1}|\widehat{\E}_{t+1}|\rfloor$.

\xhdr{Filtration-guided budget guidance}
In addition to the final node and edge budgets $(|\widehat{\V}_{t+1}|,|\widehat{\E}_{t+1}|)$, the filtration sequences $\Phi(\G_t)\coloneqq (X_t,Y_t).$ provide multiscale structural targets.
We use these sequences to define intermediate checkpoints on cumulative node and edge counts after each decoding phase, and apply a soft penalty when the partially generated graph deviates from the predicted filtration profile.
This guidance steers generation toward the forecasted multiscale structure without enforcing hard degree constraints at individual filtration levels. We elaborate on how the topological descriptor $\Phi(\G_t)$ guides graph construction step by step in Appendix~\ref{appendix:inverse_topology}.

\xhdr{Phase 1: old--old bank edges}
We begin by generating recurring edges between old nodes. The candidate set consists of previously seen edges whose endpoints are old nodes,
$\{(u,v)\in \mathcal{E}_{t-i:t} : u,v \in \V^{\text{old}}_{t+1}\}$. A dedicated predictor head scores these pairs, and we add the highest-scoring pairs until we reach $\hat{c}^{~\text{oo-bank}}_{t+1}$.

\xhdr{Phase 2: old--old nobank edges}
Next, we generate previously unseen edges between old nodes.
Candidate pairs $(u,v)\in\V^{\text{old}}_{t+1}\times\V^{\text{old}}_{t+1}$ are scored by a dedicated predictor head, and the highest-scoring pairs are added until reaching $\hat{c}^{~\text{oo-nobank}}_{t+1}$.

\xhdr{Phase 3: old--new edges}
We then connect new nodes to the existing graph.
We initialize new nodes with cold-start embeddings (Section~\ref{sec:multiheaded}), then score and select old--new candidate pairs until we reach $\hat{c}^{~\text{on}}_{t+1}$.

\xhdr{Phase 4: new--new edges}
Finally, we generate edges among new nodes by scoring new--new candidate pairs and adding edges until we meet the budget $\hat{c}^{~\text{nn}}_{t+1}$.
This produces the completed snapshot $\widehat{\G}_{t+1}$.
After generation, we update the node memory using $\widehat{\G}_{t+1}$.

\subsection{Edge Decoder: Multi-Head Predictor}
\label{sec:multiheaded}

Edge generation is performed by a multi-head predictor with a shared
embedding trunk. At each decoding phase, the trunk computes node embeddings
$\{h_v\}_{v\in \widehat{\V}_{t+1}}$ on the current partially constructed
graph using a GCN~\cite{DBLP:conf/iclr/KipfW17} over the generated adjacency. Four
type-specific heads then score candidate pairs corresponding to the inductive
edge categories:
\[
\begin{aligned}
M_{\mathrm{oo\text{-}bank}} &: (h_u,h_v)\mapsto p^{(\mathrm{oo\text{-}bank})}_{uv},
&\quad
M_{\mathrm{oo\text{-}nobank}} &: (h_u,h_v)\mapsto p^{(\mathrm{oo\text{-}nobank})}_{uv},\\
M_{\mathrm{on}} &: (h_u,h_v)\mapsto p^{(\mathrm{on})}_{uv},
&\quad
M_{\mathrm{nn}} &: (h_u,h_v)\mapsto p^{(\mathrm{nn})}_{uv}.
\end{aligned}
\]

For edge type $z$, the corresponding head outputs probabilities
$\hat p^z_{uv}\in[0,1]$ for candidate pairs $(u,v)$. Candidate pairs are
ranked by $\hat p^z_{uv}$, and the top-ranked edges are added until the
phase-specific budget $\widehat c^z_{t+1}$ is reached. To keep decoding
tractable, each phase scores only a sampled candidate pool, whose size is
chosen proportional to the predicted edge budget.

\textbf{Cold start for new nodes.}
New nodes introduced in phases 3 and 4 initially have no incident edges and
therefore lack structural context. We initialize their embeddings using
degree-conditioned prototypes computed from the predicted old nodes. Let
$\kappa(v)$ denote the degree bin assigned to a new node $v$ from the
predicted filtration and edge-budget targets. 

For bin $k$, define
\quad $\mathcal{C}_k = \{u \in \widehat{\V}^{\mathrm{old}}_{t+1} : \deg(u)=k\},$ and $h_v^{\mathrm{new}} =
\dfrac{1}{|\mathcal{C}_{\kappa(v)}|}
\sum_{u\in\mathcal{C}_{\kappa(v)}} h_u.$
When $\mathcal{C}_{\kappa(v)}$ is empty, we use the nearest non-empty bin,
and fall back to a global prototype if no such bin exists. The binning scheme
is detailed in Appendix~\ref{sec:app_binning}.

Since the filtration descriptor is not injective, many non-isomorphic graphs
can realize the same predicted descriptor. The multi-phase decoder resolves
this ambiguity by combining global filtration-guided targets with learned
edge scores trained from historical positive and negative edges. This yields
graph realizations that are encouraged to match the predicted multiscale
structure while remaining consistent with empirical edge-formation patterns.

To the best of our knowledge, only the degree filtration lets us construct a graph from a given descriptor, but degree is discrete. \textbf{From the broader field of representation learning, we know that a smooth, non-discrete representation may yield better results. However, our long discussions could not find a smooth encoder; hence, we leave this as an open problem.}

\subsection{Training and Inference}
\label{sec:train_infer}

\method is trained by supervising both the evolution of the topological descriptor and the edge generation, using two corresponding components. First, the topology predictor forecasts the next-step descriptor and probabilities. Given the past $k$ encodings, it outputs $\widehat{\Phi}(\G_{t+1})$ and the predicted new-node fraction $\widehat{p}^{\text{new}}_{t+1}$, and is optimized with a regression loss against the true descriptor sequence.

Second, the edge decoder is trained with positive and negative edges from historical snapshots. For each inductive edge type
$z \in \{\text{oo-bank}, \text{oo-nobank}, \text{on}, \text{nn}\}$,
let $\E_{t+1}^{z}$ be the corresponding ground-truth edges in snapshot $t+1$. Each decoder head is trained using binary cross-entropy with negative sampling:
\(
\mathcal{L}_{\text{edge}}^{z}
= - \sum_{(u,v)\in \E_{t+1}^{z}} \log p_{uv}^{z}
  - \sum_{(u',v')\sim \mathcal{N}^{z}} \log\!\big(1-p_{u'v'}^{z}\big),
\)
where $\mathcal{N}^{z}$ denotes uniformly sampled negative pairs from the corresponding candidate space. This encourages decoded graphs to match empirical edge-formation patterns. During training, we use \emph{teacher forcing}~\cite{DBLP:journals/neco/WilliamsZ89}: each decoding phase conditions on the ground-truth partial graph formed by earlier-phase edges in $\G_{t+1}$, stabilizing optimization and decoupling phase-wise errors. The total decoder loss is $\sum_z \mathcal{L}_{\text{edge}}^{z}$, and the topology predictor and decoder are trained with separate optimizers.

\xhdr{Inference}
At inference time, given snapshots up to time $t$ and the current memory state, we predict $\widehat{\Phi}(\G_{t+1})$, the node and edge budgets $(|\widehat{\V}_{t+1}|,|\widehat{\E}_{t+1}|)$, and the new-node fraction $\widehat{p}^{\text{new}}_{t+1}$. We then select reappearing nodes with the memory module, instantiate new nodes, and reconstruct $\widehat{\G}_{t+1}$ by rolling out the four decoding phases autoregressively.   

\xhdr{Computational costs}
The dominant per-snapshot computational cost of \method is
$O\!\left(M|\widehat{\V}_{t+1}|s(d)+L|\widehat{\E}_{t+1}|d\right)$, where $M$ is the number of sampled candidate partners per node, $s(d)$ is the cost of scoring one edge at embedding dimension $d$, and $L$ is the number of GCN layers; thus edge scoring and message passing dominate runtime. We detail the cost breakdown in App.~\ref{sec:costs}.

\vspace{-1em}
\section{Experiments}
\label{sec:exp}

\begin{wraptable}{r}{0.6\linewidth}
\vspace{-2em}
        \captionof{table}{Summary of dataset statistics. Avg \% metrics represent independent averages across all snapshots; totals may not sum to 100\% due to outlier variance and temporal skew. } 
    \label{tab:datasets}
    \resizebox{\linewidth}{!}{%

\begin{tabular}{l | c c c c c c c c}
\toprule
 
\textbf{Dataset} & 
\shortstack{\textbf{Num} \\ \textbf{Graphs}} & 
\shortstack{\textbf{Avg} \\ $\mathbf{|V|}$} & 
\shortstack{\textbf{Avg} \\ $\mathbf{|E|}$} & 
\shortstack{\textbf{Avg \%} \\ $\V^{\mathrm{old}}$} & 
\shortstack{\textbf{Avg \%} \\ oo-bank} & 
\shortstack{\textbf{Avg \%} \\ oo-nobank} & 
\shortstack{\textbf{Avg \%} \\ on} & 
\shortstack{\textbf{Avg \%} \\ nn} \\
\midrule
CollegeMsg   & 181 & 123  & 185  & 91 & 49 & 39 & 10 & 3  \\
MathOverflow & 189 & 142  & 142  & 82 & 28 & 43 & 22 & 6  \\
Adex         & 299 & 249  & 200  & 51 & 28 & 10 & 52 & 9  \\
Aeternity    & 235 & 303  & 282  & 54 & 34 & 12 & 43 & 11 \\
Aion         & 196 & 285  & 239  & 47 & 24 & 13 & 49 & 14 \\
Aragon       & 343 & 336  & 309  & 45 & 18 & 4  & 74 & 4  \\
Bancor       & 317 & 275  & 224  & 47 & 24 & 9  & 54 & 13 \\
Centra       & 264 & 264  & 252  & 49 & 23 & 14 & 55 & 8  \\
Cindicator   & 226 & 300  & 264  & 48 & 24 & 13 & 47 & 16 \\
Coindash     & 274 & 282  & 267  & 58 & 31 & 14 & 52 & 3  \\
DGD          & 725 & 69   & 61   & 61 & 38 & 10 & 47 & 5  \\
Iconomi      & 548 & 179  & 158  & 55 & 34 & 8  & 54 & 5  \\
Reddit\_B    & 405 & 403  & 272  & 87 & 43 & 38 & 17 & 2  \\
TGBL-Wiki    & 31  & 1451 & 1284 & 78 & 53 & 21 & 20 & 6  \\
\bottomrule
 
\end{tabular}
    }    
    \vspace{-1em}
\end{wraptable}
Given a sequence of snapshots $\{\G_t\}_{t=1}^{T}$, we forecast the next snapshot $\widehat{\G}_{t+1}$ using information up to time $t$, and then evaluate downstream performance with a \emph{forecast-then-predict} pipeline.

\textbf{Datasets.}
We evaluate \method on 14 temporal interaction benchmarks: College Message~\cite{DBLP:journals/jasis/PanzarasaOC09}, MathOverflow~\citep{DBLP:conf/wsdm/ParanjapeBL17} and Reddit-Body~\citep{DBLP:conf/www/KumarHLJ18}, TGBL-Wiki~\citep{DBLP:conf/nips/HuangPDFHRLBRR23} and 10 ERC20 token transaction networks extracted from the Ethereum blockchain~\cite{DBLP:conf/iclr/ShamsiPHNCA24}. 
 The ERC20 datasets exhibit substantial node churn, providing a challenging testbed for inductive forecasting. As we show in App.\ref{sec:detailed_results}, edge-type distributions shift markedly over time. Table~\ref{tab:datasets} summarizes dataset statistics.

\xhdr{Temporal predictors} For predicting $\Phi(\widehat{\G}_{t+1})$ and probabilities, we evaluated SMA~\cite{yule1921time}, VAR~\cite{sims1980macroeconomics}, V-EWMA~\cite{brown1959statistical}, SSM~\cite{kalman1960new}, and VECM~\cite{johansen1991estimation}. We adopted V-EWMA because it showed a $0.50\%$ improvement over the nearest competitor in node-count prediction and a $0.88\%$ improvement in edge-count prediction (See App.~\ref{sec:vector_prediction}).  

\xhdr{Edge prediction methods}
We compare \method against prominent and recent temporal GNNs: ROLAND~\cite{DBLP:conf/kdd/YouDL22}, EvolveGCN~\citep{DBLP:conf/aaai/ParejaDCMSKKSL20}, VGRNN~\cite{DBLP:conf/nips/HajiramezanaliH19}, GC-LSTM~\cite{DBLP:journals/apin/ChenWX22}, and HTGN~\citep{DBLP:conf/kdd/0001ZKHK21}. App.~\ref{appendix:baselines} provides further information about these models. We train models for $25$ to $100$ epochs, with early stopping based on validation loss and a patience of $10$ epochs. We use binary cross-entropy loss for the multi-head link prediction task. We optimize using Adam~\citep{DBLP:journals/corr/KingmaB14}. Each model undergoes hyperparameter grid search, and we select the best parameters based on validation AUC over 3 independent runs. We discuss modifications to the temporal GNNs, including parameter tuning, in Appendix~\ref{sec:benchmarker_details}. We partition each dataset into $70\%$ for training, $15\%$ for validation, and $15\%$ for testing.

\xhdr{Computational Environment}
Experiments were conducted on a cluster with NVIDIA H200 NVL GPU (141 GB) nodes, dual 64-core AMD EPYC 9555 processors (128 physical cores), and 1.5 TB of system RAM. 

\xhdr{Reproducibility} Our code is available at \url{https://github.com/FDataLab/TopoGED}.

\xhdr{Hyperparameters}
We train all neural models with Adam, starting with a learning rate of $10^{-3}$ and early stopping on validation performance with patience $10$. We select the edge decoder model using validation ROC-AUC. We tune all models by grid search on the same validation window. \method uses history window $k=5$, batch size $32$, hidden dimension $128$, one GCN layer, dropout $0.1$, and weight decay $0$. We evaluate each task over $3$ independent runs.

\xhdr{Metrics}
We report results in four stages. First, we evaluate descriptor prediction by forecasting the terminal node and edge counts encoded in $\Phi(\G_{t+1})$ and compare five temporal forecasting methods in Appendix~\ref{sec:vector_prediction}. Second, we evaluate constructed graphs at the structure level using twelve metrics, including average degree, triangle count, density, clustering coefficient, and descriptor $\ell_2$ error. Third, we evaluate node-level behavior using seven metrics that measure old-node reappearance and new-node count accuracy. Finally, we evaluate edge-level behavior using eleven metrics that measure edge formation across inductive types, including precision and recall for old--old edges. App. Section~\ref{sec:evaluation_metrics} defines the metrics.

\begin{table}[t]
\centering
\scriptsize
 \resizebox{\linewidth}{!}{
\begin{threeparttable}
\caption{\small \textbf{Forecast results.} For each dataset and metric group, we report \method and the strongest baseline. Structure metric wins is the number of times each model was closest to 0 percent error across assortativity coefficient, clustering coefficient, average degree, density, triangle count, extra edges, extra nodes, missing edges, and missing nodes. Node metrics display the F1 of all nodes. Edge metrics are the F1 scores across all edge types. Detailed results can be seen in Appendix \ref{sec:detailed_results} and~\ref{sec:evaluation_metrics}.}
\label{tab:forecast_fidelity_values}
\begin{tabular}{lcc|cc|ccccc}
\toprule
 
\multirow{2}{*}{Dataset}
& \multicolumn{2}{c}{Structure metric wins $\uparrow$}
& \multicolumn{2}{c}{Node metrics $\uparrow$}
& \multicolumn{2}{c}{Edge metrics $\uparrow$} \\
\cmidrule(lr){2-3}
\cmidrule(lr){4-5}
\cmidrule(lr){6-7}
& \method & Best baseline 
& \method & Best baseline 
& \method & Best baseline  \\
\midrule
CollegeMsg   &  8 & \base{4}{EvolveGCN}      & $0.44\pm0.10$ & \base{$0.04\pm0.02$}{EvolveGCN} & $0.19\pm0.07$ & \base{$0.00\pm0.00$}{multiple}   \\
MathOverflow & 9 & \base{2}{multiple}   & $0.41\pm0.04$ & \base{$0.01\pm0.01$}{multiple}  & $0.02\pm0.01$ & \base{$0.00\pm0.00$}{multiple} \\
Adex         & 6 & \base{4}{HTGN}   & $0.54\pm0.09$ & \base{$0.09\pm0.02$}{GCLSTM}  & $0.09\pm0.04$ & \base{$0.01\pm0.00$}{GCLSTM}\\
Aeternity    & 9 & \base{2}{multiple}   & $0.46\pm0.11$ & \base{$0.01\pm0.01$}{ROLAND}  & $0.11\pm0.04$ & \base{$0.00\pm0.00$}{multiple} \\
Aion         & 6 & \base{3}{multiple}  & $0.45\pm0.05$ & \base{$0.01\pm0.00$}{GCLSTM}  & $0.15\pm0.05$ & \base{$0.00\pm0.00$}{multiple}  \\
Aragon       & 6 & \base{4}{HTGN}  & $0.51\pm0.04$ & \base{$0.01\pm0.00$}{GCLSTM}    & $0.22\pm0.08$ & \base{$0.00\pm0.00$}{multiple}  \\
Bancor       & 6 & \base{4}{HTGN}  & $0.57\pm0.07$ & \base{$0.03\pm0.01$}{GCLSTM}  & $0.14\pm0.07$ & \base{$0.00\pm0.00$}{multiple} \\
Centra       & 5 & \base{4}{ROLAND}  & $0.47\pm0.14$ & \base{$0.01\pm0.01$}{multiple}   & $0.07\pm0.06$ & \base{$0.00\pm0.00$}{multiple}  \\
Cindicator   & 6 & \base{4}{ROLAND}    & $0.51\pm0.04$ & \base{$0.02\pm0.01$}{GCLSTM}  & $0.13\pm0.07$ & \base{$0.00\pm0.00$}{multiple}   \\
Coindash     & 5 & \base{4}{HTGN}    & $0.47\pm0.12$ & \base{$0.03\pm0.01$}{GCLSTM} & $0.13\pm0.06$ & \base{$0.00\pm0.00$}{multiple} \\
DGD          & 8 & \base{3}{EvolveGCN}  & $0.53\pm0.09$ & \base{$0.02\pm0.01$}{GCLSTM}    & $0.08\pm0.05$ & \base{$0.00\pm0.00$}{multiple}   \\
Iconomi      & 6 & \base{3}{multiple}  & $0.51\pm0.09$ & \base{$0.01\pm0.01$}{ROLAND}    &  $0.10\pm0.06$ & \base{$0.00\pm0.00$}{multiple}  \\
Reddit-B     & 8 & \base{2}{multiple} & $0.38\pm 0.03$ & \base{$0.00\pm0.00$}{multiple}   & $0.03\pm0.02$ & \base{$0.00\pm0.00$}{multiple}   \\
TGBL-Wiki    & 7 & \base{4}{multiple}  & $0.65 \pm 0.01$ & \base{$0.35 \pm 0.01$}{HTGN}   & $0.18\pm0.01$ & \base{$0.01 \pm 0.00$}{HTGN}  \\
\midrule
Macro avg.   & $6.78\pm1.32$ & \base{$3.35\pm0.81$}{HTGN}  & $0.49\pm0.07$ & \base{$0.05\pm0.09$}{GCLSTM}  & $0.12\pm0.06$ & \base{$0.00\pm0.01$}{multiple} \\
\bottomrule
 
\end{tabular}
\end{threeparttable}}
\vspace{-.2in}
\end{table}

\xhdr{Graph prediction} \Cref{tab:forecast_fidelity_values} reports wins and aggregate scores across structure, node, and edge evaluations. \method dominates structure metrics with a macro average of $6.78$ wins compared to $3.35$ for the strongest baseline. This advantage is consistent across datasets, with near-complete coverage on Centra and DGD. These results show that the predicted descriptor translates into accurate global properties in the constructed graphs.

At the node level, \method achieves a macro average score of $0.49$, while the strongest baseline reaches $0.05$. The gap is stable across datasets and is largest on Bancor and DGD. Baseline node performance is limited by the absence of explicit node-set prediction. Standard temporal graph models score edges over a fixed or sampled node universe and rely on negative sampling for AUC evaluation. This setup measures ranking quality within a candidate set but does not constrain the size or composition of $\hat{\V}_{t+1}$. When these models are used for full graph construction, the active node set is induced implicitly by predicted edges and can grow without control. This exploding node-set problem leads to low precision and recall for node reappearance. In contrast, \method predicts node budgets, selects reappearing nodes through the memory module, and introduces new nodes explicitly, which yields consistent improvements.

\begin{wraptable}{l}{0.63\linewidth}
    \vspace{-.2in}
    \caption{Ablation for CollegeMsg. For all metrics, closer to 0 is better.     \textbf{Bold} indicates best, \underline{underlined} is second-best.}
    \label{tab:college_structure_ablation}
    \resizebox{0.98\linewidth}{!}{%
        \begin{tabular}{l | c c c | c}
\toprule
 
\textbf{Metric} & \textbf{F. Probs} & \textbf{F. $\Phi(\widehat{\mathcal{G}}_t)$} & \textbf{\shortstack{Oracle}} & \textbf{\method} \\ \midrule
Avg Node Degree & $3.63 \pm 4.91$ & $\mathbf{0.17 \pm 0.35}$ & $\underline{0.29 \pm 0.09}$ & $0.36 \pm 0.19$ \\
Unique Degree Count & $2.32 \pm 0.82$ & $0.90 \pm 0.55$ & $\mathbf{0.60 \pm 0.32}$ & $\underline{0.64 \pm 0.48}$ \\
Degree Centrality & $\mathbf{0.02 \pm 0.82}$ & $\underline{0.09 \pm 0.35}$ & $0.29 \pm 0.09$ & $0.28 \pm 0.24$ \\
Assortativity Coefficient & $\mathbf{-0.87 \pm 14.05}$ & $2.53 \pm 5.82$ & $1.59 \pm 6.98$ & $\underline{1.13 \pm 7.92}$ \\
Clustering Coefficient & $0.75 \pm 2.82$ & $0.34 \pm 1.25$ & $\mathbf{0.05 \pm 0.05}$ & $\underline{0.07 \pm 0.06}$ \\
Density & $\mathbf{0.02 \pm 0.82}$ & $\underline{0.09 \pm 0.35}$ & $0.29 \pm 0.09$ & $0.28 \pm 0.24$ \\
Num Triangles & $7929.25 \pm 31327.34$ & $7.32 \pm 6.29$ & $\mathbf{1.86 \pm 2.16}$ & $\underline{3.11 \pm 3.20}$ \\
\midrule
Descriptor Norm & $906.68 \pm 1901.83$ & $41.74 \pm 10.11$ & $\mathbf{27.67 \pm 7.88}$ & $\underline{35.42 \pm 11.91}$ \\
Median Extra Nodes & $\underline{123.00 \pm 12.85}$ & $\mathbf{0.00 \pm 1.15}$ & $\mathbf{0.00 \pm 0.00}$ & $\mathbf{0.00 \pm 1.16}$ \\
Median Missing Nodes & $\mathbf{0.00 \pm 0.00}$ & $\underline{0.50 \pm 1.18}$ & $\mathbf{0.00 \pm 0.00}$ & $\underline{0.50 \pm 1.18}$ \\
Median Extra Edges & $413.50 \pm 47.61$ & $\mathbf{6.00 \pm 2.80}$ & $8.00 \pm 1.25$ & $\underline{7.50 \pm 2.67}$ \\
Median Missing Edges & $\mathbf{0.00 \pm 0.00}$ & $\mathbf{0.00 \pm 0.12}$ & $\mathbf{0.00 \pm 0.00}$ & $\mathbf{0.00 \pm 0.00}$ \\
\bottomrule
\end{tabular}

    }
    \vspace{2pt} 
   
    \vspace{-.1in}
\end{wraptable}

At the edge level, \method attains a macro average of $0.12$ compared to $0.01$ for the strongest baseline. The improvement appears across all datasets, with clear gains on Aragon, Coindash, and TGBL-Wiki. Baselines often achieve near-zero scores because they do not model inductive edge types and therefore fail to allocate edges across old--old, old--new, and new--new categories. 

\begin{wrapfigure}{r}{0.38\textwidth}
    \centering
    \includegraphics[width=\linewidth]{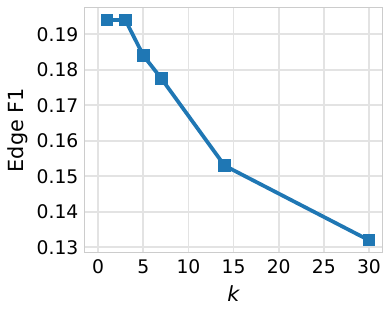}
    \caption{Sensitivity of edge prediction to history window $k$ in TGBL-Wiki.}
    \label{fig:k_sensitivity}

    \medskip

    \includegraphics[width=\linewidth]{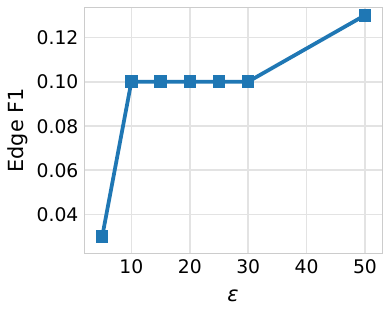}
    \caption{Sensitivity of edge prediction to the number of buckets $\epsilon$ in Adex.}
    \label{fig:epsilon_sensitivity}
    \vspace{-1.5em}
\end{wrapfigure}

Methods such as VGRNN may appear competitive on isolated metrics by overpredicting edges, which increases recall on sampled evaluations but introduces structural noise in full-graph reconstruction. HTGN performs competitively on datasets with dominant old--old interactions such as CollegeMsg, where most edges fall into this category. This masks its limitation to old--old edges. However, it does not model node churn or inductive edge types and degrades under settings with higher turnover.

\method predicts node counts, edge counts, and all inductive edge types explicitly. It aligns node selection, edge allocation, and global structure with the predicted descriptor. This alignment yields consistent gains across structure, node, and edge evaluations and produces forecasted graphs that remain coherent under node churn and structural drift.

\xhdr{Ablations} We run three ablations. In \textit{False $\Phi$}, we replace the descriptor $\widehat{\Phi}(\G_{t+1})$ with a random monotone sequence $(\widetilde{X},\widetilde{Y})$ that preserves the final node and edge counts but discards multiscale structure. In \textit{False Probs}, we retain $\widehat{\Phi}(\G_{t+1})$ but replace node and edge-type probabilities with random values. In \textit{Oracle}, we provide the decoder with the ground-truth descriptor and probabilities from $t+1$, which serves as an upper bound for reconstruction.

Table~\ref{tab:college_structure_ablation} shows that each component controls a distinct failure mode. Randomizing probabilities (\textit{F. Probs}) leads to severe over-generation, with large errors in triangle count and descriptor norm, indicating that correct edge allocation requires calibrated node and edge-type probabilities. In contrast, corrupting the descriptor (\textit{F. $\Phi$}) preserves budgets but distorts higher-order structure, producing large deviations in assort. and clustering coefficient. This confirms that matching final counts alone is insufficient; the multiscale profile in $\Phi$ is necessary to recover structural organization.\begin{wraptable}{r}{0.8\linewidth}
\vspace{-1em}
\centering
\caption{Resource usage across 14 datasets.}
\label{tab:costs}
\resizebox{\linewidth}{!}{
\begin{tabular}{l|rrr}
\toprule
\textbf{Model} &
\textbf{Time $\downarrow$  (sec)}&
\textbf{RAM $\downarrow$ (MB)} &
\textbf{GPU $\downarrow$ (MB)} \\
\midrule
ROLAND & $319.3\pm235.7$ & $4512.7\pm3315.6$ & $1124.5\pm569.4$ \\
TGCN & $28.1\pm19.1$ & $6251.4\pm3829.5$ & $869.8\pm528.4$ \\
GCLSTM & $686.7\pm365.2$ & $6183.0\pm3805.5$ & $1534.1\pm877.1$ \\
VGRNN & $432.3\pm269.5$ & $3135.9\pm1093.7$ & $35359.2\pm22840.8$ \\
HTGN & $391.2\pm377.8$ & $5950.0\pm3578.7$ & $2923.0\pm1646.9$ \\
EvolveGCN & $92.0\pm49.6$ & $6411.0\pm3853.3$ & $1351.7\pm878.8$ \\
{\method (ours)} & $\mathbf{19.0\pm12.2}$ & $\mathbf{2196.4\pm351.5}$ & $\mathbf{839.0\pm1469.6}$ \\
\bottomrule
\end{tabular}}
\vspace{-.1in}
\end{wraptable}
The oracle configuration yields the lowest errors across nearly all metrics, as expected.  More importantly, \method approaches this oracle without access to ground-truth future information. The gap between \method and the oracle is small across structure metrics, while both ablations degrade substantially. This shows that accurate reconstruction requires the joint use of descriptor prediction and probabilistic decoding, and that \method effectively aligns these components to recover both global structure and node-level behavior.

\textbf{Efficiency.} \quad As Table~\ref{tab:costs} summarizes, \method is scalable and uses substantially fewer computational resources than the baselines.

\begin{wrapfigure}{l}{0.35\linewidth}
\vspace{-.2in}
    \centering

    \includegraphics[width=\linewidth]{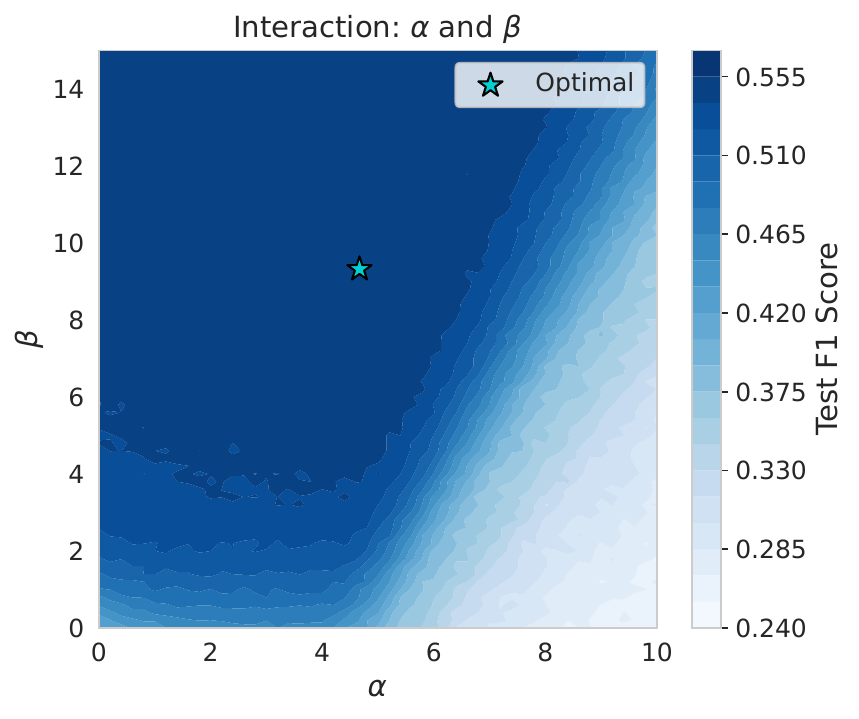}
    \vspace{-0.5em}
    \caption{Sensitivity of node reappearance performance to $\alpha$ and $\beta$ in TGBL-Wiki.}
    \label{fig:alpha_beta_sensitivity}
    \vspace{-1.5em}
\end{wrapfigure}

\textbf{Sensitivity.} \quad
Figures~\ref{fig:alpha_beta_sensitivity} and~\ref{fig:k_sensitivity} show that the memory module is stable over a broad parameter range. The $\alpha$--$\beta$ sweep shows a large high-F1 region for moderate $\alpha$ and high $\beta$, indicating that persistence across recent snapshots is the strongest signal for node reappearance, while degree should receive a nonzero but not dominant weight. 
 Very large $\alpha$ values reduce performance, since they overemphasize high-degree nodes and weaken the contribution of temporal recurrence. 
 
The history-window sweep shows that performance improves sharply from very short memory windows and then declines steadily as $k$ grows. Longer windows enlarge the candidate pool which can introduce stale candidates, supporting the use of a finite recent memory for reappearing-node selection. We provide further analysis about \method node predictor in Appendix~\ref{sec:node_prediction_methods}.

\xhdr{Limitations}  
\method\ forecasts discrete-time snapshots rather than event-level interaction streams. Newly arriving nodes are assigned the next available ID: the model predicts the number and structural roles of new nodes without using their external identities, so it does not predict who the new nodes are before they are observed. These assumptions match the inductive snapshot-forecasting setting studied here, while continuous-time event generation and identity prediction for unseen nodes remain natural directions for future work.

\section{Conclusion}
We introduced \method, a topology-guided encoder--decoder framework for inductive temporal graph forecasting under node churn. Rather than reducing future prediction to edge ranking over a fixed node set, \method forecasts a multiscale filtration descriptor, predicts node and edge budgets, and decodes a complete future snapshot through arrival-aware edge generation. 

This formulation makes node reappearance, new-node arrival, edge-type allocation, and global structural fidelity explicit parts of the prediction problem. Across diverse temporal graph benchmarks, \method produces forecasted graphs that better match future node sets, edge structure, and graph-level statistics than the existing temporal graph models, while remaining computationally efficient. These results suggest that filtration-guided snapshot reconstruction is a viable alternative to purely discriminative temporal graph prediction, and provide a step toward temporal graph models that can forecast coherent future network states rather than only score isolated future interactions.

\subsection*{Acknowledgements}
This work was partially supported by the National Science Foundation under grant DMS-2229417 and by the Simons Foundation under grant SFM-21269. The authors acknowledge the UCF Advanced Research Computing Center and Texas Advanced Computing Center (TACC) at UT Austin.

\bibliography{references}
\bibliographystyle{plainnat}

\newpage
\onecolumn
\appendix
\section*{Appendix}

\section{Broader Impact}\label{sec:broader_impact}
\method can broaden the use of temporal graph learning in settings where decisions must be made before the next network state is observed. By forecasting an entire inductive snapshot, including node arrivals, reappearances, and future edge structure, it supports early risk assessment, anomaly monitoring, resource planning, and downstream analysis in evolving networks such as transaction ecosystems, online communities, and scientific interaction graphs. Its topology-guided design makes forecasts more interpretable than isolated edge scores, since users can inspect predicted node counts, edge budgets, churn, and structural properties before applying downstream models. Because the output is a full forecasted graph, multiple tasks can be evaluated on the same predicted snapshot without task-specific retraining. The efficient descriptor-based pipeline also reduces computational cost, making graph forecasting more accessible for large dynamic networks.

\section{Extended Related Work}
\label{sec:extened_related_work}

\textbf{Temporal graph representation learning.}
Temporal graph models learn time-evolving node and edge representations to predict links, nodes, or graph-level labels.
Early methods update embeddings along interaction streams with memory, e.g., JODIE~\cite{DBLP:conf/kdd/KumarZL19} and DyRep~\citep{DBLP:conf/iclr/TrivediFBZ19}.
More recent temporal GNNs incorporate temporal encoding and attention to support inductive learning over time-stamped edges, e.g., TGAT~\cite{DBLP:conf/iclr/XuRKKA20} and TGN \citep{DBLP:journals/corr/abs-2006-10637}, while snapshot-based models attend over sequences of graphs, e.g., DySAT \citep{DBLP:conf/wsdm/SankarWGZY20}. PRES is an example of a dynamic memory model that optimizes for efficient training between snapshots~\cite{DBLP:conf/iclr/SuZ024}.

\textbf{CTDG Methods.}  CTDGs~\cite{DBLP:conf/iclr/TianQG24,DBLP:conf/icml/LuoYHJQJZS23, DBLP:conf/kdd/YouDL22} model the network as a stream of time-stamped node-level events. Memory-based CTDG methods update embeddings along interaction streams~\cite{DBLP:conf/kdd/KumarZL19,DBLP:conf/iclr/TrivediFBZ19}, while temporal GNNs add attention and temporal encoding for inductive learning~\cite{DBLP:conf/iclr/XuRKKA20,DBLP:journals/corr/abs-2006-10637}. 

\textbf{DTDG Methods.}DTDGs represent the network as a sequence of snapshot graphs collected at discrete timestamps. Snapshot-based models attend over sequences of graphs, e.g., DySAT \citep{DBLP:conf/wsdm/SankarWGZY20}. PRES is an example of a dynamic memory model that optimizes for efficient training between snapshots~\cite{DBLP:conf/iclr/SuZ024}. HTGN~\cite{DBLP:conf/kdd/0001ZKHK21} project each snapshot into hyperbolic space to model node representation. While existing methods focus on node-level tasks, GraphPulse~\cite{DBLP:conf/iclr/ShamsiPHNCA24} and MiNT~\cite{DBLP:conf/nips/ShamsiNSHPARCRA25} are snapshot-based models for graph property prediction tasks.

\textbf{Dynamic graph generation.} Dynamic graph generation has been studied in both discrete-time and continuous-time regimes.
In discrete time, prior methods generate sequences of snapshots by learning recurrent or variational snapshot latents \citep{DBLP:conf/sdm/0002ZP021,DBLP:conf/icde/LiWCCZL25}. In continuous time, event-based generators model timestamped interactions, often via temporal walks or point-process machinery \citep{DBLP:conf/icml/GravinaLGBG24,DBLP:conf/kdd/ZhouZ0H20,DBLP:conf/aaai/GuptaMBR22,DBLP:conf/aaai/HosseiniSVH25}; while effective for edge-event synthesis, they do not directly yield coherent {snapshot} forecasts under churn and are commonly tied to training node IDs. This leaves a gap for discrete-time snapshot forecasting methods that can inductively predict future graphs with new nodes while maintaining global structural fidelity.


\section{Forecast-to-Graph Reconstruction Details}

This section provides the implementation details for converting the quantities
forecast by \method into a concrete future snapshot
$\widehat{\G}_{t+1}=(\widehat{\V}_{t+1},\widehat{\E}_{t+1})$. We first describe
the degree-based filtration descriptor and its inverse reconstruction role in
Appendix~\ref{appendix:inverse_topology}. We then detail how the predicted node
set is assembled, including reappearing-node sampling from a recent memory
window in Appendix~\ref{sec:node_prediction_methods} and degree-bin assignment
for cold-start nodes in Appendix~\ref{sec:app_binning}. Next, we define the
edge-type probabilities used to allocate the predicted edge budget across
old--old-bank, old--old-nobank, old--new, and new--new interactions in
Appendix~\ref{app:edgeProb}. These components are combined in the full
budgeted multi-phase reconstruction algorithm in Algorithm~\ref{alg:topoged_decode}.
Finally, Appendix~\ref{sec:benchmarker_details} explains how link-prediction
baselines are adapted to the same full-graph construction setting through
thresholding and bounded adjacency reconstruction.

\subsection{Degree-Based Filtration Encoder and Inverse Topology Decoder} 
\label{appendix:inverse_topology}

Let $\G_t=(\V_t,\E_t)$ 
denote a graph snapshot at time $t$, where $\V_t$ is the node set and
$\E_t$ is the edge set. We encode each snapshot by a filtration-based
multiscale topological descriptor. The goal of this descriptor is to
summarize the evolution of the graph structure across increasing degree
thresholds, while retaining enough information to later reconstruct a graph
snapshot consistent with the encoded topology.

We use the degree function
\[
f_t(v)=\deg_{\G_t}(v), \qquad v\in \V_t,
\]
as the filtration function. Given degree thresholds
\[
\epsilon_1 < \epsilon_2 < \cdots < \epsilon_n,
\]
we define the active vertex set at filtration level $\epsilon_i$ by
\[
\V_{t,\epsilon_i}
=
\{v\in \V_t : f_t(v)\leq \epsilon_i\}.
\]
The corresponding filtered graph is the induced subgraph
\[
\G_{t,\epsilon_i}
=
\G_t[\V_{t,\epsilon_i}]
=
(\V_{t,\epsilon_i},\E_{t,\epsilon_i}),
\]
where
\[
\E_{t,\epsilon_i}
=
\{(u,v)\in \E_t : u,v\in \V_{t,\epsilon_i}\}.
\]
Thus, the filtration produces a nested sequence of induced subgraphs
\[
\G_{t,\epsilon_1}
\subseteq
\G_{t,\epsilon_2}
\subseteq
\cdots
\subseteq
\G_{t,\epsilon_n}
=
\G_t.
\]
At each filtration level, we record two structural quantities: the number of
active vertices and the number of induced edges. Specifically, we define
\[
x_{t,i}=|\V_{t,\epsilon_i}|,
\qquad
y_{t,i}=|\E_{t,\epsilon_i}|.
\]
This gives two monotone sequences
\[
X_t=(x_{t,1},x_{t,2},\ldots,x_{t,n})
\]
and
\[
Y_t=(y_{t,1},y_{t,2},\ldots,y_{t,n}).
\]
The filtration-based topological descriptor of the snapshot $\G_t$ is then
defined as
\[
\Phi(\G_t)\coloneqq (X_t,Y_t).
\]
By construction, the final filtration level recovers the size of the original
snapshot:
\[
x_{t,n}=|\V_t|,
\qquad
y_{t,n}=|\E_t|.
\]
Therefore, $\Phi(\G_t)$ provides a compact multiscale representation of the
snapshot, recording how the number of vertices and induced edges grows as the
degree threshold increases.

The use of the degree filtration has two advantages. First, degree is
computationally inexpensive and directly available from the graph structure.
Second, and more importantly, degree provides a natural handle on the inverse
topology problem. In the forward direction, the encoder maps a graph snapshot
to its descriptor,
\[
\G_t \longmapsto \Phi(\G_t)=(X_t,Y_t).
\]
The inverse topology problem asks for the reverse operation: given a predicted
descriptor
\[
\widehat{\Phi}_{t+1}
=
(\widehat X_{t+1},\widehat Y_{t+1}),
\]
construct a graph snapshot
\[
\widehat{\G}_{t+1}
=
(\widehat{\V}_{t+1},\widehat{\E}_{t+1})
\]
whose filtration realizes the predicted descriptor. That is, we seek
\[
\Phi(\widehat{\G}_{t+1})
\approx
\widehat{\Phi}_{t+1}.
\]
Equivalently, for each threshold $\epsilon_i$, the induced subgraph of
$\widehat{\G}_{t+1}$ should satisfy
\[
|\widehat{\V}_{t+1,\epsilon_i}|
=
\widehat{x}_{t+1,i},
\qquad
|\widehat{\E}_{t+1,\epsilon_i}|
=
\widehat{y}_{t+1,i}.
\]

This inverse problem should be understood as a graph realization problem rather
than exact graph recovery. The descriptor $(X_t,Y_t)$ does not encode the full
adjacency matrix of $\G_t$; instead, it records the number of active vertices
and induced edges at each degree threshold. Consequently, many non-isomorphic
graphs may share the same descriptor. The decoder therefore aims to construct
one graph that is consistent with the predicted multiscale topology.

\begin{figure}[t]
    \centering
    \includegraphics[width=\textwidth]{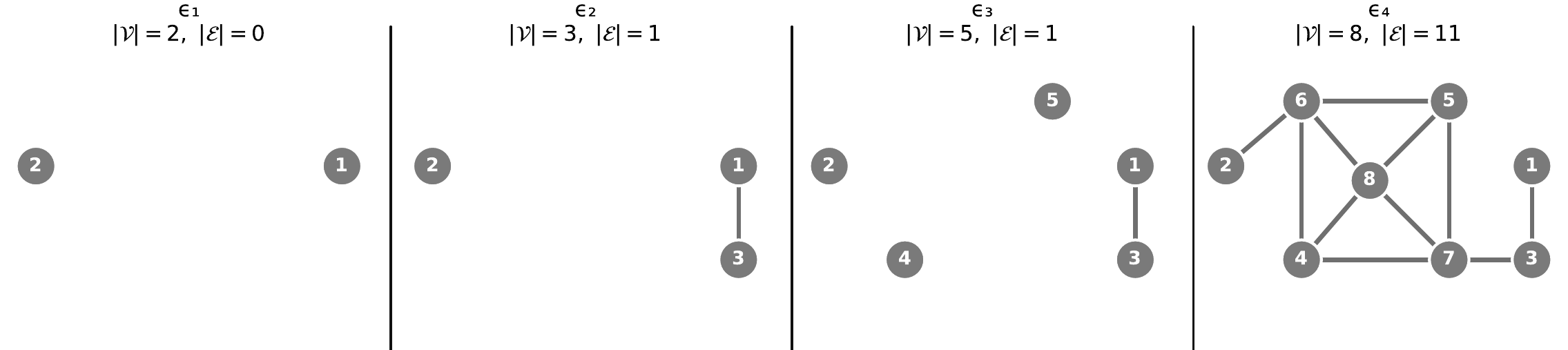}
    \caption{
    Degree-threshold filtration of a graph snapshot. 
    At each filtration level $\epsilon_i$, the induced subgraph records
    the number of active vertices $|\mathcal{V}|$ and induced edges
    $|\mathcal{E}|$.
    }
    \label{fig:filtered_graph}
\end{figure}

A natural decoding procedure proceeds level by level through the predicted
filtration. Suppose the predicted descriptor is
\[
\widehat X=(\widehat x_1,\ldots,\widehat x_n),
\qquad
\widehat Y=(\widehat y_1,\ldots,\widehat y_n).
\]
At the first filtration level, the decoder creates $\widehat x_1$ vertices and
adds $\widehat y_1$ induced edges among them. At the next level, it adds
\[
\widehat x_2-\widehat x_1
\]
new vertices and then adds edges so that the induced edge count becomes
$\widehat y_2$. Continuing in this way, at level $i$ the decoder adds
\[
\widehat x_i-\widehat x_{i-1}
\]
new vertices and introduces
\[
\widehat y_i-\widehat y_{i-1}
\]
additional induced edges, subject to the constraint that all previously
constructed subgraphs remain unchanged. Thus, the predicted descriptor controls
both node inclusion and edge growth across the filtration.

Formally, the decoder constructs nested vertex sets
\[
\widehat{\V}_{\epsilon_1}
\subseteq
\widehat{\V}_{\epsilon_2}
\subseteq
\cdots
\subseteq
\widehat{\V}_{\epsilon_n}
=
\widehat{\V}
\]
with
\[
|\widehat{\V}_{\epsilon_i}|=\widehat x_i,
\]
and edge sets satisfying
\[
|\widehat{\E}_{\epsilon_i}|=\widehat y_i,
\qquad
\widehat{\E}_{\epsilon_i}
=
\{(u,v)\in \widehat{\E}:u,v\in \widehat{\V}_{\epsilon_i}\}.
\]
The final output is the graph
\[
\widehat{\G}=(\widehat{\V},\widehat{\E}),
\]
where
\[
|\widehat{\V}|=\widehat x_n,
\qquad
|\widehat{\E}|=\widehat y_n.
\]

This construction makes the descriptor directly actionable. The sequence
$\widehat X$ determines how many vertices must be introduced at each degree
threshold, while the sequence $\widehat Y$ determines how many induced edges
must be present among those vertices. Therefore, the degree-based descriptor is
not only a representation of the graph; it also serves as a blueprint for
constructing a graph snapshot from predicted topological data.

Figure~\ref{fig:filtered_graph} illustrates this process. For a graph
$\G=(\V,\E)$, the filtration levels $\epsilon_1,\epsilon_2,\epsilon_3,\epsilon_4$
produce induced subgraphs with descriptors
\[
(|\V_{\epsilon_1}|,|\E_{\epsilon_1}|),
\quad
(|\V_{\epsilon_2}|,|\E_{\epsilon_2}|),
\quad
(|\V_{\epsilon_3}|,|\E_{\epsilon_3}|),
\quad
(|\V_{\epsilon_4}|,|\E_{\epsilon_4}|).
\]
In the displayed example, these values are
\[
(2,0),\qquad (3,1),\qquad (5,1),\qquad (8,11).
\]
Thus,
\[
X=(2,3,5,8),
\qquad
Y=(0,1,1,11).
\]
Reading the filtration forward encodes the graph into the descriptor
$\Phi(\G)=(X,Y)$, while reading it in reverse shows how the same descriptor can
be used to decode a graph realization. The large increase from
$|\E_{\epsilon_3}|=1$ to $|\E_{\epsilon_4}|=11$ indicates that most of the edge
structure appears when the final group of vertices enters the filtration. Hence,
the filtration descriptor captures not only the final size of the graph, but
also the scale at which its connectivity emerges.

In summary, the encoder maps each snapshot $\G_t$ to a compact degree-based
topological descriptor $\Phi(\G_t)=(X_t,Y_t)$. The inverse topology decoder then
uses a predicted descriptor to construct a graph snapshot whose thresholded
induced subgraphs match the predicted node and edge counts. This provides a
direct bridge between topological prediction and graph generation: instead of
predicting every edge independently, the model predicts a multiscale structural
summary, and the decoder realizes this summary as a concrete graph.

\begin{figure}
    \centering
    \includegraphics[width=0.98\linewidth]{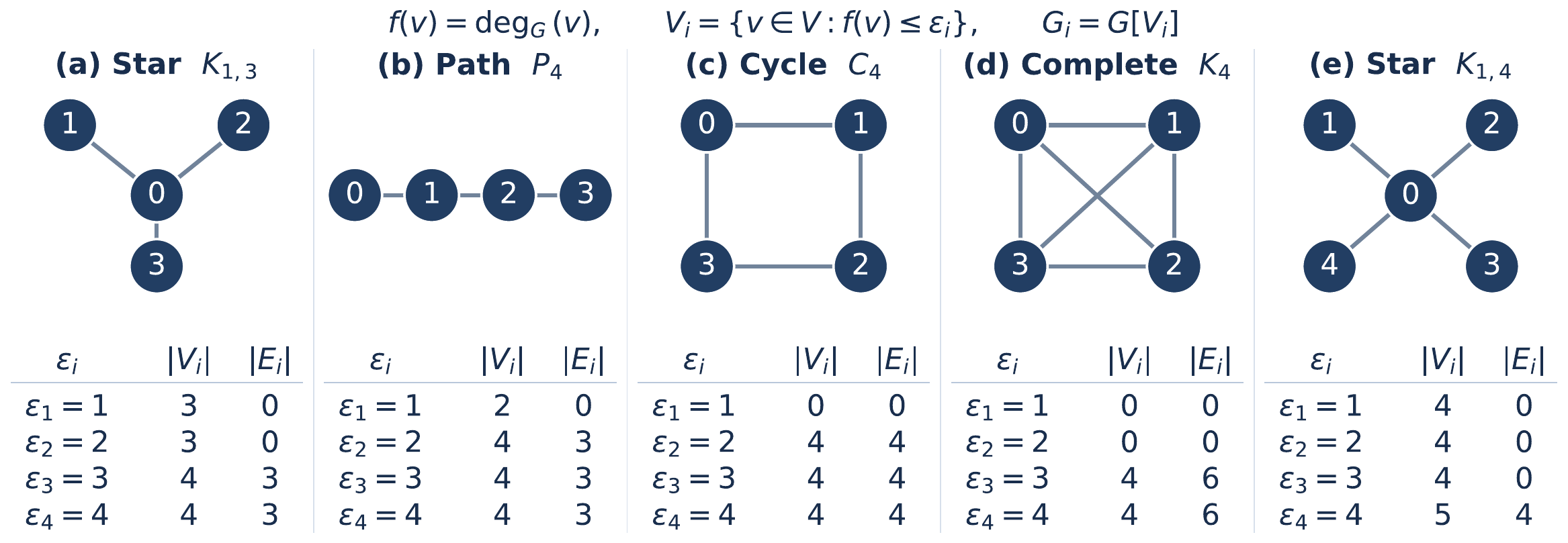}
    \caption{Degree-based structural descriptors for the star graphs $K_{1,3}$ and $K_{1,4}$, path $P_4$, cycle $C_4$, and complete graph $K_4$. Each table reports the vertex count $|V_i|$ and edge count $|E_i|$ of the induced subgraph $G_i=G[\{v\in V:\deg_G(v)\leq\epsilon_i\}]$ at thresholds $\epsilon_i=i$ for $i\in\{1,2,3,4\}$. Node labels denote vertex identifiers, and $d$ lists degrees in the original graph in vertex order.}
    \label{fig:example_graphs_desxcriptors}
\end{figure}

\subsection{Node Set Prediction and Reappearance Sampling}
\label{sec:node_prediction_methods}
A primary contribution of \method is the node prediction heuristic for old nodes. For each timestamp $T$, instead of predicting edges with the entire node universe for the dataset, we assign nodes a 'reappearance probability' for weighted random sampling. 

For each $v\in\mathcal{V}_{t-k:t}$, we assign
\begin{equation}
P(v,t)=\exp\!\left(-\frac{t-\ell_v}{\lambda}\right)
\left(\frac{\deg_v}{\deg_{\max}}\right)^\alpha f_v^\beta ,
\end{equation}
where $\ell_v$ is the last observed time of $v$, $\deg_v$ is its degree aggregated over the window, $\deg_{\max}$ is the maximum degree in the window, and $f_v$ is the number of snapshots in which $v$ appears within the window. The exponential term enforces recency, the normalized degree captures structural prominence, and the frequency term captures persistence across snapshots. We keep these factors separate to avoid conflating instantaneous connectivity with temporal activity. The normalization by $\deg_{\max}$ keeps scores on a comparable scale across datasets.

According to the descriptor and probability vector, $n$ nodes are randomly sampled and added to the current graph to be constructed. 

Weighted sampling prevents the decoder from always selecting the same highest-score nodes, while still favoring nodes with stronger recency, degree, and frequency evidence.

Tables~\ref{tab:opt_params} and~\ref{tab:performance} summarize the fitted parameters and the resulting node reappearance performance. The optimal $\beta$ values are consistently high across datasets, showing that repeated activity within the memory window is the strongest signal for old-node reappearance. The $\alpha$ values vary more substantially, which indicates that degree is useful but dataset-dependent. The fitted $\lambda$ values are generally below $1$, so recent activity receives much larger weight than older activity in most datasets.

The F1 scores show that node reappearance is highly dataset-dependent. TGBL-Wiki and CollegeMsg achieve the strongest test performance, while sparse token graphs such as Aion and Aragon remain difficult. In several datasets, validation and test performance exceed training performance, suggesting that the heuristic is not simply memorizing the training snapshots. Overall, these results support the use of a small memory module: recency, degree, and frequency provide enough signal to select plausible old nodes without scoring the full historical node universe.

\begin{table}[H]
\centering
\begin{minipage}{0.48\textwidth}
\centering
\caption{Optimal Parameters}
\label{tab:opt_params}
\small 
\begin{tabular}{l ccc}
\toprule
\toprule
\textbf{Dataset} & \boldmath$\alpha$ & \boldmath$\beta$ & \boldmath$\lambda$ \\
\midrule
CollegeMsg   & 3.88 & 8.81 & 0.31 \\
Adex         & 2.64 & 8.94 & 0.85 \\
Aeternity    & 3.53 & 9.24 & 0.53 \\
Aion         & 1.94 & 8.75 & 0.60 \\
Aragon       & 3.53 & 9.99 & 0.55 \\
Bancor       & 4.19 & 9.51 & 0.64 \\
Centra       & 3.25 & 9.78 & 0.35 \\
Cindicator   & 3.74 & 9.24 & 0.51 \\
Coindash     & 2.97 & 9.47 & 0.63 \\
Dgd          & 1.32 & 9.96 & 0.31 \\
Iconomi      & 2.94 & 9.70 & 0.37 \\
mathoverflow & 5.94 & 7.46 & 0.15 \\
Reddit\_B    & 3.92 & 9.96 & 0.96 \\
TGBL-Wiki         & 4.67 & 9.32 & 0.21 \\
\bottomrule  
\bottomrule  
\end{tabular}
\end{minipage}
\hfill 
\begin{minipage}{0.48\textwidth}
\centering
\caption{Node reappearance prediction F1 Score}
\label{tab:performance}
\small
\begin{tabular}{l ccc}
\toprule
\toprule
\textbf{Dataset} & \textbf{Train} & \textbf{Val} & \textbf{Test} \\
\midrule
CollegeMsg   & 0.4563 & 0.4366 & 0.3976 \\
Adex         & 0.0915 & 0.1637 & 0.2257 \\
Aeternity    & 0.1276 & 0.2103 & 0.2353 \\
Aion         & 0.0645 & 0.0917 & 0.0944 \\
Aragon       & 0.0719 & 0.0670 & 0.1128 \\
Bancor       & 0.1402 & 0.1710 & 0.2383 \\
Centra       & 0.1230 & 0.2380 & 0.1723 \\
Cindicator   & 0.0655 & 0.1197 & 0.1491 \\
Coindash     & 0.1311 & 0.2483 & 0.2866 \\
Dgd          & 0.2203 & 0.1999 & 0.2428 \\
Iconomi      & 0.1306 & 0.1136 & 0.2212 \\
mathoverflow & 0.3160 & 0.3125 & 0.2993 \\
Reddit\_B    & 0.2511 & 0.2556 & 0.2777 \\
TGBL-Wiki         & 0.5542 & 0.5608 & 0.5525 \\
\bottomrule
\bottomrule
\end{tabular}
\end{minipage}
\end{table}

\subsection{Node Binning and Cold-Start Initialization}
\label{sec:app_binning}
For nodes $v\in \mathcal{V}_{t+1}$ in the node set $\mathcal{V}$ of graph $G_{t+1}$, we assign a maximum degree according to the current filtration sequence. 

Previously existing nodes, $v\in\mathcal{V}_{\text{old}}$ are assigned a maximum degree equivalent to their most recent appearance, rounded up to the threshold $\epsilon_k$ for the nearest non-empty bin.
$$\deg(v) = \min \{ \epsilon_k \in \mathcal{B} \mid \epsilon_k \geq \deg(v)_{\tau} \text{ and } V_k > 0 \}$$

Where $\epsilon_k$ is the threshold for each bin in $\mathcal{B}$, $\deg(v)_\tau$ is the most recently seen degree for node v, and $\mathcal{V}_k$ is the number of nodes in bin $k$.

For $v\in\mathcal{V_{\text{new}}}$, new nodes initially have no information. Therefore, for each node, a maximum degree $\epsilon$ is randomly assigned if the bin is not empty. 
$$\text{deg}(v) = \text{random}(\{ \epsilon_k \in \mathcal{B} \mid V_k > 0 \})$$

\begin{table}[t]
\centering
\caption{Glossary of complexity variables.}
\label{tab:complexity_variables}
\small
\begin{tabular}{ll}
\toprule
\toprule
Symbol & Meaning \\
\midrule
$\mathcal{T}$ & Total training epochs, fixed at $100$ \\
$N_{\mathrm{train}}$ & Number of training snapshots, using the $70\%$ training partition \\
$N_{\mathrm{val}}$ to $N$ & Test set range, using the final $15\%$ partition \\
$L$ & Number of GCN/GAT layers in the encoder \\
$d$ & Node embedding dimensionality \\
$H$ & Predictive heads for edge types $\{\mathrm{o\text{-}o}, \mathrm{o\text{-}n}, \mathrm{n\text{-}n}\}$ \\
$|\mathcal{E}_{i,h}^{+}|$ & Number of positive edge samples per head during training \\
$|\mathcal{E}_{i,h}^{-}|$ & Number of negative edge samples per head during training \\
$E_t, V_t$ & Observed edges and nodes in the input context snapshot \\
$\widehat{\E}, \widehat{\V}$ & Target edges and nodes defined by TopER and probability vectors \\
$\mathcal{B}$ & Number of filtration buckets used for node-degree assignment \\
$M$ & Number of potential edge candidates sampled per node \\
$s(d)$ & Computational cost of the MLP scoring function \\
$\log \widehat{\E}$ & Complexity term for top-$K$ edge selection through sorting \\
\bottomrule
\bottomrule

\end{tabular}
\end{table}

\subsection{Edge-Type Counts and Probabilities}
\label{app:edgeProb}

We study the relative prevalence of inductive edge types in snapshot $\G_t = (\V_t, \E_t)$ using the decomposition defined in Section~\ref{sec:prelim}. Recall that nodes are partitioned into old and new at each timestep $t$ as
$\V_t = \V^{\mathrm{old}}_t \cup \V^{\mathrm{new}}_t$, and edges are grouped into four disjoint types:
old--old-bank (recurring edges between old nodes),
old--old-nobank (new edges between old nodes),
old--new (edges connecting old and new nodes), and
new--new (edges between new nodes).

The total number of edges at time $t$ is:
\[
|\E_t| = |\E^{\text{oo-bank}}_{t}| + |\E^{\text{oo-nobank}}_{t}| + |\E^{\text{on}}_{t}| + |\E^{\text{nn}}_t|.
\]

We define the normalized edge-type probabilities as the share of each type in the snapshot:
\begin{alignat*}{2}
\pi^{\text{oo-bank}}_{t} &= \frac{|\E^{\text{oo-bank}}_{t}|}{|\E_t|}, \quad
\pi^{\text{oo-nobank}}_{t} = \frac{|\E^{\text{oo-nobank}}_{t}|}{|\E_t|}, \\
\pi^{\text{on}}_{t} &= \frac{|\E^{\text{on}}_{t}|}{|\E_t|}, \quad
\pi^{\text{nn}}_{t} = \frac{|\E^{\text{nn}}_t|}{|\E_t|}.
\end{alignat*}

These probabilities quantify the inductive edge-type composition of the current snapshot. For example, $p^{\text{bank}}_{\text{oo}}(t)$ measures the recurrence rate of previously observed edges, while $p_{\text{nn}}(t)$ reflects purely novel interactions among newly arriving nodes.

\begin{algorithm}[H]
\caption{\method Inference: Budgeted Multi-Phase Snapshot Reconstruction}
\label{alg:topoged_decode}
\KwIn{Observed snapshots $\{\G_{t-k+1},\ldots,\G_{t}\}$; Edge Set $\mathcal{B}$; encoder $\Phi(\cdot)$; topology predictor $\mathsf{Pred}$; trunk encoder $\mathsf{Enc}$; type heads $\{M_{z}\}_{z \in \mathcal{Z}}$.}
\KwOut{Predicted snapshot $\widehat{\G}_{t+1}=(\widehat{\V}_{t+1},\widehat{\E}_{t+1})$ and updated EdgeBank $\mathcal{B}$.}

\vspace{0.25em}
\textbf{1. Global Budget Forecasting.} \\
Compute $\Phi(\G_\tau)$ for $\tau=t-k+1,\ldots,t$. \\
Predict $\widehat{\Phi}(\G_{t+1}),\ |\widehat{\V}_{t+1}|,\ \widehat{\E}_{t+1},\ \widehat{p}^{\mathrm{new}}_{t+1}$ and edge-type proportions $\boldsymbol{\pi}_{t+1}$ using $\mathsf{Pred}$. \\
Set $\widehat{\V}^{\mathrm{old}}_{t+1}=\lfloor |\widehat{\V}_{t+1}|(1-\widehat{p}^{\mathrm{new}}_{t+1})\rfloor$ and $\widehat{\V}^{\mathrm{new}}_{t+1}=|\widehat{\V}_{t+1}|-\widehat{\V}^{\mathrm{old}}_{t+1}$. \\
Set per-type edge budgets $\widehat{c}_z(t+1)=\lfloor \pi_z(t+1)\widehat{\E}_{t+1}\rfloor$ for $z\in\{\mathrm{oo\text{-}bank},\mathrm{oo\text{-}nobank},\mathrm{on},\mathrm{nn}\}$. \\

\vspace{0.25em}
\textbf{2. Inductive Node Set Assembly.} \\
\textit{Reappearing Nodes:} Sample $\widehat{N}_{\mathrm{old}}$ nodes from the historical universe $\bigcup_{\tau=t-k+1}^t \V_\tau$ using weighted random sampling as in Section~\ref{sec:memorymodule}. Initialize $v \in \widehat{\V}^{\mathrm{old}}_{t+1}$ with their most recent observed degree. \\
\textit{Cold-Start Nodes:} Initialize $\widehat{N}_{\mathrm{new}}$ node identifiers. Assign each $v \in \widehat{\V}^{\mathrm{new}}_{t+1}$ an initial degree value from the remaining capacity in the predicted descriptor $\widehat{\Phi}(\G_{t+1})$. \\
Define $\widehat{\V}_{t+1} = \widehat{\V}^{\mathrm{old}}_{t+1} \cup \widehat{\V}^{\mathrm{new}}_{t+1}$.

\vspace{0.25em}
\textbf{3. Representation Learning.} \\
Initialize embeddings $h$: for $v \in \widehat{\V}^{\mathrm{old}}_{t+1}$, use the most recently seen embedding from history; for $v \in \widehat{\V}^{\mathrm{new}}_{t+1}$, initialize with default vectors. \\
Compute inductive representations $h \leftarrow \mathsf{Enc}(\widehat{\V}_{t+1}, \G_{t-k+1:t})$.

\vspace{0.25em}
\textbf{4. Multi-Phase Edge Reconstruction (Top-K Filtering).} \\
\textit{Phase I (Bank Recurrence):} Define $\mathcal{C}_{\mathrm{oo\text{-}bank}} = \{(u, v) \in \mathcal{B} \mid u, v \in \widehat{\V}^{\mathrm{old}}_{t+1}\}$. Add top $\widehat{c}_{\mathrm{oo\text{-}bank}}$ pairs scored by $M_{\mathrm{oo\text{-}bank}}(h_u, h_v)$ to $\widehat{\E}_{t+1}$. \\
\textit{Phase II (Discovery):} Define $\mathcal{C}_{\mathrm{oo\text{-}nobank}} = \{(u, v) \in \widehat{\V}^{\mathrm{old}}_{t+1} \times \widehat{\V}^{\mathrm{old}}_{t+1} \mid (u, v) \notin \mathcal{B}\}$. Add top $\widehat{c}_{\mathrm{oo\text{-}nobank}}$ pairs scored by $M_{\mathrm{oo\text{-}nobank}}$ to $\widehat{\E}_{t+1}$. \\
\textit{Phase III (Inductive Growth):} Define $\mathcal{C}_{\mathrm{on}} = \widehat{\V}^{\mathrm{old}}_{t+1} \times \widehat{\V}^{\mathrm{new}}_{t+1}$. Add top $\widehat{c}_{\mathrm{on}}$ pairs scored by $M_{\mathrm{on}}$ to $\widehat{\E}_{t+1}$. \\
\textit{Phase IV (New Community):} Define $\mathcal{C}_{\mathrm{nn}} = \{(u, v) \in \widehat{\V}^{\mathrm{new}}_{t+1} \times \widehat{\V}^{\mathrm{new}}_{t+1}, u \neq v\}$. Add top $\widehat{c}_{\mathrm{nn}}$ pairs scored by $M_{\mathrm{nn}}$ to $\widehat{\E}_{t+1}$. \\
\textit{Note:} Embeddings $h$ may optionally be updated via $\mathsf{Enc}$ between phases to reflect incremental structural changes.

\vspace{0.25em}
\textbf{5. EdgeBank Maintenance.} \\
Update $\mathcal{B} \leftarrow (\mathcal{B} \cup \widehat{\E}_{t+1}) \setminus \mathcal{E}_{expired}$ where $\mathcal{E}_{expired}$ are edges older than $k$ snapshots. \\

\Return{$\widehat{\G}_{t+1}=(\widehat{\V}_{t+1}, \widehat{\E}_{t+1}),\ \mathcal{B}$.}
\end{algorithm}

\subsection{Adapting Link-Prediction Baselines to Graph Construction} \label{sec:benchmarker_details}
Benchmark models are based on their ability to predict edge existence. For each potential edge candidate, the probability of edge $(u,v)$, the model outputs a probability $p_{uv}^t$ representing the likelihood of an edge appearing at timestamp $t$. We learn a global threshold for the dataset and the model to produce a binary adjacency matrix from the probability values. Specifically, an edge is included in the predicted graph if and only if
$$e_{uv}^{(t)}=\begin{cases}1 & \text{ if }p_{uv}^{(t)}\geq T\\ 0 & \text{otherwise} \end{cases}$$


Existing methods primarily perform link prediction by receiving a set of candidate edges, which are a subset of known possible edges in the universe. That is
$$\mathcal{E}_{cand} = \mathcal{E}_{pos} \cup \mathcal{E}_{neg}$$
Where $\mathcal{E}_{pos}$ is the set of ``true'' edges existing in the current snapshot (the positive candidates) and $\mathcal{E}_{neg}$ is a sampled set of non-existent edges (the negative candidates), where typically $|\mathcal{E}_{neg}| = X \cdot |\mathcal{E}_{pos}|$ for some $X \in \mathbb{Z}^+$.

However, for graph construction, because the set of known nodes in $\widehat{G}_{t+1}$ is unknown, we selected candidate edges as all possible edges between all nodes in the node universe across all timesteps $T$. That is
$$\E_{cand} = \{(u, v, t) \mid u, v \in V, u \neq v, t \in T\}$$

\xhdr{Edge selection and threshold optimization: } To translate the continuous probability scores from baseline models into a discrete adjacency structure, we use a global decision threshold $\tau$. We optimize $\tau$ using the validation partition $G_{val}$ by performing a grid search over the interval $[0, 1]$ with a step size of $0.01$. The optimal threshold $\tau^*$ is selected to maximize the F1-score across all validation snapshots:$$\tau^* = \arg\max_{\tau} \text{F1}(\widehat{\mathcal{E}}_{\tau}, \mathcal{E}_{val})$$where $\widehat{\mathcal{E}}_{\tau} = \{e \in \mathcal{E}_{cand} \mid p_e \geq \tau\}$. We then apply this optimized threshold to the test partition to reconstruct the final adjacency matrices.

\xhdr{Bounded Adjacency Reconstruction} To ensure scalability across datasets of varying densities and to prevent memory overflow (OOM) during the construction phase, we implement a heuristic cap on graph growth. For each predicted snapshot $\widehat{\mathcal{G}}_t$, the total number of edges is restricted to at most $k_t$, defined relative to the cardinality of the preceding snapshot:$$k_t = 5 \times |\mathcal{E}_{t-1}|$$If the number of candidate edges exceeding the decision threshold $\tau$ surpasses this limit, we perform top-$k$ filtering, selecting only the $k_t$ edges with the highest predicted probabilities. This approach ensures computational tractability while preserving the most structurally significant interactions according to the model's confidence scores.

\section{Stability Theorem}
\label{sec:proof}

In this section, we prove Proposition~\ref{prop:stability_edge_edits_main}. 
A natural question is whether the filtration-based descriptor $\Phi(\G_t)$ 
is a reliable snapshot state for forecasting: if two snapshots differ by 
only a few edges, do their descriptors remain close? This is essential for 
\method because the topology predictor is trained to forecast 
$\Phi(\G_{t+1})$ from past descriptors, and its predictions will inevitably 
carry small errors. If the descriptor were sensitive to minor structural 
perturbations, these errors could cascade into large reconstruction 
mistakes in the decoder. The following proposition shows that the 
degree-sublevel filtration descriptor is stable in a precise sense: editing 
at most $\delta$ edges in a snapshot perturbs the node-count sequence $X$ 
by at most $2\delta$ in $\ell_1$ norm, and perturbs the induced-edge-count 
sequence $Y$ by an amount controlled by $\delta$ and the filtration 
thresholds. This stability justifies using $\Phi(\G_t)$ as a forecasting 
target: small prediction errors in descriptor space correspond to small 
structural deviations in the reconstructed snapshot, making the 
inverse topology approach principled rather than heuristic.

\noindent \textbf{Proposition~\ref{prop:stability_edge_edits_main}} (Stability under edge edits).
\textit{Let $\G=(\V,\E)$ and $\G'=(\V,\E')$ be graphs on the same node set with $|\E\triangle \E'|\le \delta$.
Let $X,Y$ and $X',Y'$ be the degree-sublevel filtration node and induced-edge count sequences computed using thresholds
$\epsilon_1<\cdots<\epsilon_n$.}

\textit{(i) If the thresholds are consecutive integers $\{0,1,\ldots,D\}$, then
$\|X-X'\|_1 \le 2\delta$.}

\textit{(ii) For any thresholds, 
$\|Y-Y'\|_1 \le \delta\sum_{i=1}^n (1+2\epsilon_i)$.}
\begin{proof}
Let $\Delta d(v)=\deg_{\G}(v)-\deg_{\G'}(v)$.
Each edge in $\E\triangle \E'$ changes the degrees of exactly two
endpoints by $\pm 1$, hence
\[
\sum_{v\in V}|\Delta d(v)|\le 2|\E\triangle \E'|\le 2\delta.
\]

\textbf{Part (i).}
Assume thresholds are $k\in\{0,1,\ldots,D\}$ and write
$x_k=|\V_k(\G)|$ where $\V_k(\G)=\{v:\deg_{\G}(v)\le k\}$,
and similarly $x_k'$.
For each $k$,
\[
x_k-x_k'
=\sum_{v\in \V}
\Bigl(\mathbf{1}\{\deg_{\G}(v)\le k\}-\mathbf{1}\{\deg_{\G'}(v)\le k\}\Bigr),
\]
so by the triangle inequality,
\[
|x_k-x_k'|
\le\sum_{v\in \V}
\Bigl|\mathbf{1}\{\deg_{\G}(v)\le k\}-\mathbf{1}\{\deg_{\G'}(v)\le k\}\Bigr|.
\]
Summing over $k$ and swapping the order of summation gives
\[
\|X-X'\|_1
=\sum_{k=0}^D|x_k-x_k'|
\le\sum_{v\in \V}\sum_{k=0}^D
\Bigl|\mathbf{1}\{\deg_{\G}(v)\le k\}
     -\mathbf{1}\{\deg_{\G'}(v)\le k\}\Bigr|.
\]
For a fixed $v$, the inner sum counts how many integer thresholds
lie strictly between $\deg_{\G}(v)$ and $\deg_{\G'}(v)$, which is
exactly $|\Delta d(v)|$ whenever
$D\ge\max\{\deg_{\G}(v),\deg_{\G'}(v)\}$ and at most $|\Delta d(v)|$
otherwise.
Therefore,
\[
\|X-X'\|_1\le\sum_{v\in V}|\Delta d(v)|\le 2\delta.
\]

\textbf{Part (ii).}
Fix a threshold $\epsilon_i$ and let $\mS=\V_{\epsilon_i}(\G)$ and
$\mS'=\V_{\epsilon_i}(\G')$.
Let $\F=\E\triangle \E'$ denote the set of edited edges and
$\C=\E\cap \E'$ the set of common edges.
Any edge in the symmetric difference
$\E(\G[\mS])\triangle \E(\G'[\mS'])$
is either an edited edge (in $\F$) or a common edge whose
membership in the induced subgraph changes due to a vertex moving
between $\mS$ and $\mS'$.
Thus
\[
|\E(\G[\mS])\triangle \E(\G'[\mS'])|
\le|\F|
  +\bigl|\C\cap\bigl((\mS\times \mS)\triangle(\mS'\times \mS')\bigr)\bigr|.
\]
The first term satisfies $|\F|\le\delta$.
For the second term, a common edge changes membership in the
induced subgraph only if at least one of its endpoints lies in
$\mS\triangle \mS'$, so
\[
\bigl|\C\cap\bigl((\mS\times \mS)\triangle(\mS'\times \mS')\bigr)\bigr|
\le\sum_{v\in \mS\triangle \mS'}\deg_{\C}(v),
\]
where $\deg_{\C}(v)$ denotes the degree of $v$ in the common-edge
graph $(\V,\C)$.
For any $v\in \mS\triangle \mS'$, membership differs between $\mS$ and
$\mS'$ only if $\deg_{\G}(v)\ne\deg_{\G'}(v)$, and either $v\in \mS$ or
$v\in \mS'$, so $\min(\deg_{\G}(v),\deg_{\G'}(v))\le\epsilon_i$.
Since $\C\subseteq \E$ and $\C\subseteq \E'$, we have
\[
\deg_{\C}(v)\le\min\!\bigl(\deg_{\G}(v),\deg_{\G'}(v)\bigr)\le\epsilon_i.
\]
Moreover, $|\mS\triangle \mS'|\le\sum_{v\in \V}|\Delta d(v)|\le 2\delta$.
Combining,
\[
\bigl|\C\cap\bigl((\mS\times \mS)\triangle(\mS'\times \mS')\bigr)\bigr|
\le\epsilon_i\,|\mS\triangle \mS'|\le 2\delta\epsilon_i.
\]
Therefore,
\[
|y_i-y_i'|
\le|\E(\G[\mS])\triangle \E(\G'[\mS'])|
\le\delta+2\delta\epsilon_i
=\delta(1+2\epsilon_i).
\]
Summing over all $i$ gives the stated $\ell_1$ bound
\[
\|Y-Y'\|_1\le\delta\sum_{i=1}^n(1+2\epsilon_i). \qedhere
\]
\end{proof}

\section{Experimental Details}

This section provides additional experimental details omitted from the main text.
Appendix~\ref{appendix:baselines} summarizes the temporal graph baselines,
Appendix~\ref{sec:evaluation_metrics}
defines the graph construction metrics; Appendix~\ref{sec:vector_prediction}
reports the descriptor and budget prediction results used by \method, and Appendix~\ref{sec:costs} analyzes computational complexity.

\subsection{Baseline Models}
\label{appendix:baselines}
In this section, we give further details about the temporal graph learning models we used as a baseline for our work.

\textbf{TGCN} ~\citep{DBLP:journals/tits/ZhaoSZLWLDL20} is a combination of GCN and GRU. In particular, GCN is used to learn complex topological structures, while GRU is used to model embeddings dynamically to capture temporal dependence.

\textbf{HTGN}~\citep{DBLP:conf/kdd/0001ZKHK21} leverages the power of hyperbolic geometry, which is well-suited for capturing hierarchical structures and complex relationships in temporal networks. HTGN maps the temporal graph into hyperbolic space and utilizes hyperbolic graph neural networks and hyperbolic gated recurrent neural networks to model the evolving dynamics. It incorporates two key modules that are hyperbolic temporal contextual self-attention (HTA) and hyperbolic temporal consistency (HTC)-to ensure that temporal dependencies are effectively captured and that the model is both stable and generalizable across various tasks.

\textbf{GCLSTM}~\citep{DBLP:journals/apin/ChenWX22} combines a Graph Convolutional Network (GCN) and Long Short-Term Memory units to handle both the structural and temporal aspects of evolving networks. The GCN is used to capture the local structural properties of the network at each snapshot, while the LSTM learns the temporal evolution of these snapshots over time.

\textbf{EvolveGCN}~\citep{DBLP:conf/aaai/ParejaDCMSKKSL20} is designed to capture the temporal dynamics of graph-structured data. Instead of relying on static node embeddings, EvolveGCN evolves the parameters of a graph convolutional network (GCN) over time. By using a recurrent neural network (RNN) to adapt the GCN parameters, this model is capable of dynamically adjusting during both training and testing. This makes EvolveGCN suitable for evolving graphs with changing observed nodes, although it does not explicitly generate unseen nodes or full future snapshots.

\textbf{ROLAND}~\citep{DBLP:conf/kdd/YouDL22} is a dynamic graph learning framework that models node representations as hierarchical states, updated recurrently to capture temporal dependencies in evolving graphs. It supports scalable training using techniques like truncated backpropagation through time and meta-learning. In our DTDG setting, we use ROLAND to benchmark its performance and adaptability across diverse temporal networks.

 \textbf{VGRNN}~\cite{DBLP:conf/nips/HajiramezanaliH19} introduces a hierarchical variational model for dynamic graphs, extending the variational autoencoder to capture temporal dependencies. The framework utilizes high-level latent random variables within a graph recurrent neural network to model evolving topology and node attributes over time.

\subsection{Evaluation Metrics}
\label{sec:evaluation_metrics}

We evaluate each predicted snapshot
$\widehat{\G}_{t+1}=(\widehat{\V}_{t+1},\widehat{\E}_{t+1})$
against the corresponding ground-truth snapshot
$\G_{t+1}=(\V_{t+1},\E_{t+1})$ over the $N$ test timestamps. Our metrics
measure three complementary aspects of forecast quality: structural fidelity,
node and edge recovery, and budget errors.

\textbf{Structural fidelity.}
For a graph statistic $f$, we report the mean relative error
\[
\mathcal{S}_f
=
\frac{1}{N}
\sum_{t=1}^{N}
\left(
\frac{
f(\widehat{\G}_{t+1}) - f(\G_{t+1})
}{
f(\G_{t+1})+\varepsilon
}
\right),
\]
where $\varepsilon>0$ is a small constant used to avoid division by zero. Lower
values indicate closer agreement between the predicted and true snapshots. We
apply this evaluation to average node degree, number of unique degrees, average
degree centrality, assortativity coefficient, clustering coefficient, density,
number of triangles, and the filtration-descriptor error. For the descriptor
metric, both graphs are encoded using the degree-based filtration descriptor
$\Phi(\cdot)$, and we report the relative error between the resulting
descriptor vectors.

\begin{itemize}
    \item \textbf{Avg Node Degree}: Measures the error magnitude in matching the average node connectivity of the ground truth graph.
    \item \textbf{Unique Degree Count}: Evaluates the difference in distribution of distinct degree values across nodes.
    \item \textbf{Degree Centrality}: Tracks differences in centrality distribution errors between true and predicted graphs.
    \item \textbf{Assortativity Coefficient}: Measures the error in mixing patterns (e.g., whether high-degree nodes connect to other high-degree nodes).
    \item \textbf{Clustering Coefficient}: Evaluates local closure and clustering density
    \item \textbf{Density}: Measures the error in overall graph density.
    \item \textbf{Num Triangles}: Tracks the difference in local motif structure by counting triangle formations.
    \item \textbf{Descriptor Norm}: Measures the difference between the predicted structural descriptor and target descriptor through the L2 Norm.
    \item \textbf{Median Extra Nodes}: Measures the median number of over-predicted nodes across all graphs.
    \item \textbf{Median Missing Nodes}: Measures the median number of under-predicted nodes across all graphs.
    \item \textbf{Median Extra Edges}: Measures the median count of over-predicted edges across all graphs.
    \item \textbf{Median Missing Edges}: Measures the median count of under-predicted edges across all graphs.
\end{itemize}

\textbf{Node and edge recovery.}
We evaluate recovery of the true node and edge sets using precision, recall,
and F1 scores. For nodes, these metrics compare the predicted active node set
$\widehat{\V}_{t+1}$ with the true active node set $\V_{t+1}$. For edges, they
compare $\widehat{\E}_{t+1}$ with $\E_{t+1}$, either over all edges or within
the inductive edge categories old--old-bank, old--old-nobank, old--new, and
new--new. Aggregate scores are computed by averaging the corresponding metric
over all test snapshots:
\[
\mathcal{S}_f
=
\frac{1}{N}\sum_{t=1}^{N}
f(\widehat{\G}_{t+1},\G_{t+1}).
\]
For these metrics, higher values indicate better recovery.

\begin{itemize}
    \item \textbf{Precision Nodes / Recall Nodes / F1 Nodes}: Evaluates the overall accuracy of predicting the complete active node set through node IDs.
    \item \textbf{Precision Old Nodes / Recall Old Nodes / F1 Old Nodes}: Specifically measures how accurately the model identifies and retains reappearing historical (old) nodes through node IDs.
    \item \textbf{New Nodes Predicted}: Measures the number of new nodes predicted by counting the occurrences in the predicted graph.
    \item \textbf{oo-bank Precision / Recall / F1}: Measures accuracy for old-to-old, previously seen edges through Precision, Recall, and F1 by comparing involved node IDs.
    \item \textbf{oo-nobank Precision / Recall / F1}: Measures accuracy for old-to-old, never seen, edges through Precision, Recall, and F1 by comparing involved node IDs.
    \item \textbf{Edge Precision / Recall / F1}: Measures accuracy for all edge types through Precision, Recall, and F1 by comparing involved node IDs.
    \item \textbf{Num o-n Predicted}: Tracks old-to-new edge predictions.
    \item \textbf{Num n-n Predicted}: Tracks new-to-new edge predictions.
\end{itemize}

\textbf{Budget errors.}
We also report median errors in the predicted node and edge budgets. These
metrics quantify over-generation and under-generation separately:
\[
\mathrm{ExtraNodes}_t =
\max\{|\widehat{\V}_{t+1}|-|\V_{t+1}|,0\},
\qquad
\mathrm{MissingNodes}_t =
\max\{|\V_{t+1}|-|\widehat{\V}_{t+1}|,0\},
\]
and
\[
\mathrm{ExtraEdges}_t =
\max\{|\widehat{\E}_{t+1}|-|\E_{t+1}|,0\},
\qquad
\mathrm{MissingEdges}_t =
\max\{|\E_{t+1}|-|\widehat{\E}_{t+1}|,0\}.
\]
We summarize each quantity by its median over test snapshots:
\[
\mathcal{S}_f
=
\operatorname{median}
\left(
\{f_t\}_{t=1}^{N}
\right).
\]
Lower values indicate better calibration of the predicted graph size.
Together, these metrics assess whether a model reconstructs the correct global
structure, recovers the correct active nodes and edges, and avoids excessive
over- or under-generation.

\subsection{Descriptor Prediction Performance}
\label{sec:vector_prediction}
To predict the descriptor and probability vectors used to guide graph construction, we tested multiple prediction methods. We tested the models SMA~\cite{yule1921time}, VAR~\cite{sims1980macroeconomics}, V-EWMA~\cite{brown1959statistical}, SSM~\cite{kalman1960new}, and VECM~\cite{johansen1991estimation}.

The performance is detailed in Tables~\ref{tab:percent_error_nodes} and \ref{tab:percent_error_edges}. As the tables show, V-EWMA has the best average performance, achieving the lowest mean error for both node and edge prediction, with average errors of $8.12\%$ and $10.82\%$, respectively. This indicates that V-EWMA provides the most accurate joint prediction of the descriptor and probability vectors used in graph construction.

\begin{table}[H]
\centering
\setlength{\tabcolsep}{6pt}
\small

\caption{Percent Error Nodes}
\label{tab:percent_error_nodes}
\centering
\begin{tabular}{l | cccccc}
\toprule
\toprule
\textbf{Dataset} & \textbf{SMA} & \textbf{VAR} & \textbf{V-EWMA} & \textbf{SSM} & \textbf{VECM} \\
\midrule
Collegemsg & \underline{15.10} & 23.43 & \textbf{11.64} & 27.88 & 18.94 \\
Mathoverflow & \textbf{0.28} & -1.27 & \underline{0.41} & -3.65 & -1.76 \\
Adex & \textbf{7.41} & 60.63 & \underline{9.84} & 11.23 & 44.21 \\
Aeternity & -1.64 & 17.88 & \underline{0.85} & 3.42 & \textbf{-0.81} \\
Aion & \textbf{0.02} & 10.08 & \underline{-0.27} & 22.31 & 13.67 \\
Aragon & \underline{6.69} & 47.72 & \textbf{5.51} & 18.55 & 39.06 \\
Bancor & \textbf{0.28} & 46.64 & \underline{1.31} & 5.42 & 27.22 \\
Centra & \underline{43.65} & 63.00 & \textbf{37.33} & 64.64 & 57.76 \\
Cindicator & \underline{5.85} & 60.88 & \textbf{4.52} & 40.49 & 48.90 \\
Coindash & \textbf{1.89} & 119.84 & \underline{8.42} & 15.36 & 117.00 \\
Dgd & 14.15 & \textbf{4.95} & 10.85 & 22.94 & \underline{9.20} \\
Iconomi & \underline{17.30} & 26.76 & \textbf{15.62} & 34.18 & 24.04 \\
Reddit\_b & 0.84 & -3.24 & \underline{0.67} & \textbf{0.65} & -3.86 \\
Tgbl-wiki & 9.13 & \textbf{5.32} & 6.96 & 7.26 & \underline{6.44} \\
\midrule[\heavyrulewidth]
\textbf{Average} & \underline{8.64} & 34.47 & \textbf{8.12} & 19.33 & 28.57 \\
\bottomrule
\bottomrule

\end{tabular}
\end{table}

\vspace{2em} 

\begin{table}[ht]
\caption{Percent Error Edges}
\label{tab:percent_error_edges}
\centering
\begin{tabular}{l | cccccc}
\toprule
\toprule

\textbf{Dataset} & \textbf{SMA} & \textbf{VAR} & \textbf{V-EWMA} & \textbf{SSM} & \textbf{VECM} \\
\midrule
Collegemsg & 29.06 & 41.57 & \underline{23.65} & 46.19 & \textbf{20.99} \\
Mathoverflow & \textbf{1.11} & -1.25 & \underline{1.23} & -3.22 & -1.38 \\
Adex & \textbf{12.25} & 71.53 & 16.57 & \underline{14.23} & 55.26 \\
Aeternity & -2.21 & 13.14 & \textbf{-0.01} & \underline{2.02} & -2.08 \\
Aion & \underline{0.75} & 7.79 & \textbf{0.49} & 19.71 & 10.93 \\
Aragon & \underline{7.48} & 52.68 & \textbf{6.25} & 19.80 & 43.73 \\
Bancor & \textbf{1.50} & 45.89 & \underline{2.93} & 6.45 & 27.41 \\
Centra & \underline{51.01} & 56.46 & \textbf{41.90} & 80.14 & 65.42 \\
Cindicator & \underline{6.42} & 48.65 & \textbf{5.11} & 40.96 & 37.83 \\
Coindash & \textbf{3.67} & 103.73 & \underline{10.09} & 18.76 & 112.69 \\
Dgd & 18.87 & \textbf{9.40} & 14.79 & 29.81 & \underline{13.72} \\
Iconomi & \underline{20.77} & 26.55 & \textbf{18.55} & 37.92 & 23.83 \\
Reddit\_b & \underline{2.64} & -4.84 & \textbf{2.10} & 2.70 & -5.02 \\
Tgbl-wiki & 10.55 & \textbf{6.88} & \underline{7.88} & 8.42 & 9.59 \\
\midrule[\heavyrulewidth]
\textbf{Average} & \underline{11.70} & 34.16 & \textbf{10.82} & 23.13 & 29.49 \\
\bottomrule
\bottomrule

\end{tabular}
\end{table}

\subsection{Computational Complexity Analysis}
\label{sec:costs}
 
Let $n_t=|\V_t|$ and $m_t=|\E_t|$ denote the number of nodes and edges in the observed snapshot, and let $\widehat{n}_{t+1}=|\widehat{\V}_{t+1}|$ and $\widehat{m}_{t+1}=|\widehat{\E}_{t+1}|$ denote the predicted node and edge budgets. Let $d$ be the embedding dimension, $L$ the number of GCN layers, and $M$ the number of sampled candidate partners per node.

The filtration encoder computes degrees and cumulative node and edge counts over degree thresholds, which costs $O(n_t+m_t)$ per snapshot once the threshold set is fixed. The descriptor predictor operates on a fixed-length sequence of filtration vectors, so its cost does not scale with the graph size. The memory module scans the last $k$ snapshots to recover candidate old nodes and compute reappearance scores, with worst-case cost $O(\sum_{\tau=t-k}^{t} m_\tau)$. This avoids scoring the full historical node universe.

The decoder is the dominant term. Each decoding phase scores a restricted candidate set instead of all possible node pairs. With at most $M$ candidate partners per predicted node, candidate scoring costs $O(M\widehat{n}_{t+1}s(d))$, where $s(d)$ is the cost of scoring one candidate edge. Selecting the top edges adds $O(M\widehat{n}_{t+1}\log \widehat{m}_{t+1})$. Recomputing embeddings with an $L$-layer sparse GCN over the partially generated graph costs $O(L(\widehat{m}_{t+1}d+\widehat{n}_{t+1}d^2))$ across a constant number of decoding phases.

Thus, the per-snapshot inference cost is $O(n_t+m_t+\sum_{\tau=t-k}^{t}m_\tau+M\widehat{n}_{t+1}(s(d)+\log \widehat{m}_{t+1})+L(\widehat{m}_{t+1}d+\widehat{n}_{t+1}d^2))$. For fixed $k$, $M$, $L$, and descriptor length, the cost scales near-linearly in the observed memory window and the predicted snapshot size. Training has the same leading terms, with the edge-scoring cost multiplied by $(1+r)$ when $r$ negative edges are sampled per positive edge.

\section{Results By Dataset}
\label{sec:detailed_results}

In this section, we present a granular performance analysis across all evaluated datasets to contextualize the macro averages provided in Table~\ref{tab:forecast_fidelity_values}. \method consistently demonstrates superior predictive capacity, with the most significant gains occurring in \texttt{F1 Nodes}, \texttt{Edge F1}, and \texttt{F1 Old Nodes}.

\begin{figure}
\vspace{-1em}
\centering
\includegraphics[width=\linewidth]{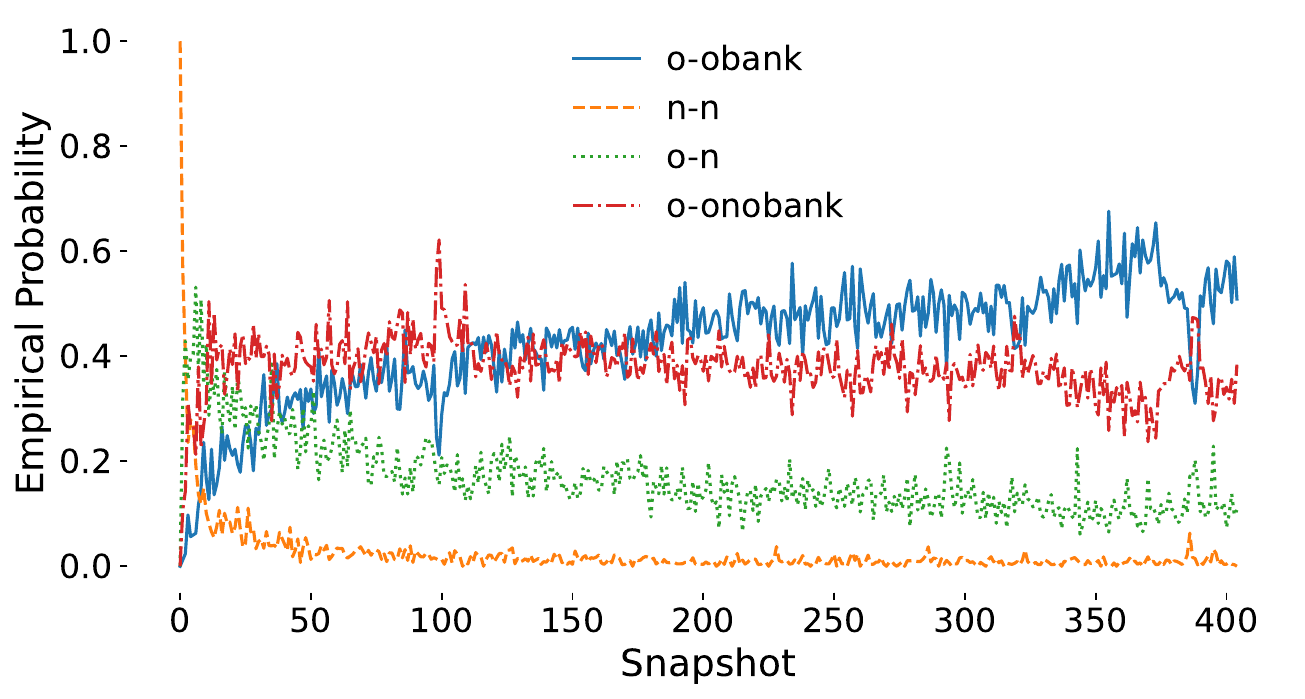}
\caption{Empirical probabilities of edge types over time for the Reddit-B network.}
\label{fig:redditEdges}
\vspace{-1em}
\end{figure}
 
Figure~\ref{fig:redditEdges} visualizes why fixed edge priors fail on temporal datasets. The empirical mix of o-obank, o-onobank, o-n, and n-n edges shifts over time, so a model that treats edge formation as stationary will allocate probability mass to the wrong edge types during regime changes.

Notably, while most baseline models struggle to maintain non-zero scores in edge-level forecasting, \method maintains a robust performance margin. Furthermore, the model exhibits substantial precision in capturing graph topology, as evidenced by the \texttt{Structure Evaluation} metrics. On average, \method outperforms the strongest baseline in $60\%$ of graph property tasks, frequently achieving results orders of magnitude closer to the ground truth than the nearest competing model.

Detailed results for all datasets are reported in Tables~\ref{tab:collegeMsg_results}--\ref{tab:tgbl_wiki_results}, covering CollegeMessage, MathOverflow, Adex, Aeternity, Aion, Aragon, Bancor, Centra, Cindicator, Coindash, DGD, Iconomi, Reddit\_B, and TGBL-Wiki.

\begin{table}[H] 
    \centering
    \caption{Comprehensive evaluation results—including edge, node, and structural metrics—for all benchmark models on \textbf{CollegeMsg}. \tozero indicates that values closer to 0 are better. \higherbetter denotes that higher values are better, while \lowerbetter denotes that lower values are better. \textbf{Bold} indicates best, the second-best is \underline{underlined}.}
    \label{tab:collegeMsg_results}
    \resizebox{1.00\linewidth}{!}{\begin{tabular}{l | c | c c c c c c | c}
\toprule
\toprule
& \textbf{Metric} & \textbf{HTGN} & \textbf{ROLAND} & \textbf{VGRNN} & \textbf{TGCN} & \textbf{GCLSTM} & \textbf{EvolveGCN} & \textbf{TopoGED} \\ \midrule
\multirow{11}{*}{\rotatebox{90}{\textbf{Edge Evaluation}}}&oo-bank Precision\higherbetter    & $0.00 \pm 0.00$ & $0.00 \pm 0.00$ & $0.00 \pm 0.00$ & $0.00 \pm 0.00$ & $0.00 \pm 0.00$ & $0.00 \pm 0.00$ & $\mathbf{0.26 \pm 0.12}$ \\
&oo-bank Recall\higherbetter       & $0.00 \pm 0.00$ & $0.00 \pm 0.00$ & $0.00 \pm 0.00$ & $0.00 \pm 0.00$ & $0.00 \pm 0.00$ & $0.00 \pm 0.00$ & $\mathbf{0.45 \pm 0.21}$  \\
&oo-bank F1\higherbetter           & $0.00 \pm 0.00$ & $0.00 \pm 0.00$ & $0.00 \pm 0.00$ & $0.00 \pm 0.00$ & $0.00 \pm 0.00$ & $0.00 \pm 0.00$ & $\mathbf{0.32 \pm 0.12}$ \\ 
&oo-nobank Precision\higherbetter  & $0.00 \pm 0.00$ & $0.00 \pm 0.00$ & $0.00 \pm 0.00$ & $0.00 \pm 0.00$ & $0.00 \pm 0.00$ & $0.00 \pm 0.00$ & $\mathbf{0.01 \pm 0.03}$ \\
&oo-nobank Recall\higherbetter     & $0.00 \pm 0.00$ & $0.00 \pm 0.00$ & $0.00 \pm 0.00$ & $0.00 \pm 0.00$ & $0.00 \pm 0.00$ & $0.00 \pm 0.00$ & $\mathbf{0.01 \pm 0.05}$ \\
&oo-nobank F1\higherbetter         & $0.00 \pm 0.00$ & $0.00 \pm 0.00$ & $0.00 \pm 0.00$ & $0.00 \pm 0.00$ & $0.00 \pm 0.00$ & $0.00 \pm 0.00$ & $\mathbf{0.01 \pm 0.04}$ \\ 
&Edge Precision\higherbetter       & $0.00 \pm 0.00$ & $0.00 \pm 0.00$ & $0.00 \pm 0.00$ & $0.00 \pm 0.00$ & $0.00 \pm 0.00$ & $0.00 \pm 0.00$ & $\mathbf{0.16 \pm 0.07}$ \\
&Edge Recall\higherbetter          & $0.00 \pm 0.00$ & $0.00 \pm 0.00$ & $0.00 \pm 0.00$ & $0.00 \pm 0.00$ & $0.00 \pm 0.00$ & $0.00 \pm 0.00$ & $\mathbf{0.25 \pm 0.13}$ \\
&Edge F1\higherbetter              & $0.00 \pm 0.00$ & $0.00 \pm 0.00$ & $0.00 \pm 0.00$ & $0.00 \pm 0.00$ & $0.00 \pm 0.00$ & $0.00 \pm 0.00$ & $\mathbf{0.19 \pm 0.07}$ \\ 
&Num o-n Predicted \tozero         & $1.57 \pm 2.62$ & $0.54 \pm 2.22$ & $\underline{-0.16 \pm 2.00}$ & $1.75 \pm 8.89$ & $\mathbf{0.15 \pm 2.07}$ & $1.15 \pm 3.13$ & $0.96 \pm 2.42$ \\
&Num n-n Predicted \tozero         & $\mathbf{-0.04 \pm 0.19}$ & $\mathbf{-0.04 \pm 0.19}$ & $\mathbf{-0.04 \pm 0.19}$ & $0.14 \pm 0.71$ & $\mathbf{-0.04 \pm 0.19}$ & $0.09 \pm 0.33$ & $\underline{0.07 \pm 0.60}$ \\

\midrule
\multirow{7}{*}{\rotatebox{90}{\textbf{Node Evaluation}}}&Precision Nodes\higherbetter     & $0.02 \pm 0.02$ & $0.02 \pm 0.02$ & $\underline{0.03 \pm 0.04}$ & $0.01 \pm 0.02$ & $0.02 \pm 0.02$ & $0.02 \pm 0.02$ & $\mathbf{0.43 \pm 0.11}$ \\
&Recall Nodes\higherbetter         & $0.06 \pm 0.04$ & $0.02 \pm 0.02$ & $0.02 \pm 0.02$ & $0.01 \pm 0.02$ & $0.01 \pm 0.01$ & $\underline{0.09 \pm 0.05}$ & $\mathbf{0.46 \pm 0.13}$ \\
&F1 Nodes\higherbetter             & $0.03 \pm 0.02$ & $0.02 \pm 0.02$ & $0.02 \pm 0.03$ & $0.01 \pm 0.02$ & $0.01 \pm 0.02$ & $\underline{0.04 \pm 0.02}$ & $\mathbf{0.44 \pm 0.10}$ \\
&Precision Old Nodes\higherbetter  & $0.02 \pm 0.02$ & $0.02 \pm 0.02$ & $\underline{0.03 \pm 0.04}$ & $0.00 \pm 0.01$ & $0.02 \pm 0.02$ & $0.02 \pm 0.01$ & $\mathbf{0.43 \pm 0.11}$ \\
&Recall Old Nodes\higherbetter     & $0.06 \pm 0.04$ & $0.02 \pm 0.02$ & $0.02 \pm 0.03$ & $0.00 \pm 0.01$ & $0.01 \pm 0.02$ & $\underline{0.09 \pm 0.05}$ & $\mathbf{0.46 \pm 0.13}$ \\
&F1 Old Nodes\higherbetter         & $\underline{0.04 \pm 0.02}$ & $0.02 \pm 0.02$ & $0.02 \pm 0.03$ & $0.00 \pm 0.01$ & $0.01 \pm 0.02$ & $\underline{0.04 \pm 0.02}$ & $\mathbf{0.43 \pm 0.10}$ \\
&New Nodes Predicted\higherbetter  & $\mathbf{-0.15 \pm 0.64}$ & $\underline{-0.30 \pm 0.71}$ & $-0.57 \pm 0.67$ & $-0.31 \pm 1.58$ & $-0.53 \pm 0.72$ & $1.03 \pm 2.96$ & $0.42 \pm 1.41$ \\

\midrule
\multirow{11}{*}{\rotatebox{90}{\textbf{Structure Evaluation}}}&Avg Node Degree \tozero      & $0.37 \pm 0.21$ & $2.59 \pm 1.19$ & $4.75 \pm 1.60$ & $3.18 \pm 0.73$ & $4.70 \pm 1.69$ & $\mathbf{-0.05 \pm 0.10}$ & $\underline{0.36 \pm 0.19}$  \\
&Unique Degree Count\tozero         & $0.72 \pm 0.55$ & $2.20 \pm 1.10$ & $1.89 \pm 1.03$ & $2.52 \pm 0.97$ & $2.44 \pm 1.03$ & $\mathbf{0.37 \pm 0.47}$ & $\underline{0.64 \pm 0.48}$ \\
&Degree Centrality\tozero           & $\underline{-0.45 \pm 0.11}$ & $3.02 \pm 2.58$ & $9.07 \pm 3.84$ & $4.20 \pm 1.01$ & $9.07 \pm 4.30$ & $-0.73 \pm 0.07$ & $\mathbf{0.28 \pm 0.24}$ \\
&Assortativity Coefficient\tozero   & $\underline{-5.19 \pm 6.47}$ & $6.98 \pm 17.90$ & $6.02 \pm 14.07$ & $5.45 \pm 13.81$ & $5.49 \pm 13.43$ & $-8.03 \pm 12.29$ & $\mathbf{1.13 \pm 7.92}$ \\
&Clustering Coefficient\tozero      & $0.39 \pm 1.03$ & $1.24 \pm 3.92$ & $1.64 \pm 4.80$ & $1.33 \pm 3.75$ & $1.52 \pm 4.12$ & $\underline{0.20 \pm 0.48}$ & $\mathbf{0.07 \pm 0.06}$ \\
&Density\tozero                     & $\underline{-0.45 \pm 0.11}$ & $3.02 \pm 2.58$ & $9.07 \pm 3.84$ & $4.20 \pm 1.01$ & $9.07 \pm 4.30$ & $-0.73 \pm 0.07$ & $\mathbf{0.28 \pm 0.24}$ \\
&Num Triangles\tozero               & $22.46 \pm 16.69$ & $130.32 \pm 100.27$ & $214.75 \pm 143.20$ & $147.96 \pm 99.07$ & $219.50 \pm 165.61$ & $\underline{18.43 \pm 22.63}$ & $\mathbf{3.11 \pm 3.20}$ \\ 
&Descriptor Norm\tozero             & $157.41 \pm 64.71$ & $91.78 \pm 36.40$ & $103.52 \pm 45.81$ & $\underline{86.98 \pm 27.94}$ & $100.15 \pm 37.70$ & $318.29 \pm 120.85$ & $\mathbf{35.42 \pm 11.91}$ \\
&Median Extra Nodes\tozero          & ${57.00 \pm 6.14}$ & ${0.00 \pm 1.34}$ & $\mathbf{0.00 \pm 0.00}$ & $\mathbf{0.00 \pm 0.00}$ & $\mathbf{0.00 \pm 0.00}$ & $93.50 \pm 10.56$ & $\underline{0.00 \pm 1.19}$ \\
&Median Missing Nodes\tozero        & $\mathbf{0.00 \pm 0.00}$ & $2.50 \pm 2.22$ & $15.00 \pm 1.34$ & $9.00 \pm 0.99$ & $15.00 \pm 2.19$ & $\mathbf{0.00 \pm 0.00}$ & $\underline{0.50 \pm 1.19}$ \\
&Median Extra Edges\tozero          & ${57.00 \pm 8.15}$ & $63.50 \pm 8.08$ & $\underline{57.00 \pm 8.11}$ & ${57.00 \pm 8.39}$ & ${57.00 \pm 8.39}$ & ${57.00 \pm 8.20}$ & $\mathbf{7.50 \pm 2.67}$ \\
&Median Missing Edges\tozero        & $\mathbf{0.00 \pm 0.00}$ & $\mathbf{0.00 \pm 0.00}$ & $\mathbf{0.00 \pm 0.00}$ & $\mathbf{0.00 \pm 0.00}$ & $\mathbf{0.00 \pm 0.00}$ & $\mathbf{0.00 \pm 0.00}$ & $\mathbf{0.00 \pm 0.00}$ \\

\bottomrule
\bottomrule

\end{tabular}}

\end{table} 

\begin{table}[H] 
    \centering
    \caption{Comprehensive evaluation results—including edge, node, and structural metrics—for all benchmark models on \textbf{MathOverflow}. \tozero indicates that values closer to 0 are better. \higherbetter denotes that higher values are better, while \lowerbetter denotes that lower values are better. \textbf{Bold} indicates best, the second-best is \underline{underlined}.}
    \label{tab:mathoverflow_results}
    \resizebox{1.00\linewidth}{!}{\begin{tabular}{l | c | c c c c c c | c}
\toprule
\toprule
& \textbf{Metric} & \textbf{HTGN} & \textbf{ROLAND} & \textbf{VGRNN} & \textbf{TGCN} & \textbf{GCLSTM} & \textbf{EvolveGCN} & \textbf{TopoGED} \\ \midrule
\multirow{11}{*}{\rotatebox{90}{\textbf{Edge Evaluation}}}&oo-bank Precision\higherbetter & $0.00 \pm 0.00$ & $0.00 \pm 0.00$ & $0.00 \pm 0.00$ & $0.00 \pm 0.00$ & $0.00 \pm 0.00$ & $0.00 \pm 0.00$ & $\mathbf{0.07 \pm 0.03}$ \\
&oo-bank Recall\higherbetter & $0.00 \pm 0.00$ & $0.00 \pm 0.00$ & $0.00 \pm 0.00$ & $0.00 \pm 0.00$ & $0.00 \pm 0.00$ & $0.00 \pm 0.00$ & $\mathbf{0.09 \pm 0.04}$ \\
&oo-bank F1\higherbetter & $0.00 \pm 0.00$ & $0.00 \pm 0.00$ & $0.00 \pm 0.00$ & $0.00 \pm 0.00$ & $0.00 \pm 0.00$ & $0.00 \pm 0.00$ & $\mathbf{0.08 \pm 0.04}$ \\
&oo-nobank Precision\higherbetter & $0.00 \pm 0.00$ & $0.00 \pm 0.00$ & $0.00 \pm 0.00$ & $0.00 \pm 0.00$ & $0.00 \pm 0.00$ & $0.00 \pm 0.00$ & $\mathbf{0.00 \pm 0.01}$ \\
&oo-nobank Recall\higherbetter & $0.00 \pm 0.00$ & $0.00 \pm 0.00$ & $0.00 \pm 0.00$ & $0.00 \pm 0.00$ & $0.00 \pm 0.00$ & $0.00 \pm 0.00$ & $\mathbf{0.00 \pm 0.01}$ \\
&oo-nobank F1\higherbetter & $0.00 \pm 0.00$ & $0.00 \pm 0.00$ & $0.00 \pm 0.00$ & $0.00 \pm 0.00$ & $0.00 \pm 0.00$ & $0.00 \pm 0.00$ & $\mathbf{0.00 \pm 0.01}$ \\
&Edge Precision\higherbetter & $0.00 \pm 0.00$ & $0.00 \pm 0.00$ & $0.00 \pm 0.00$ & $0.00 \pm 0.00$ & $0.00 \pm 0.00$ & $0.00 \pm 0.00$ & $\mathbf{0.02 \pm 0.01}$ \\
&Edge Recall\higherbetter & $0.00 \pm 0.00$ & $0.00 \pm 0.00$ & $0.00 \pm 0.00$ & $0.00 \pm 0.00$ & $0.00 \pm 0.00$ & $0.00 \pm 0.00$ & $\mathbf{0.03 \pm 0.01}$ \\
&Edge F1\higherbetter & $0.00 \pm 0.00$ & $0.00 \pm 0.00$ & $0.00 \pm 0.00$ & $0.00 \pm 0.00$ & $0.00 \pm 0.00$ & $0.00 \pm 0.00$ & $\mathbf{0.02 \pm 0.01}$ \\
&Num o-n Predicted\tozero & $1.67 \pm 1.16$ & $1.99 \pm 1.45$ & $2.56 \pm 1.91$ & $\underline{-0.36 \pm 0.31}$ & $3.57 \pm 2.53$ & $2.48 \pm 1.69$ & $\mathbf{0.14 \pm 0.40}$ \\
&Num n-n Predicted\tozero & $85.49 \pm 76.27$ & $82.35 \pm 79.78$ & $76.85 \pm 68.83$ & $98.11 \pm 87.26$ & $\underline{69.71 \pm 61.45}$ & $79.39 \pm 75.28$ & $\mathbf{0.45 \pm 1.63}$ \\

\midrule
\multirow{7}{*}{\rotatebox{90}{\textbf{Node Evaluation}}}&Precision Nodes\higherbetter & $0.01 \pm 0.00$ & $0.01 \pm 0.01$ & $0.01 \pm 0.02$ & $0.01 \pm 0.01$ & $\underline{0.02 \pm 0.02}$ & $0.01 \pm 0.00$ & $\mathbf{0.42 \pm 0.06}$ \\
&Recall Nodes\higherbetter & $0.01 \pm 0.01$ & $0.01 \pm 0.01$ & $0.00 \pm 0.01$ & $0.00 \pm 0.01$ & $0.01 \pm 0.01$ & $\underline{0.02 \pm 0.01}$ & $\mathbf{0.41 \pm 0.04}$ \\
&F1 Nodes\higherbetter & $\underline{0.01 \pm 0.01}$ & $\underline{0.01 \pm 0.01}$ & $\underline{0.01 \pm 0.01}$ & $\underline{0.01 \pm 0.01}$ & $\underline{0.01 \pm 0.01}$ & ${0.01 \pm 0.00}$ & $\mathbf{0.41 \pm 0.04}$ \\
&Precision Old Nodes\higherbetter & $0.05 \pm 0.04$ & $0.09 \pm 0.07$ & $0.11 \pm 0.20$ & $\underline{0.18 \pm 0.23}$ & $0.09 \pm 0.08$ & $0.05 \pm 0.03$ & $\mathbf{0.36 \pm 0.06}$ \\
&Recall Old Nodes\higherbetter & $0.01 \pm 0.01$ & $0.01 \pm 0.01$ & $0.01 \pm 0.01$ & $0.00 \pm 0.01$ & $0.01 \pm 0.01$ & $\underline{0.02 \pm 0.01}$ & $\mathbf{0.35 \pm 0.04}$ \\
&F1 Old Nodes\higherbetter & $0.02 \pm 0.01$ & $0.02 \pm 0.01$ & $0.01 \pm 0.01$ & $0.01 \pm 0.01$ & $0.02 \pm 0.01$ & $\underline{0.03 \pm 0.01}$ & $\mathbf{0.35 \pm 0.04}$ \\
&New Nodes Predicted\higherbetter & $10.94 \pm 5.47$ & $4.75 \pm 2.16$ & $\underline{1.61 \pm 1.21}$ & $2.61 \pm 1.61$ & $2.26 \pm 1.64$ & $17.13 \pm 8.81$ & $\mathbf{0.07 \pm 0.51}$ \\

\midrule
\multirow{11}{*}{\rotatebox{90}{\textbf{Structure Evaluation}}}&Avg Node Degree\tozero & $0.73 \pm 0.15$ & $2.47 \pm 0.80$ & $6.79 \pm 2.14$ & $4.92 \pm 0.48$ & $4.82 \pm 1.08$ & $\underline{0.13 \pm 0.15}$ & $\mathbf{0.07 \pm 0.06}$ \\
&Unique Degree Count\tozero & $\underline{0.80 \pm 0.28}$ & $2.22 \pm 0.57$ & $1.65 \pm 0.73$ & $2.72 \pm 0.58$ & $2.65 \pm 0.63$ & $1.52 \pm 0.41$ & $\mathbf{0.46 \pm 0.27}$ \\
&Degree Centrality\tozero & $\underline{0.11 \pm 0.17}$ & $3.79 \pm 2.31$ & $23.63 \pm 14.41$ & $12.46 \pm 2.13$ & $12.11 \pm 4.04$ & $-0.51 \pm 0.16$ & $\mathbf{0.09 \pm 0.17}$ \\
&Assortativity Coefficient\tozero & $-188.57 \pm 1073.62$ & $141.27 \pm 811.37$ & $144.25 \pm 825.39$ & $127.30 \pm 740.41$ & $147.56 \pm 851.57$ & $\underline{-54.55 \pm 314.30}$ & $\mathbf{51.44 \pm 300.12}$ \\
&Clustering Coefficient\tozero & $13.17 \pm 9.39$ & $18.98 \pm 14.78$ & $30.79 \pm 23.61$ & $29.39 \pm 20.61$ & $26.02 \pm 18.98$ & $\underline{6.50 \pm 5.57}$ & $\mathbf{2.45 \pm 3.26}$ \\
&Density \tozero & $\underline{0.11 \pm 0.17}$ & $3.79 \pm 2.31$ & $23.63 \pm 14.41$ & $12.46 \pm 2.13$ & $12.11 \pm 4.04$ & $-0.51 \pm 0.16$ & $\mathbf{0.09 \pm 0.17}$ \\
&Num Triangles\tozero & $\underline{174.69 \pm 103.80}$ & $418.85 \pm 271.60$ & $972.69 \pm 616.75$ & $646.03 \pm 386.53$ & $707.80 \pm 425.57$ & $209.57 \pm 153.77$ & $\mathbf{6.45 \pm 6.70}$ \\
&Descriptor Norm\tozero & $349.89 \pm 72.70$ & $\underline{287.23 \pm 48.77}$ & $378.51 \pm 103.85$ & $330.13 \pm 45.08$ & $329.96 \pm 49.60$ & $626.17 \pm 167.32$ & $\mathbf{98.76 \pm 25.30}$ \\
&Median Extra Nodes\tozero & ${80.00 \pm 4.28}$ & $\mathbf{0.00 \pm 0.00}$ & $\mathbf{0.00 \pm 0.00}$ & $\mathbf{0.00 \pm 0.00}$ & $\mathbf{0.00 \pm 0.00}$ & $206.00 \pm 14.61$ & $\underline{0.00 \pm 1.87}$ \\
&Median Missing Nodes\tozero & $\mathbf{0.00 \pm 0.00}$ & $25.00 \pm 6.56$ & $91.00 \pm 7.21$ & $78.00 \pm 4.21$ & $80.00 \pm 4.70$ & $\mathbf{0.00 \pm 0.00}$ & $\underline{3.00 \pm 3.53}$ \\
&Median Extra Edges\tozero & $230.00 \pm 14.01$ & $\underline{209.00 \pm 23.18}$ & $227.00 \pm 16.99$ & $227.00 \pm 15.30$ & $227.00 \pm 16.16$ & $227.00 \pm 16.05$ & $\mathbf{8.00 \pm 4.38}$ \\
&Median Missing Edges\tozero & $\mathbf{0.00 \pm 0.00}$ & $\mathbf{0.00 \pm 0.00}$ & $\mathbf{0.00 \pm 0.00}$ & $\mathbf{0.00 \pm 0.00}$ & $\mathbf{0.00 \pm 0.00}$ & $\mathbf{0.00 \pm 0.00}$ & $\underline{0.00 \pm 2.34}$ \\

\bottomrule
\bottomrule

\end{tabular}}

\end{table} 

\begin{table}[H] 
    \centering
    \caption{Comprehensive evaluation results—including edge, node, and structural metrics—for all benchmark models on \textbf{Adex}. \tozero indicates that values closer to 0 are better. \higherbetter denotes that higher values are better, while \lowerbetter denotes that lower values are better. \textbf{Bold} indicates best, the second-best is \underline{underlined}.}
    \label{tab:adex_results}
    \resizebox{1.00\linewidth}{!}{\begin{tabular}{l | c | c c c c c c | c}
\toprule
\toprule
& \textbf{Metric} & \textbf{HTGN} & \textbf{ROLAND} & \textbf{VGRNN} & \textbf{TGCN} & \textbf{GCLSTM} & \textbf{EvolveGCN} & \textbf{TopoGED} \\ \midrule
\multirow{11}{*}{\rotatebox{90}{\textbf{Edge Evaluation}}}&oo-bank Precision\higherbetter & $0.01 \pm 0.05$ & $0.00 \pm 0.00$ & $0.09 \pm 0.21$ & $0.00 \pm 0.00$ & $\mathbf{0.25 \pm 0.23}$ & $0.03 \pm 0.15$ & $\underline{0.20 \pm 0.12}$ \\
&oo-bank Recall\higherbetter & $0.00 \pm 0.00$ & $0.00 \pm 0.00$ & $0.01 \pm 0.03$ & $0.00 \pm 0.00$ & $\underline{0.06 \pm 0.06}$ & $0.00 \pm 0.02$ & $\mathbf{0.27 \pm 0.13}$ \\
&oo-bank F1\higherbetter & $0.00 \pm 0.01$ & $0.00 \pm 0.00$ & $0.02 \pm 0.05$ & $0.00 \pm 0.00$ & $\underline{0.09 \pm 0.08}$ & $0.00 \pm 0.03$ & $\mathbf{0.22 \pm 0.10}$ \\
&oo-nobank Precision\higherbetter & $0.00 \pm 0.00$ & $0.00 \pm 0.00$ & $0.00 \pm 0.00$ & $0.00 \pm 0.00$ & $0.00 \pm 0.00$ & $0.00 \pm 0.00$ & $\mathbf{0.01 \pm 0.03}$ \\
&oo-nobank Recall\higherbetter & $0.00 \pm 0.00$ & $0.00 \pm 0.00$ & $0.00 \pm 0.00$ & $0.00 \pm 0.00$ & $\mathbf{0.01 \pm 0.02}$ & $0.00 \pm 0.00$ & $\mathbf{0.01 \pm 0.03}$ \\
&oo-nobank F1\higherbetter & $0.00 \pm 0.00$ & $0.00 \pm 0.00$ & $0.00 \pm 0.00$ & $0.00 \pm 0.00$ & $0.00 \pm 0.00$ & $0.00 \pm 0.00$ & $\mathbf{0.01 \pm 0.03}$ \\
&Edge Precision\higherbetter & $0.00 \pm 0.00$ & $0.00 \pm 0.00$ & $0.00 \pm 0.00$ & $0.00 \pm 0.00$ & $0.00 \pm 0.00$ & $0.00 \pm 0.00$ & $\mathbf{0.10 \pm 0.05}$ \\
&Edge Recall\higherbetter & $0.00 \pm 0.00$ & $0.00 \pm 0.00$ & $0.00 \pm 0.01$ & $0.00 \pm 0.00$ & $\underline{0.02 \pm 0.02}$ & $0.00 \pm 0.01$ & $\mathbf{0.09 \pm 0.05}$ \\
&Edge F1\higherbetter & $0.00 \pm 0.00$ & $0.00 \pm 0.00$ & $0.00 \pm 0.00$ & $0.00 \pm 0.00$ & $\underline{0.01 \pm 0.00}$ & $0.00 \pm 0.00$ & $\mathbf{0.09 \pm 0.04}$ \\
&Num o-n Predicted\tozero & $-0.88 \pm 0.10$ & $3.41 \pm 1.97$ & $2.55 \pm 2.19$ & $-0.91 \pm 0.10$ & $4.05 \pm 3.35$ & $\underline{0.29 \pm 2.92}$ & $\mathbf{0.19 \pm 0.82}$ \\
&Num n-n Predicted\tozero & $\mathbf{-0.50 \pm 0.68}$ & $57.73 \pm 45.97$ & $43.89 \pm 45.76$ & $-0.89 \pm 0.32$ & $24.11 \pm 23.73$ & $21.25 \pm 52.82$ & $\underline{0.62 \pm 1.69}$ \\

\midrule
\multirow{7}{*}{\rotatebox{90}{\textbf{Node Evaluation}}}&Precision Nodes\higherbetter & $0.00 \pm 0.00$ & $0.04 \pm 0.03$ & $0.05 \pm 0.03$ & $0.00 \pm 0.00$ & $\underline{0.08 \pm 0.02}$ & $0.01 \pm 0.01$ & $\mathbf{0.55 \pm 0.12}$ \\
&Recall Nodes\higherbetter & $0.00 \pm 0.01$ & $0.02 \pm 0.01$ & $0.02 \pm 0.01$ & $0.00 \pm 0.00$ & $\underline{0.10 \pm 0.03}$ & $0.03 \pm 0.05$ & $\mathbf{0.56 \pm 0.11}$ \\
&F1 Nodes\higherbetter & $0.00 \pm 0.00$ & $0.03 \pm 0.01$ & $0.03 \pm 0.01$ & $0.00 \pm 0.00$ & $\underline{0.09 \pm 0.02}$ & $0.01 \pm 0.02$ & $\mathbf{0.54 \pm 0.09}$ \\
&Precision Old Nodes\higherbetter & $0.00 \pm 0.00$ & $0.07 \pm 0.03$ & $0.10 \pm 0.07$ & $0.00 \pm 0.00$ & $\underline{0.14 \pm 0.04}$ & $0.01 \pm 0.03$ & $\mathbf{0.31 \pm 0.09}$ \\
&Recall Old Nodes\higherbetter & $0.00 \pm 0.01$ & $0.03 \pm 0.02$ & $0.04 \pm 0.02$ & $0.00 \pm 0.00$ & $\underline{0.17 \pm 0.05}$ & $0.04 \pm 0.07$ & $\mathbf{0.32 \pm 0.07}$ \\
&F1 Old Nodes\higherbetter& $0.00 \pm 0.00$ & $0.04 \pm 0.02$ & $0.05 \pm 0.02$ & $0.00 \pm 0.00$ & $\underline{0.15 \pm 0.03}$ & $0.02 \pm 0.04$ & $\mathbf{0.30 \pm 0.05}$ \\
&New Nodes Predicted\higherbetter & $-0.85 \pm 0.12$ & $-0.33 \pm 0.35$ & $-0.48 \pm 0.29$ & $-0.98 \pm 0.01$ & $0.41 \pm 0.46$ & $\mathbf{0.08 \pm 1.82}$ & $\underline{0.12 \pm 0.59}$ \\

\midrule
\multirow{11}{*}{\rotatebox{90}{\textbf{Structure Evaluation}}}&Avg Node Degree\tozero & $-0.40 \pm 0.08$ & $7.91 \pm 2.93$ & $7.59 \pm 3.32$ & $4.46 \pm 1.44$ & $3.26 \pm 3.56$ & $\underline{-0.12 \pm 0.39}$ & $\mathbf{0.02 \pm 0.15}$ \\
&Unique Degree Count\tozero & $-0.60 \pm 0.11$ & $1.11 \pm 1.04$ & $0.61 \pm 0.56$ & $1.87 \pm 0.78$ & $0.83 \pm 0.95$ & $\mathbf{-0.06 \pm 0.65}$ & $\underline{0.09 \pm 0.28}$ \\
&Degree Centrality\tozero & $-0.85 \pm 0.05$ & $21.15 \pm 22.28$ & $21.76 \pm 14.62$ & $9.77 \pm 3.67$ & $3.10 \pm 5.80$ & $\underline{-0.71 \pm 0.15}$ & $\mathbf{0.05 \pm 0.43}$ \\
&Assortativity Coefficient\tozero & $-2.58 \pm 0.89$ & $1.29 \pm 0.93$ & $\underline{0.99 \pm 0.67}$ & $1.19 \pm 0.60$ & $1.30 \pm 0.67$ & $-1.33 \pm 1.27$ & $\mathbf{0.88 \pm 0.56}$ \\
&Clustering Coefficient\tozero & $\mathbf{0.37 \pm 1.45}$ & $38.29 \pm 72.55$ & $34.53 \pm 65.81$ & $35.92 \pm 67.31$ & $31.98 \pm 61.46$ & $6.35 \pm 13.46$ & $\underline{1.54 \pm 4.16}$ \\
&Density\tozero & $-0.85 \pm 0.05$ & $21.15 \pm 22.28$ & $21.76 \pm 14.62$ & $9.77 \pm 3.67$ & $3.10 \pm 5.80$ & $\underline{-0.71 \pm 0.15}$ & $\mathbf{0.05 \pm 0.43}$ \\
&Num Triangles\tozero & $\mathbf{2.92 \pm 11.49}$ & $2182.74 \pm 6414.98$ & $1277.94 \pm 1705.64$ & $755.62 \pm 1227.33$ & $895.09 \pm 2437.47$ & $99.45 \pm 498.74$ & $\underline{5.06 \pm 7.96}$ \\
&Descriptor Norm\tozero & $1479.50 \pm 1422.56$ & $501.81 \pm 485.32$ & $470.69 \pm 486.77$ & $\underline{351.54 \pm 442.18}$ & $495.11 \pm 627.77$ & $754.83 \pm 441.84$ & $\mathbf{180.45 \pm 279.10}$ \\
&Median Extra Nodes\tozero & $314.00 \pm 32.35$ & $\mathbf{0.00 \pm 0.00}$ & $\mathbf{0.00 \pm 0.00}$ & $\mathbf{0.00 \pm 0.00}$ & $36.00 \pm 9.65$ & $211.00 \pm 14.68$ & $\underline{4.00 \pm 3.93}$ \\
&Median Missing Nodes\tozero& $\mathbf{0.00 \pm 0.00}$ & $\underline{39.00 \pm 5.65}$ & $51.00 \pm 6.32$ & $41.00 \pm 6.52$ & $\mathbf{0.00 \pm 0.00}$ & $\mathbf{0.00 \pm 0.00}$ & $\underline{0.00 \pm 1.91}$ \\
&Median Extra Edges\tozero & $132.00 \pm 14.89$ & $294.00 \pm 25.96$ & $203.00 \pm 19.02$ & $148.00 \pm 19.28$ & $308.00 \pm 29.51$ & $\underline{129.00 \pm 16.17}$ & $\mathbf{1.00 \pm 4.89}$ \\
&Median Missing Edges\tozero & $\mathbf{0.00 \pm 0.00}$ & $\mathbf{0.00 \pm 0.00}$ & $\mathbf{0.00 \pm 0.00}$ & $\mathbf{0.00 \pm 0.00}$ & $\mathbf{0.00 \pm 0.00}$ & $\mathbf{0.00 \pm 0.00}$ & $\underline{0.00 \pm 2.50}$ \\

\bottomrule
\bottomrule

\end{tabular}}

\end{table} 

\begin{table}[H] 
    \centering
    \caption{Comprehensive evaluation results—including edge, node, and structural metrics—for all benchmark models on \textbf{Aeternity}. \tozero indicates that values closer to 0 are better. \higherbetter denotes that higher values are better, while \lowerbetter denotes that lower values are better. \textbf{Bold} indicates best, the second-best is \underline{underlined}.}
    \label{tab:aeternity_results}
    \resizebox{1.00\linewidth}{!}{\begin{tabular}{l | c | c c c c c c | c}
\toprule
\toprule
& \textbf{Metric} & \textbf{HTGN} & \textbf{ROLAND} & \textbf{VGRNN} & \textbf{TGCN} & \textbf{GCLSTM} & \textbf{EvolveGCN} & \textbf{TopoGED} \\ \midrule
\multirow{11}{*}{\rotatebox{90}{\textbf{Edge Evaluation}}}&oo-bank Precision\higherbetter & $0.00 \pm 0.00$ & $\underline{0.08 \pm 0.28}$ & $0.03 \pm 0.17$ & $0.00 \pm 0.00$ & $0.06 \pm 0.23$ & $0.06 \pm 0.17$ & $\mathbf{0.20 \pm 0.07}$ \\
&oo-bank Recall\higherbetter & $0.00 \pm 0.00$ & $0.00 \pm 0.00$ & $0.00 \pm 0.00$ & $0.00 \pm 0.00$ & $0.00 \pm 0.00$ & $0.00 \pm 0.01$ & $\mathbf{0.28 \pm 0.13}$ \\
&oo-bank F1\higherbetter & $0.00 \pm 0.00$ & $0.00 \pm 0.00$ & $0.00 \pm 0.00$ & $0.00 \pm 0.00$ & $0.00 \pm 0.00$ & $0.00 \pm 0.01$ & $\mathbf{0.22 \pm 0.07}$ \\
&oo-nobank Precision\higherbetter & $0.00 \pm 0.00$ & $0.00 \pm 0.00$ & $0.00 \pm 0.00$ & $0.00 \pm 0.00$ & $0.00 \pm 0.00$ & $0.00 \pm 0.02$ & $\mathbf{0.01 \pm 0.02}$ \\
&oo-nobank Recall\higherbetter & $0.00 \pm 0.00$ & $0.00 \pm 0.00$ & $0.00 \pm 0.00$ & $0.00 \pm 0.00$ & $0.00 \pm 0.00$ & $0.00 \pm 0.00$ & $\mathbf{0.01 \pm 0.03}$ \\
&oo-nobank F1\higherbetter  & $0.00 \pm 0.00$ & $0.00 \pm 0.00$ & $0.00 \pm 0.00$ & $0.00 \pm 0.00$ & $0.00 \pm 0.00$ & $0.00 \pm 0.00$ & $\mathbf{0.01 \pm 0.02}$ \\
&Edge Precision\higherbetter & $0.00 \pm 0.00$ & $0.00 \pm 0.00$ & $0.00 \pm 0.00$ & $0.00 \pm 0.00$ & $0.00 \pm 0.00$ & $0.00 \pm 0.00$ & $\mathbf{0.11 \pm 0.04}$  \\
&Edge Recall\higherbetter & $0.00 \pm 0.00$ & $0.00 \pm 0.00$ & $0.00 \pm 0.00$ & $0.00 \pm 0.00$ & $0.00 \pm 0.00$ & $0.00 \pm 0.00$ & $\mathbf{0.11 \pm 0.05}$ \\
&Edge F1\higherbetter & $0.00 \pm 0.00$ & $0.00 \pm 0.00$ & $0.00 \pm 0.00$ & $0.00 \pm 0.00$ & $0.00 \pm 0.00$ & $0.00 \pm 0.00$ & $\mathbf{0.11 \pm 0.04}$ \\
&Num o-n Predicted\tozero& $-0.81 \pm 0.12$ & $-0.10 \pm 0.44$ & $0.68 \pm 1.02$ & $\underline{-0.02 \pm 2.52}$ & $0.84 \pm 1.47$ & $0.31 \pm 2.05$ & $\mathbf{0.00 \pm 0.33}$  \\
&Num n-n Predicted\tozero & $128.35 \pm 75.97$ & $\underline{105.26 \pm 58.75}$ & $196.59 \pm 115.27$ & $114.37 \pm 83.36$ & $214.64 \pm 131.71$ & $233.41 \pm 141.90$ & $\mathbf{0.31 \pm 1.06}$ \\

\midrule
\multirow{7}{*}{\rotatebox{90}{\textbf{Node Evaluation}}}&Precision Nodes\higherbetter & $0.00 \pm 0.00$ & $\underline{0.01 \pm 0.01}$ & $\underline{0.01 \pm 0.01}$ & $0.00 \pm 0.00$ & ${0.01 \pm 0.00}$ & ${0.01 \pm 0.00}$ & $\mathbf{0.48 \pm 0.11}$ \\
&Recall Nodes\higherbetter & $0.00 \pm 0.00$ & $0.01 \pm 0.01$ & $0.00 \pm 0.00$ & $0.00 \pm 0.00$ & $0.01 \pm 0.01$ & $\underline{0.02 \pm 0.01}$ & $\mathbf{0.47 \pm 0.14}$  \\
&F1 Nodes\higherbetter & $0.00 \pm 0.00$ & $\underline{0.01 \pm 0.01}$ & ${0.01 \pm 0.00}$ & $0.00 \pm 0.00$ & ${0.01 \pm 0.00}$ & ${0.01 \pm 0.00}$ & $\mathbf{0.46 \pm 0.11}$ \\
&Precision Old Nodes\higherbetter & $0.06 \pm 0.05$ & $0.29 \pm 0.17$ & $\underline{0.37 \pm 0.33}$ & $0.02 \pm 0.06$ & $0.34 \pm 0.11$ & $\mathbf{0.41 \pm 0.12}$ & $0.30 \pm 0.09$ \\
&Recall Old Nodes\higherbetter & $0.01 \pm 0.01$ & $0.01 \pm 0.01$ & $0.01 \pm 0.00$ & $0.00 \pm 0.00$ & $0.02 \pm 0.01$ & $\underline{0.03 \pm 0.02}$ & $\mathbf{0.29 \pm 0.10}$ \\
&F1 Old Nodes\higherbetter & $0.01 \pm 0.01$ & $0.03 \pm 0.02$ & $0.01 \pm 0.01$ & $0.00 \pm 0.01$ & $0.03 \pm 0.02$ & $\underline{0.05 \pm 0.03}$ & $\mathbf{0.29 \pm 0.08}$  \\
&New Nodes Predicted\higherbetter & $11.32 \pm 4.69$ & $1.09 \pm 0.93$ & $\underline{0.59 \pm 0.73}$ & $2.29 \pm 2.83$ & $3.72 \pm 1.71$ & $6.18 \pm 3.06$ & $\mathbf{0.00 \pm 0.32}$ \\

\midrule
\multirow{11}{*}{\rotatebox{90}{\textbf{Structure Evaluation}}}&Avg Node Degree\tozero & $\underline{-0.26 \pm 0.43}$ & $2.94 \pm 1.26$ & $8.71 \pm 3.73$ & $2.68 \pm 1.89$ & $2.75 \pm 2.18$ & $2.08 \pm 3.19$ & $\mathbf{0.00 \pm 0.08}$ \\
&Unique Degree Count\tozero & $-0.33 \pm 0.38$ & $1.27 \pm 0.58$ & $0.64 \pm 0.56$ & $1.23 \pm 0.97$ & $1.41 \pm 1.21$ & $\underline{-0.14 \pm 0.85}$ & $\mathbf{-0.06 \pm 0.24}$  \\
&Degree Centrality\tozero & $\underline{-0.69 \pm 0.50}$ & $6.50 \pm 4.77$ & $30.12 \pm 56.24$ & $5.55 \pm 4.72$ & $3.52 \pm 7.69$ & $3.44 \pm 11.12$ & $\mathbf{0.10 \pm 0.44}$ \\
&Assortativity Coefficient\tozero & $-2.42 \pm 0.34$ & $0.74 \pm 0.42$ & $1.10 \pm 0.36$ & $0.80 \pm 0.45$ & $\underline{0.71 \pm 0.50}$ & $1.46 \pm 0.53$ & $\mathbf{0.64 \pm 0.31}$ \\
&Clustering Coefficient\tozero & $\underline{8.37 \pm 20.58}$ & $87.08 \pm 195.38$ & $105.80 \pm 206.62$ & $51.98 \pm 99.18$ & $88.01 \pm 196.81$ & $87.57 \pm 194.23$ & $\mathbf{6.70 \pm 14.43}$ \\
&Density\tozero & $\underline{-0.69 \pm 0.50}$ & $6.50 \pm 4.77$ & $30.12 \pm 56.24$ & $5.55 \pm 4.72$ & $3.52 \pm 7.69$ & $3.44 \pm 11.12$ & $\mathbf{0.10 \pm 0.44}$ \\
&Num Triangles\tozero & $\underline{112.86 \pm 289.18}$ & $1235.52 \pm 1708.21$ & $4023.98 \pm 7267.32$ & $832.73 \pm 871.23$ & $1655.07 \pm 3540.89$ & $1395.16 \pm 4023.10$ & $\mathbf{10.21 \pm 14.78}$ \\
&Descriptor Norm\tozero & $2185.37 \pm 1115.69$ & $\underline{666.41 \pm 657.85}$ & $1289.74 \pm 1286.05$ & $754.89 \pm 728.87$ & $1272.95 \pm 1046.26$ & $1567.32 \pm 1128.84$ & $\mathbf{384.54 \pm 599.03}$ \\
&Median Extra Nodes\tozero & $625.00 \pm 25.84$ & $\mathbf{0.00 \pm 0.00}$ & $\mathbf{0.00 \pm 0.00}$ & $\mathbf{0.00 \pm 4.27}$ & $138.00 \pm 16.87$ & $322.50 \pm 19.87$ & $\underline{1.00 \pm 5.34}$ \\
&Median Missing Nodes\tozero & $\mathbf{0.00 \pm 0.00}$ & $98.00 \pm 22.37$ & $106.00 \pm 11.25$ & $58.50 \pm 10.87$ & $\mathbf{0.00 \pm 0.44}$ & $\mathbf{0.00 \pm 0.00}$ & $\underline{2.00 \pm 12.34}$ \\
&Median Extra Edges\tozero & $325.50 \pm 33.60$ & $\underline{309.00 \pm 29.77}$ & $758.00 \pm 65.10$ & $318.50 \pm 31.70$ & $817.50 \pm 76.77$ & $859.50 \pm 86.60$ & $\mathbf{1.50 \pm 7.05}$ \\
&Median Missing Edges\tozero & $\mathbf{0.00 \pm 0.00}$ & $\mathbf{0.00 \pm 0.00}$ & $\mathbf{0.00 \pm 0.00}$ & $\mathbf{0.00 \pm 0.00}$ & $\mathbf{0.00 \pm 0.00}$ & $\mathbf{0.00 \pm 0.00}$ & $\underline{0.00 \pm 15.72}$  \\

\bottomrule
\bottomrule

\end{tabular}}

\end{table} 

\begin{table}[H] 
    \centering
    \caption{Comprehensive evaluation results—including edge, node, and structural metrics—for all benchmark models on \textbf{Aion}. \tozero indicates that values closer to 0 are better. \higherbetter denotes that higher values are better, while \lowerbetter denotes that lower values are better. \textbf{Bold} indicates best, the second-best is \underline{underlined}.}
    \label{tab:aion_results}
    \resizebox{1.00\linewidth}{!}{\begin{tabular}{l | c | c c c c c c | c}
\toprule
\toprule
& \textbf{Metric} & \textbf{HTGN} & \textbf{ROLAND} & \textbf{VGRNN} & \textbf{TGCN} & \textbf{GCLSTM} & \textbf{EvolveGCN} & \textbf{TopoGED} \\ \midrule
\multirow{11}{*}{\rotatebox{90}{\textbf{Edge Evaluation}}}&oo-bank Precision\higherbetter & $0.00 \pm 0.00$ & $0.00 \pm 0.00$ & $0.00 \pm 0.00$ & $0.00 \pm 0.00$ & $\mathbf{0.13 \pm 0.35}$ & $0.00 \pm 0.00$ & $\underline{0.07 \pm 0.03}$ \\
&oo-bank Recall\higherbetter & $0.00 \pm 0.00$ & $0.00 \pm 0.00$ & $0.00 \pm 0.00$ & $0.00 \pm 0.00$ & $\underline{0.00 \pm 0.01}$ & $0.00 \pm 0.00$ & $\mathbf{0.16 \pm 0.08}$ \\
&oo-bank F1\higherbetter & $0.00 \pm 0.00$ & $0.00 \pm 0.00$ & $0.00 \pm 0.00$ & $0.00 \pm 0.00$ & $\underline{0.01 \pm 0.01}$ & $0.00 \pm 0.00$ & $\mathbf{0.09 \pm 0.04}$ \\
&oo-nobank Precision\higherbetter  & $0.00 \pm 0.00$ & $0.00 \pm 0.00$ & $0.00 \pm 0.00$ & $0.00 \pm 0.00$ & $\underline{0.02 \pm 0.09}$ & $0.00 \pm 0.00$ & $\mathbf{0.03 \pm 0.03}$ \\
&oo-nobank Recall\higherbetter & $0.00 \pm 0.00$ & $0.00 \pm 0.00$ & $0.00 \pm 0.00$ & $0.00 \pm 0.00$ & $0.00 \pm 0.00$ & $0.00 \pm 0.00$ & $\mathbf{0.01 \pm 0.02}$ \\
&oo-nobank F1\higherbetter & $0.00 \pm 0.00$ & $0.00 \pm 0.00$ & $0.00 \pm 0.00$ & $0.00 \pm 0.00$ & $0.00 \pm 0.00$ & $0.00 \pm 0.00$ & $\mathbf{0.02 \pm 0.02}$ \\
&Edge Precision\higherbetter & $0.00 \pm 0.00$ & $0.00 \pm 0.00$ & $0.00 \pm 0.00$ & $0.00 \pm 0.00$ & $0.00 \pm 0.00$ & $0.00 \pm 0.00$ & $\mathbf{0.15 \pm 0.05}$ \\
&Edge Recall\higherbetter & $0.00 \pm 0.00$ & $0.00 \pm 0.00$ & $0.00 \pm 0.00$ & $0.00 \pm 0.00$ & $0.00 \pm 0.00$ & $0.00 \pm 0.00$ & $\mathbf{0.15 \pm 0.06}$ \\
&Edge F1\higherbetter & $0.00 \pm 0.00$ & $0.00 \pm 0.00$ & $0.00 \pm 0.00$ & $0.00 \pm 0.00$ & $0.00 \pm 0.00$ & $0.00 \pm 0.00$ & $\mathbf{0.15 \pm 0.05}$ \\
&Num o-n Predicted\tozero  & $-0.74 \pm 0.07$ & $-0.79 \pm 0.07$ & $0.52 \pm 0.77$ & $-0.84 \pm 0.11$ & $\underline{-0.17 \pm 0.22}$ & $-0.47 \pm 0.36$ & $\mathbf{0.02 \pm 0.22}$ \\
&Num n-n Predicted\tozero & $202.57 \pm 177.05$ & $206.59 \pm 200.54$ & $\underline{174.04 \pm 157.08}$ & $260.36 \pm 206.20$ & $176.97 \pm 153.41$ & $196.50 \pm 172.56$ & $\mathbf{1.05 \pm 3.46}$ \\

\midrule
\multirow{7}{*}{\rotatebox{90}{\textbf{Node Evaluation}}}&Precision Nodes\higherbetter & $0.00 \pm 0.00$ & $0.00 \pm 0.00$ & $0.00 \pm 0.00$ & $0.00 \pm 0.00$ & $\underline{0.02 \pm 0.01}$ & $0.01 \pm 0.02$ & $\mathbf{0.45 \pm 0.06}$ \\
&Recall Nodes\higherbetter & $0.00 \pm 0.00$ & $\underline{0.01 \pm 0.00}$ & $0.00 \pm 0.00$ & $0.00 \pm 0.00$ & $\underline{0.01 \pm 0.00}$ & $0.00 \pm 0.00$ & $\mathbf{0.45 \pm 0.07}$ \\
&F1 Nodes\higherbetter & $0.00 \pm 0.00$ & $0.00 \pm 0.00$ & $0.00 \pm 0.00$ & $0.00 \pm 0.00$ & $\underline{0.01 \pm 0.00}$ & $0.00 \pm 0.01$ & $\mathbf{0.45 \pm 0.05}$\\
&Precision Old Nodes\higherbetter & $0.01 \pm 0.02$ & $0.09 \pm 0.03$ & $0.08 \pm 0.08$ & $0.00 \pm 0.01$ & $\mathbf{0.35 \pm 0.11}$ & $\underline{0.18 \pm 0.21}$ & $0.14 \pm 0.04$ \\
&Recall Old Nodes\higherbetter & $0.00 \pm 0.00$ & $\underline{0.01 \pm 0.01}$ & $0.00 \pm 0.00$ & $0.00 \pm 0.00$ & $\underline{0.01 \pm 0.00}$ & $0.00 \pm 0.01$ & $\mathbf{0.13 \pm 0.04}$ \\
&F1 Old Nodes\higherbetter & $0.00 \pm 0.00$ & $\underline{0.02 \pm 0.01}$ & $0.01 \pm 0.01$ & $0.00 \pm 0.00$ & $\underline{0.02 \pm 0.01}$ & $0.01 \pm 0.01$ & $\mathbf{0.13 \pm 0.04}$ \\
&New Nodes Predicted\higherbetter & $5.77 \pm 1.39$ & $4.28 \pm 1.34$ & $\underline{0.11 \pm 0.36}$ & $5.03 \pm 3.45$ & $-0.33 \pm 0.13$ & $-0.50 \pm 0.11$ & $\mathbf{0.01 \pm 0.21}$ \\

\midrule
\multirow{11}{*}{\rotatebox{90}{\textbf{Structure Evaluation}}}&Avg Node Degree\tozero  & $\underline{-0.09 \pm 0.10}$ & $0.17 \pm 0.12$ & $5.35 \pm 1.64$ & $1.33 \pm 2.48$ & $7.78 \pm 0.98$ & $11.41 \pm 2.50$ & $\mathbf{0.01 \pm 0.06}$ \\
&Unique Degree Count\tozero& $\mathbf{0.02 \pm 0.20}$ & $\underline{0.08 \pm 0.45}$ & $0.88 \pm 0.61$ & $0.46 \pm 1.36$ & $3.11 \pm 0.84$ & $2.69 \pm 0.81$ & $-0.17 \pm 0.21$ \\
&Degree Centrality\tozero & $-0.66 \pm 0.08$ & $\underline{-0.42 \pm 0.21}$ & $14.79 \pm 9.11$ & $3.16 \pm 8.66$ & $31.15 \pm 7.68$ & $63.10 \pm 25.23$ & $\mathbf{0.05 \pm 0.22}$ \\
&Assortativity Coefficient\tozero & $-2.60 \pm 0.44$ & $1.12 \pm 0.53$ & $0.87 \pm 0.46$ & $1.27 \pm 0.83$ & $\underline{0.19 \pm 0.38}$ & $\mathbf{-0.05 \pm 0.31}$ & $1.08 \pm 0.37$ \\
&Clustering Coefficient\tozero & $10.20 \pm 11.05$ & $\mathbf{5.57 \pm 11.05}$ & $76.65 \pm 82.51$ & $12.08 \pm 30.30$ & $59.78 \pm 63.33$ & $66.35 \pm 71.37$ & $\underline{7.38 \pm 10.13}$ \\
&Density\tozero & $-0.66 \pm 0.08$ & $\underline{-0.42 \pm 0.21}$ & $14.79 \pm 9.11$ & $3.16 \pm 8.66$ & $31.15 \pm 7.68$ & $63.10 \pm 25.23$ & $\mathbf{0.05 \pm 0.22}$ \\
&Num Triangles\tozero & $87.45 \pm 65.09$ & $\underline{48.10 \pm 66.90}$ & $1912.40 \pm 1276.79$ & $490.73 \pm 767.46$ & $2594.00 \pm 1728.90$ & $3245.35 \pm 2425.08$ & $\mathbf{15.53 \pm 16.14}$ \\
&Descriptor Norm\tozero & $1330.29 \pm 301.01$ & $1164.38 \pm 502.01$ & $\underline{747.93 \pm 226.62}$ & $1902.54 \pm 1532.02$ & $790.99 \pm 218.30$ & $909.13 \pm 280.69$ & $\mathbf{279.83 \pm 115.95}$ \\
&Median Extra Nodes\tozero & $475.50 \pm 19.30$ & ${321.50 \pm 22.01}$ & $\mathbf{0.00 \pm 0.00}$ & $353.50 \pm 104.50$ & $\mathbf{0.00 \pm 0.00}$ & $\mathbf{0.00 \pm 0.00}$ & $\underline{0.00 \pm 3.01}$ \\
&Median Missing Nodes\tozero & $\mathbf{0.00 \pm 0.00}$ & $\mathbf{0.00 \pm 0.00}$ & $146.50 \pm 12.55$ & $\mathbf{0.00 \pm 0.00}$ & $184.50 \pm 22.56$ & $200.00 \pm 28.53$ & $\underline{8.50 \pm 6.23}$ \\
&Median Extra Edges\tozero & $379.50 \pm 25.05$ & $\underline{363.50 \pm 26.38}$ & $409.00 \pm 57.09$ & $465.00 \pm 65.47$ & $378.50 \pm 24.70$ & $386.00 \pm 26.84$ & $\mathbf{2.50 \pm 5.93}$ \\
&Median Missing Edges\tozero & $\mathbf{0.00 \pm 0.00}$ & $\mathbf{0.00 \pm 0.00}$ & $\mathbf{0.00 \pm 0.00}$ & $\mathbf{0.00 \pm 0.00}$ & $\mathbf{0.00 \pm 0.00}$ & $\mathbf{0.00 \pm 0.00}$ & $\underline{4.00 \pm 6.50}$ \\

\bottomrule
\bottomrule

\end{tabular}}

\end{table} 

\begin{table}[H] 
    \centering
    \caption{Comprehensive evaluation results—including edge, node, and structural metrics—for all benchmark models on \textbf{Aragon}. \tozero indicates that values closer to 0 are better. \higherbetter denotes that higher values are better, while \lowerbetter denotes that lower values are better. \textbf{Bold} indicates best, the second-best is \underline{underlined}. $OOM$ indicate out-of-memory.}
    \label{tab:aragon_results}
    \resizebox{1.00\linewidth}{!}{\begin{tabular}{l | c | c c c c c c | c}
\toprule
\toprule
& \textbf{Metric} & \textbf{HTGN} & \textbf{ROLAND} & \textbf{VGRNN} & \textbf{TGCN} & \textbf{GCLSTM} & \textbf{EvolveGCN} & \textbf{TopoGED} \\ \midrule
\multirow{11}{*}{\rotatebox{90}{\textbf{Edge Evaluation}}}&oo-bank Precision\higherbetter & $0.00 \pm 0.00$ & $0.00 \pm 0.00$ & $OOM$ & $0.00 \pm 0.00$ & $0.00 \pm 0.00$ & $0.00 \pm 0.00$ & $\mathbf{0.07 \pm 0.06}$ \\
&oo-bank Recall\higherbetter & $0.00 \pm 0.00$ & $0.00 \pm 0.00$ & $OOM$ & $0.00 \pm 0.00$ & $0.00 \pm 0.00$ & $0.00 \pm 0.00$ & $\mathbf{0.13 \pm 0.14}$ \\
&oo-bank F1\higherbetter & $0.00 \pm 0.00$ & $0.00 \pm 0.00$ & $OOM$ & $0.00 \pm 0.00$ & $0.00 \pm 0.00$ & $0.00 \pm 0.00$ & $\mathbf{0.08 \pm 0.08}$ \\
&oo-nobank Precision\higherbetter & $0.00 \pm 0.00$ & $0.00 \pm 0.00$ & $OOM$ & $0.00 \pm 0.00$ & $0.00 \pm 0.00$ & $0.00 \pm 0.00$ & $\mathbf{0.00 \pm 0.01}$ \\
&oo-nobank Recall\higherbetter & $0.00 \pm 0.00$ & $0.00 \pm 0.00$ & $OOM$ & $0.00 \pm 0.00$ & $0.00 \pm 0.00$ & $0.00 \pm 0.00$ & $\mathbf{0.00 \pm 0.01}$ \\
&oo-nobank F1\higherbetter & $0.00 \pm 0.00$ & $0.00 \pm 0.00$ & $OOM$ & $0.00 \pm 0.00$ & $0.00 \pm 0.00$ & $0.00 \pm 0.00$ & $\mathbf{0.00 \pm 0.01}$ \\
&Edge Precision\higherbetter & $0.00 \pm 0.00$ & $0.00 \pm 0.00$ & $OOM$ & $0.00 \pm 0.00$ & $0.00 \pm 0.00$ & $0.00 \pm 0.00$ & $\mathbf{0.22 \pm 0.08}$ \\
&Edge Recall\higherbetter & $0.00 \pm 0.00$ & $0.00 \pm 0.00$ & $OOM$ & $0.00 \pm 0.00$ & $0.00 \pm 0.00$ & $0.00 \pm 0.00$ & $\mathbf{0.23 \pm 0.09}$ \\
&Edge F1\higherbetter & $0.00 \pm 0.00$ & $0.00 \pm 0.00$ & $OOM$ & $0.00 \pm 0.00$ & $0.00 \pm 0.00$ & $0.00 \pm 0.00$ & $\mathbf{0.22 \pm 0.08}$ \\
&Num o-n Predicted\tozero & $\underline{-0.63 \pm 0.10}$ & $-0.87 \pm 0.04$ & $OOM$ & $-0.90 \pm 0.04$ & $-0.78 \pm 0.20$ & $-0.77 \pm 0.33$ & $\mathbf{0.07 \pm 0.29}$ \\
&Num n-n Predicted\tozero & $\underline{157.99 \pm 97.18}$ & $173.19 \pm 108.70$ & $OOM$ & $171.62 \pm 104.11$ & $335.32 \pm 196.71$ & $168.11 \pm 109.39$ & $\mathbf{0.58 \pm 1.81}$ \\

\midrule
\multirow{7}{*}{\rotatebox{90}{\textbf{Node Evaluation}}}&Precision Nodes\higherbetter & $0.00 \pm 0.00$ & $0.00 \pm 0.00$ & $OOM$ & $0.00 \pm 0.00$ & $\underline{0.01 \pm 0.00}$ & $0.00 \pm 0.00$ & $\mathbf{0.51 \pm 0.06}$ \\
&Recall Nodes\higherbetter & $0.00 \pm 0.00$ & $0.00 \pm 0.00$ & $OOM$ & $0.00 \pm 0.00$ & $\underline{0.01 \pm 0.01}$ & $0.00 \pm 0.01$ & $\mathbf{0.53 \pm 0.08}$ \\
&F1 Nodes\higherbetter & $0.00 \pm 0.00$ & $0.00 \pm 0.00$ & $OOM$ & $0.00 \pm 0.00$ & $\underline{0.01 \pm 0.00}$ & $0.00 \pm 0.00$ & $\mathbf{0.51 \pm 0.04}$ \\
&Precision Old Nodes\higherbetter & $0.00 \pm 0.01$ & $0.01 \pm 0.03$ & $OOM$ & $0.00 \pm 0.00$ & $\mathbf{0.28 \pm 0.10}$ & $0.06 \pm 0.09$ & $\underline{0.16 \pm 0.04}$ \\
&Recall Old Nodes\higherbetter & $0.00 \pm 0.00$ & $0.00 \pm 0.00$ & $OOM$ & $0.00 \pm 0.00$ & $\underline{0.02 \pm 0.01}$ & $0.01 \pm 0.02$ & $\mathbf{0.16 \pm 0.05}$ \\
&F1 Old Nodes\higherbetter & $0.00 \pm 0.00$ & $0.00 \pm 0.01$ & $OOM$ & $0.00 \pm 0.00$ & $\underline{0.04 \pm 0.02}$ & $0.01 \pm 0.02$ & $\mathbf{0.16 \pm 0.04}$ \\
&New Nodes Predicted\higherbetter & $8.14 \pm 2.58$ & $3.33 \pm 1.38$ & $OOM$ & $\underline{0.08 \pm 0.44}$ & $2.53 \pm 1.10$ & $3.36 \pm 1.30$ & $\mathbf{0.06 \pm 0.27}$ \\

\midrule
\multirow{11}{*}{\rotatebox{90}{\textbf{Structure Evaluation}}}&Avg Node Degree\tozero & $-0.44 \pm 0.03$ & $0.26 \pm 0.28$ & $OOM$ & $4.21 \pm 1.17$ & $2.14 \pm 1.14$ & $\underline{0.25 \pm 0.23}$ & $\mathbf{0.01 \pm 0.06}$ \\
&Unique Degree Count\tozero & $-0.79 \pm 0.11$ & $-0.50 \pm 0.24$ & $OOM$ & $2.53 \pm 0.95$ & $\underline{0.45 \pm 0.70}$ & $1.07 \pm 0.69$ & $\mathbf{0.13 \pm 0.24}$ \\ 
&Degree Centrality\tozero & $-0.87 \pm 0.04$ & $\underline{-0.31 \pm 0.47}$ & $OOM$ & $10.29 \pm 4.69$ & $1.09 \pm 1.33$ & $-0.37 \pm 0.28$ & $\mathbf{0.01 \pm 0.27}$ \\ 
&Assortativity Coefficient\tozero & $-2.64 \pm 1.47$ & $2.08 \pm 0.47$ & $OOM$ & $0.90 \pm 0.35$ & $1.84 \pm 0.44$ & $\mathbf{-0.07 \pm 1.11}$ & $\underline{0.79 \pm 0.33}$ \\ 
&Clustering Coefficient\tozero & $\mathbf{-0.82 \pm 0.38}$ & $\underline{-0.83 \pm 0.38}$ & $OOM$ & $23.32 \pm 19.28$ & $29.93 \pm 23.87$ & $4.42 \pm 5.72$ & $1.01 \pm 2.22$ \\ 
&Density\tozero & $-0.87 \pm 0.04$ & $\underline{-0.31 \pm 0.47}$ & $OOM$ & $10.29 \pm 4.69$ & $1.09 \pm 1.33$ & $-0.37 \pm 0.28$ & $\mathbf{0.01 \pm 0.27}$ \\ 
&Num Triangles\tozero & $\mathbf{-0.75 \pm 0.57}$ & $\underline{-0.83 \pm 0.38}$ & $OOM$ & $595.75 \pm 473.80$ & $463.09 \pm 516.32$ & $70.68 \pm 74.09$ & $2.62 \pm 3.67$ \\ 
&Descriptor Norm\tozero & $1676.71 \pm 786.88$ & $523.55 \pm 295.90$ & $OOM$ & $\underline{296.62 \pm 130.53}$ & $516.47 \pm 178.13$ & $419.98 \pm 232.31$ & $\mathbf{127.58 \pm 86.93}$ \\ 
&Median Extra Nodes\tozero & $377.50 \pm 24.71$ & $117.50 \pm 10.36$ & $OOM$ & $\mathbf{0.00 \pm 0.00}$ & $84.50 \pm 6.86$ & $120.50 \pm 10.29$ & $\underline{8.50 \pm 4.56}$ \\ 
&Median Missing Nodes\tozero & $\mathbf{0.00 \pm 0.00}$ & $\mathbf{0.00 \pm 0.00}$ & $OOM$ & $55.00 \pm 6.53$ & $\mathbf{0.00 \pm 0.00}$ & $\mathbf{0.00 \pm 0.00}$ & $\underline{0.00 \pm 1.05}$ \\ 
&Median Extra Edges\tozero & $146.00 \pm 12.86$ & $152.50 \pm 11.25$ & $OOM$ & $\underline{145.50 \pm 13.10}$ & $377.50 \pm 26.38$ & $\underline{145.50 \pm 13.57}$ & $\mathbf{9.00 \pm 4.31}$ \\ 
&Median Missing Edges\tozero & $\mathbf{0.00 \pm 0.00}$ & $\mathbf{0.00 \pm 0.00}$ & $OOM$ & $\mathbf{0.00 \pm 0.00}$ & $\mathbf{0.00 \pm 0.00}$ & $\mathbf{0.00 \pm 0.00}$ & $\underline{0.00 \pm 0.91}$ \\

\bottomrule
\bottomrule

\end{tabular}}

\end{table} 

\begin{table}[H] 
    \centering
    \caption{Comprehensive evaluation results—including edge, node, and structural metrics—for all benchmark models on \textbf{Bancor}. \tozero indicates that values closer to 0 are better. \higherbetter denotes that higher values are better, while \lowerbetter denotes that lower values are better. \textbf{Bold} indicates best, the second-best is \underline{underlined}. $OOM$ indicate out-of-memory.}
    \label{tab:bancor_results}
    \resizebox{1.00\linewidth}{!}{\begin{tabular}{l | c | c c c c c c | c}
\toprule
\toprule
& \textbf{Metric} & \textbf{HTGN} & \textbf{ROLAND} & \textbf{VGRNN} & \textbf{TGCN} & \textbf{GCLSTM} & \textbf{EvolveGCN} & \textbf{TopoGED} \\ \midrule
\multirow{11}{*}{\rotatebox{90}{\textbf{Edge Evaluation}}}&oo-bank Precision\higherbetter & $0.00 \pm 0.00$ & $0.00 \pm 0.00$ & $OOM$ & $0.00 \pm 0.00$ & $\underline{0.23 \pm 0.42}$ & $0.01 \pm 0.05$ & $\mathbf{0.25 \pm 0.11}$ \\
&oo-bank Recall\higherbetter & $0.00 \pm 0.00$ & $0.00 \pm 0.00$ & $OOM$ & $0.00 \pm 0.00$ & $\underline{0.01 \pm 0.01}$ & $0.00 \pm 0.00$ & $\mathbf{0.34 \pm 0.15}$ \\
&oo-bank F1\higherbetter & $0.00 \pm 0.00$ & $0.00 \pm 0.00$ & $OOM$ & $0.00 \pm 0.00$ & $\underline{0.01 \pm 0.03}$ & $0.00 \pm 0.00$ & $\mathbf{0.27 \pm 0.10}$ \\
&oo-nobank Precision\higherbetter & $0.00 \pm 0.00$ & $0.00 \pm 0.00$ & $OOM$ & $0.00 \pm 0.00$ & $\mathbf{0.01 \pm 0.05}$ & $0.00 \pm 0.00$ & $\mathbf{0.01 \pm 0.03}$ \\
&oo-nobank Recall\higherbetter & $0.00 \pm 0.00$ & $0.00 \pm 0.00$ & $OOM$ & $0.00 \pm 0.00$ & $0.00 \pm 0.00$ & $0.00 \pm 0.00$ & $\mathbf{0.00 \pm 0.02}$ \\
&oo-nobank F1\higherbetter & $0.00 \pm 0.00$ & $0.00 \pm 0.00$ & $OOM$ & $0.00 \pm 0.00$ & $0.00 \pm 0.01$ & $0.00 \pm 0.00$ & $\mathbf{0.01 \pm 0.03}$ \\
&Edge Precision\higherbetter & $0.00 \pm 0.00$ & $0.00 \pm 0.00$ & $OOM$ & $0.00 \pm 0.00$ & $0.00 \pm 0.00$ & $0.00 \pm 0.00$ & $\mathbf{0.15 \pm 0.08}$ \\
&Edge Recall\higherbetter & $0.00 \pm 0.00$ & $0.00 \pm 0.00$ & $OOM$ & $0.00 \pm 0.00$ & $0.00 \pm 0.00$ & $0.00 \pm 0.00$ & $\mathbf{0.14 \pm 0.07}$ \\
&Edge F1\higherbetter & $0.00 \pm 0.00$ & $0.00 \pm 0.00$ & $OOM$ & $0.00 \pm 0.00$ & $0.00 \pm 0.00$ & $0.00 \pm 0.00$ & $\mathbf{0.14 \pm 0.07}$ \\
&Num o-n Predicted\tozero & $-0.45 \pm 0.22$ & $-0.29 \pm 0.45$ & $OOM$ & $0.34 \pm 0.53$ & $0.59 \pm 0.51$ & $\underline{-0.11 \pm 0.40}$ & $\mathbf{0.04 \pm 0.35}$ \\
&Num n-n Predicted\tozero & $69.62 \pm 65.14$ & $64.20 \pm 59.69$ & $OOM$ & $\underline{55.66 \pm 52.53}$ & $107.34 \pm 97.87$ & $64.87 \pm 61.32$ & $\mathbf{0.56 \pm 1.80}$ \\

\midrule
\multirow{7}{*}{\rotatebox{90}{\textbf{Node Evaluation}}}&Precision Nodes\higherbetter & $0.00 \pm 0.00$ & $0.01 \pm 0.01$ & $OOM$ & $0.00 \pm 0.00$ & $\underline{0.03 \pm 0.01}$ & $\underline{0.03 \pm 0.02}$ & $\mathbf{0.58 \pm 0.09}$ \\
&Recall Nodes\higherbetter & $0.00 \pm 0.00$ & $0.01 \pm 0.00$ & $OOM$ & $0.00 \pm 0.00$ & $\underline{0.02 \pm 0.01}$ & $0.01 \pm 0.01$ & $\mathbf{0.57 \pm 0.10}$ \\
&F1 Nodes\higherbetter & $0.00 \pm 0.00$ & $0.01 \pm 0.00$ & $OOM$ & $0.00 \pm 0.00$ & $\underline{0.03 \pm 0.01}$ & $0.02 \pm 0.01$ & $\mathbf{0.57 \pm 0.07}$ \\
&Precision Old Nodes\higherbetter & $0.00 \pm 0.00$ & $\mathbf{0.62 \pm 0.44}$ & $OOM$ & $0.01 \pm 0.03$ & $\underline{0.51 \pm 0.10}$ & $0.34 \pm 0.16$ & $0.33 \pm 0.09$ \\
&Recall Old Nodes\higherbetter & $0.00 \pm 0.00$ & $0.01 \pm 0.01$ & $OOM$ & $0.00 \pm 0.00$ & $\underline{0.04 \pm 0.01}$ & $0.03 \pm 0.01$ & $\mathbf{0.32 \pm 0.08}$ \\
&F1 Old Nodes\higherbetter & $0.00 \pm 0.00$ & $0.02 \pm 0.01$ & $OOM$ & $0.00 \pm 0.01$ & $\underline{0.07 \pm 0.02}$ & $0.05 \pm 0.03$ & $\mathbf{0.32 \pm 0.07}$ \\
&New Nodes Predicted\higherbetter & $8.58 \pm 3.85$ & $0.48 \pm 0.61$ & $OOM$ & $\mathbf{-0.01 \pm 0.33}$ & $0.41 \pm 0.50$ & $0.27 \pm 0.47$ & $\underline{0.03 \pm 0.31}$ \\

\midrule
\multirow{11}{*}{\rotatebox{90}{\textbf{Structure Evaluation}}}&Avg Node Degree\tozero & $\underline{-0.39 \pm 0.05}$ & $3.18 \pm 1.22$ & $OOM$ & $4.44 \pm 0.87$ & $6.20 \pm 1.73$ & $3.69 \pm 1.31$ & $\mathbf{0.01 \pm 0.09}$ \\
&Unique Degree Count\tozero & $\underline{-0.63 \pm 0.07}$ & $1.31 \pm 0.59$ & $OOM$ & $1.83 \pm 0.66$ & $1.72 \pm 0.92$ & $1.63 \pm 0.65$ & $\mathbf{0.16 \pm 0.24}$ \\
&Degree Centrality\tozero & $\underline{-0.84 \pm 0.07}$ & $7.45 \pm 6.13$ & $OOM$ & $11.67 \pm 6.18$ & $12.28 \pm 9.08$ & $8.53 \pm 6.83$ & $\mathbf{0.08 \pm 0.46}$ \\
&Assortativity Coefficient\tozero & $-2.25 \pm 0.42$ & $1.39 \pm 0.57$ & $OOM$ & $\mathbf{0.94 \pm 0.42}$ & $1.81 \pm 0.70$ & $-2.27 \pm 0.74$ & $\underline{1.13 \pm 0.52}$ \\
&Clustering Coefficient\tozero & $\mathbf{-0.84 \pm 0.34}$ & $34.97 \pm 42.92$ & $OOM$ & $40.36 \pm 48.15$ & $44.31 \pm 50.13$ & $29.82 \pm 35.45$ & $\underline{2.63 \pm 4.87}$ \\
&Density\tozero & $\underline{-0.84 \pm 0.07}$ & $7.45 \pm 6.13$ & $OOM$ & $11.67 \pm 6.18$ & $12.28 \pm 9.08$ & $8.53 \pm 6.83$ & $\mathbf{0.08 \pm 0.46}$ \\
&Num Triangles\tozero & $\mathbf{-0.74 \pm 0.58}$ & $374.60 \pm 295.06$ & $OOM$ & $492.84 \pm 468.77$ & $1027.56 \pm 1080.19$ & $559.73 \pm 549.70$ & $\underline{5.15 \pm 6.79}$ \\
&Descriptor Norm\tozero & $1656.96 \pm 1283.31$ & $\underline{319.63 \pm 255.20}$ & $OOM$ & $365.52 \pm 308.79$ & $500.71 \pm 299.54$ & $374.58 \pm 338.75$ & $\mathbf{171.73 \pm 230.58}$ \\
&Median Extra Nodes\tozero & $380.00 \pm 20.37$ & $\mathbf{0.00 \pm 0.00}$ & $OOM$ & $\mathbf{0.00 \pm 0.00}$ & $\mathbf{0.00 \pm 0.00}$ & $\mathbf{0.00 \pm 0.00}$ & $\underline{0.00 \pm 2.21}$ \\
&Median Missing Nodes\tozero & $\mathbf{0.00 \pm 0.00}$ & $47.50 \pm 5.96$ & $OOM$ & $68.00 \pm 4.53$ & $45.50 \pm 4.88$ & $56.00 \pm 4.12$ & $\underline{1.50 \pm 1.74}$ \\
&Median Extra Edges\tozero & $154.00 \pm 9.87$ & $158.00 \pm 13.16$ & $OOM$ & $\underline{153.00 \pm 10.12}$ & $356.50 \pm 11.58$ & $160.00 \pm 9.92$ & $\mathbf{1.50 \pm 2.91}$ \\
&Median Missing Edges\tozero & $\mathbf{0.00 \pm 0.00}$ & $\mathbf{0.00 \pm 0.00}$ & $OOM$ & $\mathbf{0.00 \pm 0.00}$ & $\mathbf{0.00 \pm 0.00}$ & $\mathbf{0.00 \pm 0.00}$ & $\underline{0.00 \pm 1.76}$ \\

\bottomrule
\bottomrule

\end{tabular}}

\end{table} 

\begin{table}[H] 
    \centering
    \caption{Comprehensive evaluation results—including edge, node, and structural metrics—for all benchmark models on \textbf{Centra}. \tozero indicates that values closer to 0 are better. \higherbetter denotes that higher values are better, while \lowerbetter denotes that lower values are better. \textbf{Bold} indicates best, the second-best is \underline{underlined}.}
    \label{tab:centra_results}
    \resizebox{1.00\linewidth}{!}{\begin{tabular}{l | c | c c c c c c | c}
\toprule
\toprule
& \textbf{Metric} & \textbf{HTGN} & \textbf{ROLAND} & \textbf{VGRNN} & \textbf{TGCN} & \textbf{GCLSTM} & \textbf{EvolveGCN} & \textbf{TopoGED} \\ \midrule
\multirow{11}{*}{\rotatebox{90}{\textbf{Edge Evaluation}}}&oo-bank Precision\higherbetter & $0.00 \pm 0.00$ & $0.00 \pm 0.00$ & $0.00 \pm 0.00$ & $0.00 \pm 0.00$ & $\mathbf{0.12 \pm 0.22}$ & $0.05 \pm 0.22$ & $\underline{0.09 \pm 0.09}$ \\
&oo-bank Recall\higherbetter & $0.00 \pm 0.00$ & $0.00 \pm 0.00$ & $0.00 \pm 0.00$ & $0.00 \pm 0.00$ & $\underline{0.01 \pm 0.02}$ & $0.00 \pm 0.00$ & $\mathbf{0.18 \pm 0.21}$ \\
&oo-bank F1\higherbetter & $0.00 \pm 0.00$ & $0.00 \pm 0.00$ & $0.00 \pm 0.00$ & $0.00 \pm 0.00$ & $\underline{0.02 \pm 0.04}$ & $0.00 \pm 0.01$ & $\mathbf{0.11 \pm 0.11}$ \\
&oo-nobank Precision\higherbetter & $0.00 \pm 0.00$ & $0.00 \pm 0.00$ & $0.00 \pm 0.00$ & $0.00 \pm 0.00$ & $0.00 \pm 0.00$ & $0.00 \pm 0.00$ & $\mathbf{0.01 \pm 0.02}$ \\
&oo-nobank Recall\higherbetter & $0.00 \pm 0.00$ & $0.00 \pm 0.00$ & $0.00 \pm 0.00$ & $0.00 \pm 0.00$ & $0.00 \pm 0.00$ & $0.00 \pm 0.00$ & $\mathbf{0.01 \pm 0.02}$ \\
&oo-nobank F1\higherbetter & $0.00 \pm 0.00$ & $0.00 \pm 0.00$ & $0.00 \pm 0.00$ & $0.00 \pm 0.00$ & $0.00 \pm 0.00$ & $0.00 \pm 0.00$ & $\mathbf{0.01 \pm 0.01}$ \\
&Edge Precision\higherbetter & $0.00 \pm 0.00$ & $0.00 \pm 0.00$ & $0.00 \pm 0.00$ & $0.00 \pm 0.00$ & $0.00 \pm 0.00$ & $0.00 \pm 0.00$ & $\mathbf{0.07 \pm 0.06}$ \\
&Edge Recall\higherbetter & $0.00 \pm 0.00$ & $0.00 \pm 0.00$ & $0.00 \pm 0.00$ & $0.00 \pm 0.00$ & $0.00 \pm 0.00$ & $0.00 \pm 0.00$ & $\mathbf{0.08 \pm 0.07}$ \\
&Edge F1\higherbetter & $0.00 \pm 0.00$ & $0.00 \pm 0.00$ & $0.00 \pm 0.00$ & $0.00 \pm 0.00$ & $0.00 \pm 0.00$ & $0.00 \pm 0.00$ & $\mathbf{0.07 \pm 0.06}$ \\
&Num o-n Predicted\tozero & $\underline{-0.19 \pm 0.64}$ & $\mathbf{0.05 \pm 1.48}$ & $0.64 \pm 2.27$ & $-0.87 \pm 0.18$ & $0.52 \pm 1.26$ & $-0.26 \pm 0.66$ & $0.46 \pm 0.92$ \\
&Num n-n Predicted\tozero & $419.16 \pm 554.55$ & $500.92 \pm 743.93$ & $544.65 \pm 652.79$ & $\underline{153.58 \pm 101.27}$ & $473.27 \pm 499.76$ & $423.32 \pm 569.51$ & $\mathbf{0.74 \pm 1.74}$ \\

\midrule
\multirow{7}{*}{\rotatebox{90}{\textbf{Node Evaluation}}}&Precision Nodes\higherbetter & $0.00 \pm 0.00$ & $0.00 \pm 0.00$ & $0.01 \pm 0.01$ & $0.00 \pm 0.00$ & $\underline{0.02 \pm 0.01}$ & $0.01 \pm 0.01$ & $\mathbf{0.47 \pm 0.15}$ \\
&Recall Nodes\higherbetter & $0.00 \pm 0.00$ & $\underline{0.01 \pm 0.01}$ & $\underline{0.01 \pm 0.01}$ & $0.00 \pm 0.00$ & $\underline{0.01 \pm 0.01}$ & $\underline{0.01 \pm 0.01}$ & $\mathbf{0.56 \pm 0.22}$ \\
&F1 Nodes\higherbetter & $0.00 \pm 0.00$ & ${0.01 \pm 0.00}$ & $\underline{0.01 \pm 0.01}$ & $0.00 \pm 0.00$ & $\underline{0.01 \pm 0.01}$ & $\underline{0.01 \pm 0.01}$ & $\mathbf{0.47 \pm 0.14}$ \\
&Precision Old Nodes\higherbetter & $0.00 \pm 0.00$ & $0.05 \pm 0.04$ & $\mathbf{0.23 \pm 0.33}$ & $0.00 \pm 0.00$ & $0.17 \pm 0.08$ & $\underline{0.22 \pm 0.29}$ & $\mathbf{0.23 \pm 0.13}$ \\
&Recall Old Nodes\higherbetter & $0.00 \pm 0.01$ & $0.02 \pm 0.02$ & $0.01 \pm 0.01$ & $0.00 \pm 0.00$ & $\underline{0.03 \pm 0.02}$ & $0.02 \pm 0.02$ & $\mathbf{0.26 \pm 0.13}$ \\
&F1 Old Nodes\higherbetter & $0.00 \pm 0.00$ & $0.02 \pm 0.02$ & $0.02 \pm 0.02$ & $0.00 \pm 0.00$ & $\underline{0.05 \pm 0.03}$ & $0.03 \pm 0.04$ & $\mathbf{0.21 \pm 0.11}$ \\
&New Nodes Predicted\higherbetter & $10.35 \pm 6.75$ & $5.31 \pm 6.29$ & $1.75 \pm 5.46$ & $2.89 \pm 2.64$ & $1.31 \pm 2.09$ & $\underline{0.96 \pm 3.44}$ & $\mathbf{0.35 \pm 0.76}$ \\

\midrule
\multirow{11}{*}{\rotatebox{90}{\textbf{Structure Evaluation}}}&Avg Node Degree\tozero & $-0.39 \pm 0.17$ & $\underline{0.30 \pm 0.55}$ & $6.05 \pm 5.77$ & $1.49 \pm 1.64$ & $4.89 \pm 2.68$ & $4.58 \pm 3.51$ & $\mathbf{0.03 \pm 0.12}$ \\
&Unique Degree Count\tozero & $-0.46 \pm 0.48$ & $0.43 \pm 1.40$ & $0.53 \pm 0.74$ & $\mathbf{-0.05 \pm 0.73}$ & $2.15 \pm 1.72$ & $2.14 \pm 1.23$ & $\underline{0.06 \pm 0.30}$ \\
&Degree Centrality\tozero & $-0.79 \pm 0.30$ & $\mathbf{0.52 \pm 5.05}$ & $21.35 \pm 45.22$ & $3.04 \pm 4.25$ & $20.43 \pm 39.18$ & $24.73 \pm 56.03$ & $\underline{0.56 \pm 3.55}$ \\
&Assortativity Coefficient\tozero & $-1.58 \pm 0.61$ & $0.25 \pm 2.47$ & $0.26 \pm 1.95$ & $0.46 \pm 3.29$ & $\underline{0.07 \pm 2.20}$ & $-0.94 \pm 0.63$ & $\mathbf{-0.04 \pm 1.53}$ \\
&Clustering Coefficient\tozero & $\underline{4.89 \pm 18.04}$ & $14.82 \pm 63.29$ & $56.31 \pm 135.01$ & $6.74 \pm 23.95$ & $46.57 \pm 103.10$ & $35.26 \pm 83.73$ & $\mathbf{2.13 \pm 4.24}$ \\
&Density\tozero & $-0.79 \pm 0.30$ & $\mathbf{0.52 \pm 5.05}$ & $21.35 \pm 45.22$ & $3.04 \pm 4.25$ & $20.43 \pm 39.18$ & $24.73 \pm 56.03$ & $\underline{0.56 \pm 3.55}$ \\
&Num Triangles\tozero & $138.66 \pm 722.69$ & $486.66 \pm 1408.43$ & $1843.29 \pm 4451.19$ & $\underline{126.88 \pm 152.34}$ & $2946.97 \pm 8072.72$ & $1808.73 \pm 3466.10$ & $\mathbf{16.99 \pm 43.03}$ \\
&Descriptor Norm\tozero & $3578.26 \pm 4205.39$ & $1862.97 \pm 2565.60$ & $2326.09 \pm 3778.83$ & $\mathbf{650.07 \pm 1494.33}$ & $1758.78 \pm 2467.73$ & $1995.73 \pm 3215.58$ & $\underline{1042.17 \pm 1701.40}$ \\
&Median Extra Nodes\tozero & $411.00 \pm 73.46$ & $168.00 \pm 34.40$ & $\mathbf{0.00 \pm 1.30}$ & ${0.00 \pm 31.00}$ & $\underline{0.00 \pm 1.77}$ & $\mathbf{0.00 \pm 0.00}$ & ${19.50 \pm 11.98}$ \\
&Median Missing Nodes\tozero & $\mathbf{0.00 \pm 0.00}$ & $\mathbf{0.00 \pm 0.00}$ & $49.50 \pm 29.18$ & ${1.00 \pm 7.40}$ & $21.50 \pm 23.48$ & $58.00 \pm 10.03$ & $\underline{0.00 \pm 1.68}$ \\
&Median Extra Edges\tozero & $154.50 \pm 40.46$ & $186.50 \pm 53.45$ & $361.50 \pm 66.36$ & $\underline{115.00 \pm 18.19}$ & $366.50 \pm 49.04$ & $151.50 \pm 41.55$ & $\mathbf{26.50 \pm 19.25}$ \\
&Median Missing Edges\tozero & $\mathbf{0.00 \pm 0.00}$ & $\mathbf{0.00 \pm 0.00}$ & $\mathbf{0.00 \pm 0.00}$ & ${0.00 \pm 2.12}$ & $\mathbf{0.00 \pm 0.00}$ & $\mathbf{0.00 \pm 0.00}$ & $\underline{0.00 \pm 1.06}$ \\

\bottomrule
\bottomrule

\end{tabular}}

\end{table} 

\begin{table}[H] 
    \centering
    \caption{Comprehensive evaluation results—including edge, node, and structural metrics—for all benchmark models on \textbf{Cindicator}. \tozero indicates that values closer to 0 are better. \higherbetter denotes that higher values are better, while \lowerbetter denotes that lower values are better. \textbf{Bold} indicates best, the second-best is \underline{underlined}.}
    \label{tab:cindicator_results}
    \resizebox{1.00\linewidth}{!}{\begin{tabular}{l | c | c c c c c c | c}
\toprule
\toprule
& \textbf{Metric} & \textbf{HTGN} & \textbf{ROLAND} & \textbf{VGRNN} & \textbf{TGCN} & \textbf{GCLSTM} & \textbf{EvolveGCN} & \textbf{TopoGED} \\ \midrule
\multirow{11}{*}{\rotatebox{90}{\textbf{Edge Evaluation}}}&oo-bank Precision\higherbetter & $0.00 \pm 0.00$ & $\mathbf{0.15 \pm 0.36}$ & $0.06 \pm 0.24$ & $0.00 \pm 0.00$ & $\mathbf{0.15 \pm 0.36}$ & $0.00 \pm 0.00$ & $\underline{0.11 \pm 0.07}$ \\
&oo-bank Recall\higherbetter & $0.00 \pm 0.00$ & $0.00 \pm 0.01$ & $0.00 \pm 0.01$ & $0.00 \pm 0.00$ & $0.00 \pm 0.01$ & $0.00 \pm 0.00$ & $\mathbf{0.18 \pm 0.11}$ \\
&oo-bank F1\higherbetter & $0.00 \pm 0.00$ & $\underline{0.01 \pm 0.02}$ & $0.00 \pm 0.01$ & $0.00 \pm 0.00$ & $\underline{0.01 \pm 0.02}$ & $0.00 \pm 0.00$ & $\mathbf{0.13 \pm 0.07}$ \\
&oo-nobank Precision\higherbetter & $0.00 \pm 0.00$ & $0.00 \pm 0.01$ & $0.00 \pm 0.00$ & $0.00 \pm 0.00$ & $\underline{0.01 \pm 0.06}$ & $0.00 \pm 0.00$ & $\mathbf{0.02 \pm 0.03}$ \\
&oo-nobank Recall\higherbetter & $0.00 \pm 0.00$ & $0.00 \pm 0.00$ & $0.00 \pm 0.00$ & $0.00 \pm 0.00$ & $0.00 \pm 0.00$ & $0.00 \pm 0.00$ & $\mathbf{0.01 \pm 0.02}$ \\
&oo-nobank F1\higherbetter & $0.00 \pm 0.00$ & $0.00 \pm 0.00$ & $0.00 \pm 0.00$ & $0.00 \pm 0.00$ & $0.00 \pm 0.01$ & $0.00 \pm 0.00$ & $\mathbf{0.01 \pm 0.02}$ \\
&Edge Precision\higherbetter & $0.00 \pm 0.00$ & $0.00 \pm 0.00$ & $0.00 \pm 0.00$ & $0.00 \pm 0.00$ & $0.00 \pm 0.00$ & $0.00 \pm 0.00$ & $\mathbf{0.12 \pm 0.07}$ \\
&Edge Recall\higherbetter & $0.00 \pm 0.00$ & $0.00 \pm 0.00$ & $0.00 \pm 0.00$ & $0.00 \pm 0.00$ & $0.00 \pm 0.00$ & $0.00 \pm 0.00$ & $\mathbf{0.13 \pm 0.07}$ \\
&Edge F1\higherbetter & $0.00 \pm 0.00$ & $0.00 \pm 0.00$ & $0.00 \pm 0.00$ & $0.00 \pm 0.00$ & $0.00 \pm 0.00$ & $0.00 \pm 0.00$ & $\mathbf{0.13 \pm 0.07}$ \\
&Num o-n Predicted\tozero & $-0.70 \pm 0.08$ & $3.67 \pm 1.44$ & $-0.51 \pm 0.42$ & $-1.00 \pm 0.00$ & $\underline{0.22 \pm 0.25}$ & $-0.70 \pm 0.17$ & $\mathbf{0.05 \pm 0.26}$ \\
&Num n-n Predicted\tozero & $278.15 \pm 150.62$ & $\underline{11.05 \pm 11.11}$ & $268.33 \pm 146.33$ & $293.46 \pm 159.21$ & $228.10 \pm 122.24$ & $276.82 \pm 154.10$ & $\mathbf{0.52 \pm 1.58}$ \\

\midrule
\multirow{7}{*}{\rotatebox{90}{\textbf{Node Evaluation}}}&Precision Nodes\higherbetter & $0.00 \pm 0.00$ & $0.00 \pm 0.00$ & $0.01 \pm 0.01$ & $0.00 \pm 0.00$ & $\underline{0.03 \pm 0.01}$ & $0.01 \pm 0.00$ & $\mathbf{0.51 \pm 0.06}$ \\
&Recall Nodes\higherbetter & $0.00 \pm 0.00$ & $\underline{0.01 \pm 0.00}$ & $\underline{0.01 \pm 0.01}$ & $0.00 \pm 0.00$ & $\underline{0.01 \pm 0.00}$ & $\underline{0.01 \pm 0.01}$ & $\mathbf{0.52 \pm 0.06}$ \\
&F1 Nodes\higherbetter & $0.00 \pm 0.00$ & $0.01 \pm 0.00$ & $0.01 \pm 0.01$ & $0.00 \pm 0.00$ & $\underline{0.02 \pm 0.01}$ & $0.01 \pm 0.00$ & $\mathbf{0.51 \pm 0.04}$ \\
&Precision Old Nodes\higherbetter & $0.00 \pm 0.00$ & $0.17 \pm 0.10$ & $\underline{0.39 \pm 0.33}$ & $0.00 \pm 0.00$ & $\mathbf{0.55 \pm 0.16}$ & $0.20 \pm 0.09$ & $0.20 \pm 0.05$ \\
&Recall Old Nodes\higherbetter & $0.00 \pm 0.00$ & $\underline{0.02 \pm 0.01}$ & $0.01 \pm 0.01$ & $0.00 \pm 0.00$ & $\underline{0.02 \pm 0.01}$ & $\underline{0.02 \pm 0.01}$ & $\mathbf{0.20 \pm 0.04}$ \\
&F1 Old Nodes\higherbetter & $0.00 \pm 0.00$ & $0.03 \pm 0.01$ & $0.02 \pm 0.02$ & $0.00 \pm 0.00$ & $\underline{0.04 \pm 0.01}$ & $\underline{0.04 \pm 0.02}$ & $\mathbf{0.20 \pm 0.04}$ \\
&New Nodes Predicted\higherbetter & $8.54 \pm 1.90$ & $4.60 \pm 1.63$ & $\mathbf{0.00 \pm 0.41}$ & $0.22 \pm 0.22$ & $-0.21 \pm 0.14$ & $3.82 \pm 1.35$ & $\underline{0.04 \pm 0.23}$ \\

\midrule
\multirow{11}{*}{\rotatebox{90}{\textbf{Structure Evaluation}}}&Avg Node Degree\tozero & $-0.39 \pm 0.03$ & $\underline{0.05 \pm 0.07}$ & $5.81 \pm 3.09$ & $3.91 \pm 0.55$ & $6.43 \pm 0.52$ & $0.24 \pm 0.27$ & $\mathbf{0.00 \pm 0.06}$ \\
&Unique Degree Count\tozero & $-0.45 \pm 0.11$ & $\mathbf{-0.06 \pm 0.24}$ & $0.45 \pm 0.56$ & $2.33 \pm 0.70$ & $2.30 \pm 0.66$ & $1.18 \pm 0.55$ & $\underline{0.17 \pm 0.21}$ \\
&Degree Centrality\tozero & $-0.85 \pm 0.03$ & $-0.54 \pm 0.14$ & $21.57 \pm 22.64$ & $8.64 \pm 1.77$ & $20.63 \pm 3.97$ & $\underline{-0.35 \pm 0.35}$ & $\mathbf{-0.00 \pm 0.19}$ \\
&Assortativity Coefficient\tozero & $-1.95 \pm 0.26$ & $1.03 \pm 0.31$ & $0.33 \pm 0.49$ & $0.45 \pm 0.23$ & $0.53 \pm 0.24$ & $\mathbf{-0.13 \pm 0.35}$ & $\underline{0.32 \pm 0.23}$ \\
&Clustering Coefficient\tozero & $6.44 \pm 22.00$ & $\mathbf{2.41 \pm 9.06}$ & $68.92 \pm 217.82$ & $64.29 \pm 184.65$ & $66.34 \pm 198.51$ & $5.82 \pm 25.51$ & $\underline{3.44 \pm 11.07}$ \\
&Density\tozero & $-0.85 \pm 0.03$ & $-0.54 \pm 0.14$ & $21.57 \pm 22.64$ & $8.64 \pm 1.77$ & $20.63 \pm 3.97$ & $\underline{-0.35 \pm 0.35}$ & $\mathbf{-0.00 \pm 0.19}$ \\
&Num Triangles\tozero & $\underline{16.02 \pm 15.07}$ & $18.88 \pm 17.75$ & $1287.65 \pm 1062.84$ & $911.88 \pm 726.83$ & $1130.89 \pm 770.31$ & $74.73 \pm 242.41$ & $\mathbf{7.44 \pm 8.97}$ \\
&Descriptor Norm\tozero & $1599.32 \pm 396.49$ & $848.61 \pm 387.64$ & $503.62 \pm 232.09$ & $\underline{379.57 \pm 100.96}$ & $437.60 \pm 89.09$ & $555.81 \pm 203.93$ & $\mathbf{133.74 \pm 56.00}$ \\
&Median Extra Nodes\tozero & $490.50 \pm 18.93$ & $213.50 \pm 22.82$ & $\mathbf{0.00 \pm 0.00}$ & $\mathbf{0.00 \pm 0.00}$ & $\mathbf{0.00 \pm 0.00}$ & $156.00 \pm 8.67$ & $\underline{8.00 \pm 6.02}$ \\
&Median Missing Nodes\tozero & $\mathbf{0.00 \pm 0.00}$ & $\mathbf{0.00 \pm 0.00}$ & $93.00 \pm 10.35$ & ${74.00 \pm 5.71}$ & $103.00 \pm 6.58$ & $\mathbf{0.00 \pm 0.00}$ & $\underline{0.00 \pm 3.62}$ \\
&Median Extra Edges\tozero & $226.50 \pm 17.21$ & $238.50 \pm 20.94$ & $236.50 \pm 17.67$ & $\underline{223.00 \pm 17.48}$ & $234.50 \pm 15.77$ & ${223.00 \pm 18.10}$ & $\mathbf{4.00 \pm 6.47}$ \\
&Median Missing Edges\tozero & $\mathbf{0.00 \pm 0.00}$ & $\mathbf{0.00 \pm 0.00}$ & $\mathbf{0.00 \pm 0.00}$ & $\mathbf{0.00 \pm 0.00}$ & $\mathbf{0.00 \pm 0.00}$ & $\mathbf{0.00 \pm 0.00}$ & $\underline{0.00 \pm 5.21}$ \\

\bottomrule
\bottomrule

\end{tabular}}

\end{table} 

\begin{table}[H] 
    \centering
    \caption{Comprehensive evaluation results—including edge, node, and structural metrics—for all benchmark models on \textbf{Coindash}. \tozero indicates that values closer to 0 are better. \higherbetter denotes that higher values are better, while \lowerbetter denotes that lower values are better. \textbf{Bold} indicates best, the second-best is \underline{underlined}.}
    \label{tab:coindash_results}
    \resizebox{1.00\linewidth}{!}{\begin{tabular}{l | c | c c c c c c | c}
\toprule
\toprule
& \textbf{Metric} & \textbf{HTGN} & \textbf{ROLAND} & \textbf{VGRNN} & \textbf{TGCN} & \textbf{GCLSTM} & \textbf{EvolveGCN} & \textbf{TopoGED} \\ \midrule
\multirow{11}{*}{\rotatebox{90}{\textbf{Edge Evaluation}}}&oo-bank Precision\higherbetter & $0.00 \pm 0.00$ & $0.00 \pm 0.00$ & $0.00 \pm 0.00$ & $0.00 \pm 0.00$ & $\underline{0.05 \pm 0.15}$ & $0.00 \pm 0.00$ & $\mathbf{0.20 \pm 0.10}$ \\
&oo-bank Recall\higherbetter & $0.00 \pm 0.00$ & $0.00 \pm 0.00$ & $0.00 \pm 0.00$ & $0.00 \pm 0.00$ & $\underline{0.00 \pm 0.01}$ & $0.00 \pm 0.00$ & $\mathbf{0.28 \pm 0.14}$ \\
&oo-bank F1\higherbetter & $0.00 \pm 0.00$ & $0.00 \pm 0.00$ & $0.00 \pm 0.00$ & $0.00 \pm 0.00$ & $\underline{0.01 \pm 0.02}$ & $0.00 \pm 0.00$ & $\mathbf{0.22 \pm 0.10}$ \\
&oo-nobank Precision\higherbetter & $0.00 \pm 0.00$ & $0.00 \pm 0.00$ & $0.00 \pm 0.00$ & $0.00 \pm 0.00$ & $0.00 \pm 0.00$ & $0.00 \pm 0.00$ & $\mathbf{0.02 \pm 0.05}$ \\
&oo-nobank Recall\higherbetter & $0.00 \pm 0.00$ & $0.00 \pm 0.00$ & $0.00 \pm 0.00$ & $0.00 \pm 0.00$ & $0.00 \pm 0.00$ & $0.00 \pm 0.00$ & $\mathbf{0.01 \pm 0.04}$ \\
&oo-nobank F1\higherbetter & $0.00 \pm 0.00$ & $0.00 \pm 0.00$ & $0.00 \pm 0.00$ & $0.00 \pm 0.00$ & $0.00 \pm 0.00$ & $0.00 \pm 0.00$ & $\mathbf{0.02 \pm 0.04}$ \\
&Edge Precision\higherbetter & $0.00 \pm 0.00$ & $0.00 \pm 0.00$ & $0.00 \pm 0.00$ & $0.00 \pm 0.00$ & $0.00 \pm 0.00$ & $0.00 \pm 0.00$ & $\mathbf{0.13 \pm 0.06}$ \\
&Edge Recall\higherbetter & $0.00 \pm 0.00$ & $0.00 \pm 0.00$ & $0.00 \pm 0.00$ & $0.00 \pm 0.00$ & $0.00 \pm 0.00$ & $0.00 \pm 0.00$ & $\mathbf{0.13 \pm 0.07}$ \\
&Edge F1\higherbetter & $0.00 \pm 0.00$ & $0.00 \pm 0.00$ & $0.00 \pm 0.00$ & $0.00 \pm 0.00$ & $0.00 \pm 0.00$ & $0.00 \pm 0.00$ & $\mathbf{0.13 \pm 0.06}$ \\
&Num o-n Predicted\tozero & $0.19 \pm 0.78$ & $-0.41 \pm 0.32$ & $-0.34 \pm 1.58$ & $-0.93 \pm 0.40$ & $0.58 \pm 2.12$ & $\underline{-0.28 \pm 0.69}$ & $\mathbf{0.13 \pm 0.71}$ \\
&Num n-n Predicted\tozero & $\underline{179.27 \pm 383.82}$ & $187.30 \pm 95.05$ & $240.66 \pm 413.80$ & $203.72 \pm 402.28$ & $299.16 \pm 368.46$ & $212.74 \pm 530.69$ & $\mathbf{0.26 \pm 1.29}$ \\

\midrule
\multirow{7}{*}{\rotatebox{90}{\textbf{Node Evaluation}}}&Precision Nodes\higherbetter & $0.00 \pm 0.00$ & $0.01 \pm 0.01$ & $0.01 \pm 0.02$ & $0.00 \pm 0.00$ & $\underline{0.03 \pm 0.01}$ & $0.02 \pm 0.02$ & $\mathbf{0.49 \pm 0.13}$ \\
&Recall Nodes\higherbetter & $0.00 \pm 0.00$ & $0.02 \pm 0.02$ & $0.01 \pm 0.01$ & $0.00 \pm 0.00$ & $\underline{0.04 \pm 0.02}$ & $0.01 \pm 0.01$ & $\mathbf{0.49 \pm 0.14}$ \\
&F1 Nodes\higherbetter & $0.00 \pm 0.00$ & $0.01 \pm 0.01$ & $0.01 \pm 0.01$ & $0.00 \pm 0.00$ & $\underline{0.03 \pm 0.01}$ & $0.01 \pm 0.01$ & $\mathbf{0.47 \pm 0.12}$ \\
&Precision Old Nodes\higherbetter & $0.00 \pm 0.00$ & $0.13 \pm 0.09$ & $0.28 \pm 0.43$ & $0.00 \pm 0.00$ & $\mathbf{0.77 \pm 0.31}$ & $\underline{0.35 \pm 0.40}$ & $\underline{0.35 \pm 0.12}$ \\
&Recall Old Nodes\higherbetter & $0.00 \pm 0.00$ & $0.03 \pm 0.03$ & $0.01 \pm 0.01$ & $0.00 \pm 0.00$ & $\underline{0.05 \pm 0.03}$ & $0.02 \pm 0.02$ & $\mathbf{0.34 \pm 0.12}$ \\
&F1 Old Nodes\higherbetter & $0.00 \pm 0.00$ & $0.05 \pm 0.04$ & $0.02 \pm 0.03$ & $0.00 \pm 0.00$ & $\underline{0.10 \pm 0.05}$ & $0.03 \pm 0.03$ & $\mathbf{0.33 \pm 0.11}$ \\
&New Nodes Predicted\higherbetter & $17.20 \pm 13.86$ & $10.08 \pm 6.57$ & $\underline{1.25 \pm 2.07}$ & $1.49 \pm 1.20$ & $4.21 \pm 2.90$ & $3.06 \pm 12.39$ & $\mathbf{0.09 \pm 0.63}$ \\

\midrule
\multirow{11}{*}{\rotatebox{90}{\textbf{Structure Evaluation}}}&Avg Node Degree\tozero & $-0.47 \pm 0.06$ & $\underline{0.18 \pm 0.17}$ & $5.83 \pm 2.78$ & $3.19 \pm 1.83$ & $3.25 \pm 3.14$ & $3.32 \pm 1.33$ & $\mathbf{0.01 \pm 0.10}$ \\
&Unique Degree Count\tozero & $-0.75 \pm 0.28$ & $\mathbf{-0.14 \pm 0.35}$ & $0.36 \pm 0.46$ & $1.64 \pm 0.82$ & $0.39 \pm 0.91$ & $1.49 \pm 0.76$ & $\underline{0.16 \pm 0.29}$ \\
&Degree Centrality\tozero & $-0.87 \pm 0.09$ & $\underline{-0.44 \pm 0.76}$ & $18.00 \pm 24.20$ & $8.09 \pm 16.17$ & $5.51 \pm 17.65$ & $9.74 \pm 17.48$ & $\mathbf{0.21 \pm 1.07}$ \\
&Assortativity Coefficient\tozero & $-1.95 \pm 0.94$ & $1.36 \pm 0.83$ & $\underline{0.72 \pm 0.87}$ & $\mathbf{0.14 \pm 0.52}$ & $1.47 \pm 0.91$ & $0.28 \pm 0.50$ & $0.47 \pm 0.57$ \\
&Clustering Coefficient\tozero & $\mathbf{0.29 \pm 1.68}$ & $\underline{3.31 \pm 12.00}$ & $29.96 \pm 117.24$ & $25.97 \pm 101.14$ & $29.59 \pm 114.28$ & $22.53 \pm 95.11$ & $3.39 \pm 14.63$ \\
&Density\tozero & $-0.87 \pm 0.09$ & $\underline{-0.44 \pm 0.76}$ & $18.00 \pm 24.20$ & $8.09 \pm 16.17$ & $5.51 \pm 17.65$ & $9.74 \pm 17.48$ & $\mathbf{0.21 \pm 1.07}$ \\
&Num Triangles\tozero & $\mathbf{4.49 \pm 25.54}$ & $26.28 \pm 25.71$ & $830.21 \pm 976.67$ & $486.47 \pm 1005.20$ & $703.41 \pm 1325.11$ & $368.13 \pm 281.53$ & $\underline{7.10 \pm 8.94}$ \\
&Descriptor Norm\tozero & $1216.42 \pm 1642.06$ & $499.46 \pm 426.32$ & $405.05 \pm 624.87$ & $\underline{308.18 \pm 658.46}$ & $426.25 \pm 632.52$ & $488.05 \pm 1625.19$ & $\mathbf{186.88 \pm 532.55}$ \\
&Median Extra Nodes\tozero & $235.50 \pm 26.04$ & $114.50 \pm 8.74$ & $\mathbf{0.00 \pm 0.00}$ & $\mathbf{0.00 \pm 0.00}$ & $30.00 \pm 4.15$ & $\mathbf{0.00 \pm 0.00}$ & $\underline{0.50 \pm 1.95}$ \\
&Median Missing Nodes\tozero & $\mathbf{0.00 \pm 0.00}$ & $\mathbf{0.00 \pm 0.00}$ & $25.50 \pm 5.60$ & ${21.00 \pm 3.98}$ & $\mathbf{0.00 \pm 0.00}$ & $24.50 \pm 4.13$ & $\underline{0.00 \pm 1.79}$ \\
&Median Extra Edges\tozero & $86.00 \pm 10.55$ & $140.00 \pm 10.59$ & $151.00 \pm 21.81$ & $\underline{85.00 \pm 9.91}$ & $227.00 \pm 24.04$ & $85.50 \pm 9.77$ & $\mathbf{1.50 \pm 2.61}$ \\
&Median Missing Edges\tozero & $\mathbf{0.00 \pm 0.00}$ & $\mathbf{0.00 \pm 0.00}$ & $\mathbf{0.00 \pm 0.00}$ & $\mathbf{0.00 \pm 0.00}$ & $\mathbf{0.00 \pm 0.00}$ & $\mathbf{0.00 \pm 0.00}$ & $\underline{0.00 \pm 1.15}$ \\

\bottomrule
\bottomrule

\end{tabular}}

\end{table} 

\begin{table}[H] 
    \centering
    \caption{Comprehensive evaluation results—including edge, node, and structural metrics—for all benchmark models on \textbf{DGD}. \tozero indicates that values closer to 0 are better. \higherbetter denotes that higher values are better, while \lowerbetter denotes that lower values are better. \textbf{Bold} indicates best, the second-best is \underline{underlined}. $OOM$ indicate out-of-memory.}
    \label{tab:dgd_results}
    \resizebox{1.00\linewidth}{!}{\begin{tabular}{l | c | c c c c c c | c}
\toprule
\toprule
& \textbf{Metric} & \textbf{HTGN} & \textbf{ROLAND} & \textbf{VGRNN} & \textbf{TGCN} & \textbf{GCLSTM} & \textbf{EvolveGCN} & \textbf{TopoGED} \\ \midrule
\multirow{11}{*}{\rotatebox{90}{\textbf{Edge Evaluation}}}&oo-bank Precision\higherbetter & $0.00 \pm 0.00$ & $\underline{0.01 \pm 0.10}$ & $OOM$ & $0.00 \pm 0.00$ & $0.00 \pm 0.03$ & $0.00 \pm 0.00$ & $\mathbf{0.14 \pm 0.09}$ \\
&oo-bank Recall\higherbetter & $0.00 \pm 0.00$ & $0.00 \pm 0.00$ & $OOM$ & $0.00 \pm 0.00$ & $0.00 \pm 0.00$ & $0.00 \pm 0.00$ & $\mathbf{0.17 \pm 0.11}$ \\
&oo-bank F1\higherbetter & $0.00 \pm 0.00$ & $\underline{0.01 \pm 0.01}$ & $OOM$ & $0.00 \pm 0.00$ & $\underline{0.01 \pm 0.01}$ & $0.00 \pm 0.00$ & $\mathbf{0.15 \pm 0.08}$ \\
&oo-nobank Precision\higherbetter & $0.00 \pm 0.00$ & $0.00 \pm 0.00$ & $OOM$ & $0.00 \pm 0.00$ & $0.00 \pm 0.00$ & $0.00 \pm 0.00$ & $\mathbf{0.01 \pm 0.03}$ \\
&oo-nobank Recall\higherbetter & $0.00 \pm 0.00$ & $0.00 \pm 0.00$ & $OOM$ & $0.00 \pm 0.00$ & $0.00 \pm 0.00$ & $0.00 \pm 0.00$ & $\mathbf{0.01 \pm 0.02}$ \\
&oo-nobank F1\higherbetter & $0.00 \pm 0.00$ & $0.00 \pm 0.00$ & $OOM$ & $0.00 \pm 0.00$ & $0.00 \pm 0.00$ & $0.00 \pm 0.00$ & $\mathbf{0.01 \pm 0.02}$ \\
&Edge Precision\higherbetter & $0.00 \pm 0.00$ & $0.00 \pm 0.00$ & $OOM$ & $0.00 \pm 0.00$ & $0.00 \pm 0.00$ & $0.00 \pm 0.00$ & $\mathbf{0.08 \pm 0.05}$ \\
&Edge Recall\higherbetter & $0.00 \pm 0.00$ & $0.00 \pm 0.00$ & $OOM$ & $0.00 \pm 0.00$ & $0.00 \pm 0.00$ & $0.00 \pm 0.00$ & $\mathbf{0.08 \pm 0.05}$ \\
&Edge F1\higherbetter & $0.00 \pm 0.00$ & $0.00 \pm 0.00$ & $OOM$ & $0.00 \pm 0.00$ & $0.00 \pm 0.00$ & $0.00 \pm 0.00$ & $\mathbf{0.08 \pm 0.05}$ \\
&Num o-n Predicted\tozero & $-0.71 \pm 0.13$ & $-0.65 \pm 0.43$ & $OOM$ & $-0.61 \pm 0.37$ & $\mathbf{0.01 \pm 1.20}$ & $-0.63 \pm 0.23$ & $\underline{0.17 \pm 0.49}$ \\
&Num n-n Predicted\tozero & $186.38 \pm 164.41$ & $188.90 \pm 166.41$ & $OOM$ & $184.65 \pm 168.51$ & $217.60 \pm 170.96$ & $\underline{183.59 \pm 164.55}$ & $\mathbf{0.49 \pm 1.55}$ \\

\midrule
\multirow{7}{*}{\rotatebox{90}{\textbf{Node Evaluation}}}&Precision Nodes\higherbetter & $0.00 \pm 0.00$ & $\underline{0.02 \pm 0.02}$ & $OOM$ & $0.00 \pm 0.00$ & $\underline{0.02 \pm 0.01}$ & $0.00 \pm 0.00$ & $\mathbf{0.53 \pm 0.11}$ \\
&Recall Nodes\higherbetter & $0.00 \pm 0.00$ & $0.01 \pm 0.01$ & $OOM$ & $0.00 \pm 0.00$ & $\underline{0.02 \pm 0.01}$ & $0.01 \pm 0.01$ & $\mathbf{0.56 \pm 0.13}$ \\
&F1 Nodes\higherbetter & $0.00 \pm 0.00$ & $0.01 \pm 0.01$ & $OOM$ & $0.00 \pm 0.00$ & $\underline{0.02 \pm 0.01}$ & $0.01 \pm 0.01$ & $\mathbf{0.53 \pm 0.09}$ \\
&Precision Old Nodes\higherbetter & $0.02 \pm 0.05$ & $\mathbf{0.42 \pm 0.38}$ & $OOM$ & $0.00 \pm 0.00$ & $\underline{0.37 \pm 0.19}$ & $0.15 \pm 0.13$ & $0.33 \pm 0.09$ \\
&Recall Old Nodes\higherbetter & $0.00 \pm 0.01$ & $0.01 \pm 0.01$ & $OOM$ & $0.00 \pm 0.00$ & $\underline{0.04 \pm 0.02}$ & $0.02 \pm 0.02$ & $\mathbf{0.34 \pm 0.09}$ \\
&F1 Old Nodes\higherbetter & $0.00 \pm 0.01$ & $0.03 \pm 0.03$ & $OOM$ & $0.00 \pm 0.00$ & $\underline{0.07 \pm 0.04}$ & $0.04 \pm 0.03$ & $\mathbf{0.32 \pm 0.07}$ \\
&New Nodes Predicted\higherbetter & $8.39 \pm 3.56$ & $\underline{0.33 \pm 0.59}$ & $OOM$ & $0.42 \pm 0.54$ & $1.67 \pm 1.20$ & $6.73 \pm 3.08$ & $\mathbf{0.14 \pm 0.44}$ \\

\midrule
\multirow{11}{*}{\rotatebox{90}{\textbf{Structure Evaluation}}}&Avg Node Degree\tozero & $-0.27 \pm 0.09$ & $4.46 \pm 1.86$ & $OOM$ & $3.89 \pm 1.18$ & $2.38 \pm 0.94$ & $\underline{-0.12 \pm 0.11}$ & $\mathbf{0.02 \pm 0.10}$ \\
&Unique Degree Count\tozero & $-0.23 \pm 0.19$ & $1.57 \pm 0.93$ & $OOM$ & $1.79 \pm 0.75$ & $1.47 \pm 0.67$ & $\mathbf{-0.02 \pm 0.35}$ & $\underline{0.05 \pm 0.32}$ \\
&Degree Centrality\tozero & $-0.78 \pm 0.09$ & $11.56 \pm 7.42$ & $OOM$ & $9.02 \pm 5.27$ & $2.77 \pm 2.39$ & $\underline{-0.67 \pm 0.15}$ & $\mathbf{0.06 \pm 0.56}$ \\
&Assortativity Coefficient\tozero & $-2.69 \pm 0.89$ & $1.48 \pm 1.62$ & $OOM$ & $1.42 \pm 1.50$ & $\underline{1.39 \pm 1.95}$ & $-2.46 \pm 1.33$ & $\mathbf{1.04 \pm 0.99}$ \\
&Clustering Coefficient\tozero & $\underline{5.10 \pm 16.92}$ & $48.00 \pm 88.92$ & $OOM$ & $51.54 \pm 123.07$ & $48.06 \pm 121.80$ & $12.61 \pm 34.74$ & $\mathbf{3.06 \pm 7.58}$ \\
&Density\tozero & $-0.78 \pm 0.09$ & $11.56 \pm 7.42$ & $OOM$ & $9.02 \pm 5.27$ & $2.77 \pm 2.39$ & $\underline{-0.67 \pm 0.15}$ & $\mathbf{0.06 \pm 0.56}$ \\
&Num Triangles\tozero & $\underline{28.25 \pm 79.49}$ & $989.03 \pm 1802.47$ & $OOM$ & $953.29 \pm 1918.38$ & $797.07 \pm 1384.43$ & $59.66 \pm 82.10$ & $\mathbf{7.17 \pm 9.37}$ \\ 
&Descriptor Norm\tozero & $1033.25 \pm 583.15$ & $364.97 \pm 280.28$ & $OOM$ & $\underline{350.06 \pm 283.24}$ & $360.46 \pm 264.26$ & $761.41 \pm 597.24$ & $\mathbf{180.59 \pm 205.62}$ \\
&Median Extra Nodes\tozero & $297.00 \pm 19.20$ & $\mathbf{0.00 \pm 0.00}$ & $OOM$ & $\mathbf{0.00 \pm 0.00}$ & $\underline{7.00 \pm 3.74}$ & $222.00 \pm 18.75$ & $12.00 \pm 3.60$ \\
&Median Missing Nodes\tozero & $\mathbf{0.00 \pm 0.00}$ & $52.00 \pm 3.64$ & $OOM$ & ${50.00 \pm 5.57}$ & ${0.00 \pm 1.00}$ & $\mathbf{0.00 \pm 0.00}$ & $\underline{0.00 \pm 0.06}$ \\
&Median Extra Edges\tozero & $157.00 \pm 13.94$ & $\underline{148.00 \pm 15.52}$ & $OOM$ & $154.00 \pm 13.97$ & $232.00 \pm 14.32$ & $155.00 \pm 14.01$ & $\mathbf{13.00 \pm 3.87}$ \\
&Median Missing Edges\tozero & $\mathbf{0.00 \pm 0.00}$ & $\mathbf{0.00 \pm 0.00}$ & $OOM$ & $\mathbf{0.00 \pm 0.00}$ & $\mathbf{0.00 \pm 0.00}$ & $\mathbf{0.00 \pm 0.00}$ & $\underline{0.00 \pm 0.13}$ \\

\bottomrule
\bottomrule

\end{tabular}}

\end{table} 

\begin{table}[H] 
    \centering
    \caption{Comprehensive evaluation results—including edge, node, and structural metrics—for all benchmark models on \textbf{Iconomi}. \tozero indicates that values closer to 0 are better. \higherbetter denotes that higher values are better, while \lowerbetter denotes that lower values are better. \textbf{Bold} indicates best, the second-best is \underline{underlined}. $OOM$ indicate out-of-memory.}
    \label{tab:iconomi_results}
    \resizebox{1.00\linewidth}{!}{\begin{tabular}{l | c | c c c c c c | c}
\toprule
\toprule
& \textbf{Metric} & \textbf{HTGN} & \textbf{ROLAND} & \textbf{VGRNN} & \textbf{TGCN} & \textbf{GCLSTM} & \textbf{EvolveGCN} & \textbf{TopoGED} \\ \midrule
\multirow{11}{*}{\rotatebox{90}{\textbf{Edge Evaluation}}}&oo-bank Precision\higherbetter & $0.00 \pm 0.00$ & $0.00 \pm 0.00$ & $OOM$ & $0.00 \pm 0.00$ & $0.00 \pm 0.00$ & $0.00 \pm 0.00$ & $\mathbf{0.19 \pm 0.10}$ \\
&oo-bank Recall\higherbetter & $0.00 \pm 0.00$ & $0.00 \pm 0.00$ & $OOM$ & $0.00 \pm 0.00$ & $0.00 \pm 0.00$ & $0.00 \pm 0.00$ & $\mathbf{0.28 \pm 0.15}$ \\
&oo-bank F1\higherbetter & $0.00 \pm 0.00$ & $0.00 \pm 0.00$ & $OOM$ & $0.00 \pm 0.00$ & $0.00 \pm 0.00$ & $0.00 \pm 0.00$ & $\mathbf{0.22 \pm 0.09}$ \\
&oo-nobank Precision\higherbetter & $0.00 \pm 0.00$ & $0.00 \pm 0.00$ & $OOM$ & $0.00 \pm 0.00$ & $0.00 \pm 0.00$ & $0.00 \pm 0.00$ & $\mathbf{0.01 \pm 0.03}$ \\
&oo-nobank Recall\higherbetter & $0.00 \pm 0.00$ & $0.00 \pm 0.00$ & $OOM$ & $0.00 \pm 0.00$ & $0.00 \pm 0.00$ & $0.00 \pm 0.00$ & $\mathbf{0.01 \pm 0.02}$ \\
&oo-nobank F1\higherbetter & $0.00 \pm 0.00$ & $0.00 \pm 0.00$ & $OOM$ & $0.00 \pm 0.00$ & $0.00 \pm 0.00$ & $0.00 \pm 0.00$ & $\mathbf{0.01 \pm 0.02}$ \\
&Edge Precision\higherbetter & $0.00 \pm 0.00$ & $0.00 \pm 0.00$ & $OOM$ & $0.00 \pm 0.00$ & $0.00 \pm 0.00$ & $0.00 \pm 0.00$ & $\mathbf{0.10 \pm 0.06}$ \\
&Edge Recall\higherbetter & $0.00 \pm 0.00$ & $0.00 \pm 0.00$ & $OOM$ & $0.00 \pm 0.00$ & $0.00 \pm 0.00$ & $0.00 \pm 0.00$ & $\mathbf{0.11 \pm 0.06}$ \\
&Edge F1\higherbetter & $0.00 \pm 0.00$ & $0.00 \pm 0.00$ & $OOM$ & $0.00 \pm 0.00$ & $0.00 \pm 0.00$ & $0.00 \pm 0.00$ & $\mathbf{0.10 \pm 0.06}$ \\
&Num o-n Predicted\tozero & $-0.58 \pm 0.41$ & $-0.79 \pm 0.32$ & $OOM$ & $\mathbf{0.02 \pm 0.63}$ & $\underline{-0.53 \pm 0.54}$ & $-0.75 \pm 0.36$ & $0.18 \pm 0.54$ \\
&Num n-n Predicted\tozero & $157.28 \pm 145.81$ & $160.25 \pm 149.02$ & $OOM$ & $\underline{133.04 \pm 134.74}$ & $303.04 \pm 233.97$ & $175.42 \pm 285.77$ & $\mathbf{0.38 \pm 1.37}$ \\

\midrule
\multirow{7}{*}{\rotatebox{90}{\textbf{Node Evaluation}}}&Precision Nodes\higherbetter & $0.00 \pm 0.00$ & $\underline{0.01 \pm 0.01}$ & $OOM$ & $0.00 \pm 0.00$ & $\underline{0.01 \pm 0.00}$ & $0.00 \pm 0.00$ & $\mathbf{0.52 \pm 0.11}$ \\
&Recall Nodes\higherbetter & $0.00 \pm 0.00$ & $0.01 \pm 0.01$ & $OOM$ & $0.00 \pm 0.00$ & $\underline{0.03 \pm 0.01}$ & $0.01 \pm 0.01$ & $\mathbf{0.54 \pm 0.11}$ \\
&F1 Nodes\higherbetter & $0.00 \pm 0.00$ & $\underline{0.01 \pm 0.01}$ & $OOM$ & $0.00 \pm 0.00$ & ${0.01 \pm 0.00}$ & $0.00 \pm 0.01$ & $\mathbf{0.51 \pm 0.09}$ \\
&Precision Old Nodes\higherbetter & $0.00 \pm 0.02$ & $\mathbf{0.57 \pm 0.45}$ & $OOM$ & $0.00 \pm 0.00$ & $0.26 \pm 0.07$ & $0.09 \pm 0.16$ & $\underline{0.28 \pm 0.07}$ \\
&Recall Old Nodes\higherbetter & $0.00 \pm 0.00$ & $0.02 \pm 0.01$ & $OOM$ & $0.00 \pm 0.00$ & $\underline{0.06 \pm 0.02}$ & $0.01 \pm 0.02$ & $\mathbf{0.30 \pm 0.08}$ \\
&F1 Old Nodes\higherbetter & $0.00 \pm 0.01$ & $0.03 \pm 0.02$ & $OOM$ & $0.00 \pm 0.00$ & $\underline{0.09 \pm 0.03}$ & $0.02 \pm 0.04$ & $\mathbf{0.28 \pm 0.06}$ \\
&New Nodes Predicted\higherbetter & $10.24 \pm 7.16$ & $0.79 \pm 1.01$ & $OOM$ & $\underline{0.49 \pm 0.71}$ & $7.82 \pm 3.90$ & $6.63 \pm 6.09$ & $\mathbf{0.15 \pm 0.50}$ \\

\midrule
\multirow{11}{*}{\rotatebox{90}{\textbf{Structure Evaluation}}}&Avg Node Degree\tozero & $-0.37 \pm 0.06$ & $2.82 \pm 0.72$ & $OOM$ & $3.41 \pm 0.95$ & $0.59 \pm 1.16$ & $\underline{-0.04 \pm 0.15}$ & $\mathbf{0.02 \pm 0.09}$ \\
&Unique Degree Count\tozero & $-0.53 \pm 0.13$ & $1.65 \pm 0.84$ & $OOM$ & $1.89 \pm 0.82$ & $\underline{-0.24 \pm 0.94}$ & $\mathbf{0.07 \pm 0.35}$ & $0.10 \pm 0.28$ \\
&Degree Centrality\tozero & $-0.83 \pm 0.14$ & $5.28 \pm 4.95$ & $OOM$ & $7.30 \pm 5.43$ & $\underline{-0.48 \pm 0.56}$ & $-0.61 \pm 0.26$ & $\mathbf{0.07 \pm 0.81}$ \\
&Assortativity Coefficient\tozero & $-2.26 \pm 0.49$ & $\mathbf{0.06 \pm 0.37}$ & $OOM$ & $0.92 \pm 0.55$ & $1.59 \pm 0.93$ & $-1.66 \pm 0.81$ & $\underline{0.61 \pm 0.52}$ \\
&Clustering Coefficient\tozero & $\mathbf{1.36 \pm 3.48}$ & $19.54 \pm 34.79$ & $OOM$ & $25.50 \pm 47.62$ & $10.55 \pm 29.54$ & $7.60 \pm 17.10$ & $\underline{2.74 \pm 7.26}$ \\
&Density\tozero & $-0.83 \pm 0.14$ & $5.28 \pm 4.95$ & $OOM$ & $7.30 \pm 5.43$ & $\underline{-0.48 \pm 0.56}$ & $-0.61 \pm 0.26$ & $\mathbf{0.07 \pm 0.81}$ \\
&Num Triangles\tozero & $\underline{5.91 \pm 18.95}$ & $448.38 \pm 863.70$ & $OOM$ & $488.80 \pm 1095.20$ & $281.67 \pm 1744.48$ & $44.18 \pm 110.32$ & $\mathbf{4.81 \pm 3.54}$ \\ 
&Descriptor Norm\tozero & $935.66 \pm 592.46$ & $222.01 \pm 227.53$ & $OOM$ & $\underline{217.45 \pm 219.85}$ & $658.49 \pm 202.90$ & $452.03 \pm 566.43$ & $\mathbf{122.31 \pm 195.08}$ \\
&Median Extra Nodes\tozero & $237.00 \pm 15.06$ & $\mathbf{0.00 \pm 0.00}$ & $OOM$ & $\mathbf{0.00 \pm 0.00}$ & $197.00 \pm 7.56$ & $129.00 \pm 10.25$ & $\underline{5.00 \pm 2.85}$ \\
&Median Missing Nodes\tozero & $\mathbf{0.00 \pm 0.00}$ & $23.00 \pm 3.02$ & $OOM$ & $32.00 \pm 2.32$ & $\mathbf{0.00 \pm 0.00}$ & $\mathbf{0.00 \pm 0.00}$ & $\underline{0.00 \pm 0.54}$ \\
&Median Extra Edges\tozero & $110.00 \pm 8.97$ & $\underline{96.00 \pm 6.98}$ & $OOM$ & $106.00 \pm 8.67$ & $267.00 \pm 15.10$ & $109.00 \pm 8.44$ & $\mathbf{5.00 \pm 2.72}$ \\
&Median Missing Edges\tozero & $\mathbf{0.00 \pm 0.00}$ & $\mathbf{0.00 \pm 0.00}$ & $OOM$ & $\mathbf{0.00 \pm 0.00}$ & $\mathbf{0.00 \pm 0.00}$ & $\mathbf{0.00 \pm 0.00}$ & $\underline{0.00 \pm 0.38}$ \\

\bottomrule
\bottomrule

\end{tabular}}

\end{table} 

\begin{table}[H] 
    \centering
    \caption{Comprehensive evaluation results—including edge, node, and structural metrics—for all benchmark models on \textbf{Reddit B}. \tozero indicates that values closer to 0 are better. \higherbetter denotes that higher values are better, while \lowerbetter denotes that lower values are better. \textbf{Bold} indicates best, the second-best is \underline{underlined}.}
    \label{tab:reddit_b_results}
    \resizebox{1.00\linewidth}{!}{\begin{tabular}{l | c | c c c c c c | c}
\toprule
\toprule
& \textbf{Metric} & \textbf{HTGN} & \textbf{ROLAND} & \textbf{VGRNN} & \textbf{TGCN} & \textbf{GCLSTM} & \textbf{EvolveGCN} & \textbf{TopoGED} \\ \midrule
\multirow{11}{*}{\rotatebox{90}{\textbf{Edge Evaluation}}}&oo-bank Precision\higherbetter & $0.00 \pm 0.00$ & $0.00 \pm 0.00$ & $0.00 \pm 0.00$ & $0.00 \pm 0.00$ & $0.00 \pm 0.00$ & $0.00 \pm 0.00$ & $\mathbf{0.05 \pm 0.03}$ \\
&oo-bank Recall\higherbetter & $0.00 \pm 0.00$ & $0.00 \pm 0.00$ & $0.00 \pm 0.00$ & $0.00 \pm 0.00$ & $0.00 \pm 0.00$ & $0.00 \pm 0.00$ & $\mathbf{0.10 \pm 0.08}$ \\
&oo-bank F1\higherbetter & $0.00 \pm 0.00$ & $0.00 \pm 0.00$ & $0.00 \pm 0.00$ & $0.00 \pm 0.00$ & $0.00 \pm 0.00$ & $0.00 \pm 0.00$ & $\mathbf{0.06 \pm 0.03}$ \\
&oo-nobank Precision\higherbetter & $0.00 \pm 0.00$ & $0.00 \pm 0.00$ & $0.00 \pm 0.00$ & $0.00 \pm 0.00$ & $0.00 \pm 0.00$ & $0.00 \pm 0.00$ & $\mathbf{0.00 \pm 0.01}$ \\
&oo-nobank Recall\higherbetter & $0.00 \pm 0.00$ & $0.00 \pm 0.00$ & $0.00 \pm 0.00$ & $0.00 \pm 0.00$ & $0.00 \pm 0.00$ & $0.00 \pm 0.00$ & $\mathbf{0.00 \pm 0.01}$ \\
&oo-nobank F1\higherbetter & $0.00 \pm 0.00$ & $0.00 \pm 0.00$ & $0.00 \pm 0.00$ & $0.00 \pm 0.00$ & $0.00 \pm 0.00$ & $0.00 \pm 0.00$ & $\mathbf{0.00 \pm 0.01}$ \\
&Edge Precision\higherbetter & $0.00 \pm 0.00$ & $0.00 \pm 0.00$ & $0.00 \pm 0.00$ & $0.00 \pm 0.00$ & $0.00 \pm 0.00$ & $0.00 \pm 0.00$ & $\mathbf{0.03 \pm 0.02}$ \\
&Edge Recall\higherbetter & $0.00 \pm 0.00$ & $0.00 \pm 0.00$ & $0.00 \pm 0.00$ & $0.00 \pm 0.00$ & $0.00 \pm 0.00$ & $0.00 \pm 0.00$ & $\mathbf{0.03 \pm 0.02}$ \\
&Edge F1\higherbetter & $0.00 \pm 0.00$ & $0.00 \pm 0.00$ & $0.00 \pm 0.00$ & $0.00 \pm 0.00$ & $0.00 \pm 0.00$ & $0.00 \pm 0.00$ & $\mathbf{0.03 \pm 0.02}$ \\
&Num o-n Predicted\tozero & $\underline{-1.00 \pm 0.00}$ & $\underline{-1.00 \pm 0.00}$ & $\underline{-1.00 \pm 0.00}$ & $\underline{-1.00 \pm 0.00}$ & $\underline{-1.00 \pm 0.00}$ & $\underline{-1.00 \pm 0.00}$ & $\mathbf{0.11 \pm 0.51}$ \\
&Num n-n Predicted\tozero & $467.53 \pm 284.41$ & $\underline{461.78 \pm 269.78}$ & $866.86 \pm 528.87$ & $466.79 \pm 283.91$ & $782.58 \pm 470.41$ & $466.82 \pm 283.93$ & $\mathbf{0.59 \pm 1.56}$ \\

\midrule
\multirow{7}{*}{\rotatebox{90}{\textbf{Node Evaluation}}}&Precision Nodes\higherbetter & $0.00 \pm 0.00$ & $0.00 \pm 0.00$ & $0.00 \pm 0.00$ & $0.00 \pm 0.00$ & $0.00 \pm 0.00$ & $0.00 \pm 0.00$ & $\mathbf{0.38 \pm 0.04}$ \\
&Recall Nodes\higherbetter & $0.00 \pm 0.00$ & $0.00 \pm 0.00$ & $0.00 \pm 0.00$ & $0.00 \pm 0.00$ & $0.00 \pm 0.00$ & $0.00 \pm 0.00$ & $\mathbf{0.38 \pm 0.03}$ \\
&F1 Nodes\higherbetter & $0.00 \pm 0.00$ & $0.00 \pm 0.00$ & $0.00 \pm 0.00$ & $0.00 \pm 0.00$ & $0.00 \pm 0.00$ & $0.00 \pm 0.00$ & $\mathbf{0.38 \pm 0.03}$ \\
&Precision Old Nodes\higherbetter & $0.00 \pm 0.00$ & $0.00 \pm 0.00$ & $0.00 \pm 0.00$ & $0.00 \pm 0.00$ & $0.00 \pm 0.00$ & $0.00 \pm 0.00$ & $\mathbf{0.34 \pm 0.04}$ \\
&Recall Old Nodes\higherbetter & $0.00 \pm 0.00$ & $0.00 \pm 0.00$ & $0.00 \pm 0.00$ & $0.00 \pm 0.00$ & $0.00 \pm 0.00$ & $0.00 \pm 0.00$ & $\mathbf{0.34 \pm 0.03}$ \\
&F1 Old Nodes\higherbetter & $0.00 \pm 0.00$ & $0.00 \pm 0.00$ & $0.00 \pm 0.00$ & $0.00 \pm 0.00$ & $0.00 \pm 0.00$ & $0.00 \pm 0.00$ & $\mathbf{0.34 \pm 0.03}$ \\
&New Nodes Predicted\higherbetter & $23.99 \pm 6.86$ & $4.88 \pm 2.47$ & $3.82 \pm 1.78$ & $6.87 \pm 5.81$ & $\underline{3.22 \pm 1.11}$ & $8.53 \pm 4.13$ & $\mathbf{0.04 \pm 0.31}$ \\

\midrule
\multirow{11}{*}{\rotatebox{90}{\textbf{Structure Evaluation}}}&Avg Node Degree\tozero & $\underline{0.41 \pm 0.16}$ & $5.55 \pm 2.10$ & $13.17 \pm 2.96$ & $5.09 \pm 2.65$ & $13.02 \pm 1.71$ & $3.01 \pm 1.05$ & $\mathbf{0.02 \pm 0.08}$ \\
&Unique Degree Count\tozero & $\underline{1.65 \pm 0.55}$ & $2.71 \pm 0.86$ & $2.13 \pm 0.81$ & $2.31 \pm 1.14$ & $4.63 \pm 0.98$ & $2.78 \pm 0.90$ & $\mathbf{0.87 \pm 0.39}$ \\
&Degree Centrality\tozero & $\underline{-0.22 \pm 0.15}$ & $17.19 \pm 9.94$ & $43.45 \pm 16.60$ & $16.15 \pm 10.87$ & $45.20 \pm 7.95$ & $5.70 \pm 3.12$ & $\mathbf{0.03 \pm 0.11}$ \\
&Assortativity Coefficient\tozero & $-3.63 \pm 28.88$ & $6.25 \pm 65.25$ & $10.71 \pm 77.63$ & $8.28 \pm 67.75$ & $5.16 \pm 60.15$ & $\mathbf{1.00 \pm 21.05}$ & $\underline{1.10 \pm 32.12}$ \\
&Clustering Coefficient\tozero & $\underline{3.78 \pm 13.51}$ & $44.47 \pm 110.98$ & $58.99 \pm 156.80$ & $42.99 \pm 161.65$ & $46.75 \pm 133.31$ & $18.76 \pm 52.63$ & $\mathbf{2.07 \pm 6.25}$ \\
&Density\tozero & $\underline{-0.22 \pm 0.15}$ & $17.19 \pm 9.94$ & $43.45 \pm 16.60$ & $16.15 \pm 10.87$ & $45.20 \pm 7.95$ & $5.70 \pm 3.12$ & $\mathbf{0.03 \pm 0.11}$ \\
&Num Triangles\tozero & $\underline{151.71 \pm 211.40}$ & $843.41 \pm 1136.71$ & $1707.68 \pm 2023.11$ & $687.13 \pm 934.66$ & $1970.28 \pm 2612.84$ & $669.92 \pm 789.01$ & $\mathbf{23.25 \pm 38.07}$ \\ 
&Descriptor Norm\tozero & $1046.93 \pm 276.60$ & $1002.78 \pm 148.13$ & $1495.89 \pm 206.22$ & $1030.84 \pm 319.63$ & $1421.72 \pm 174.75$ & $\underline{755.22 \pm 126.19}$ & $\mathbf{446.58 \pm 67.84}$ \\
&Median Extra Nodes\tozero & $352.00 \pm 16.18$ & $\underline{4.00 \pm 0.00}$ & $\mathbf{0.00 \pm 0.00}$ & $269.00 \pm 0.00$ & $\mathbf{0.00 \pm 0.00}$ & $71.00 \pm 0.00$ & $23.00 \pm 4.24$ \\
&Median Missing Nodes\tozero & $\mathbf{0.00 \pm 0.00}$ & $253.00 \pm 10.95$ & $291.00 \pm 6.25$ & $265.00 \pm 9.92$ & $299.00 \pm 8.01$ & ${161.00 \pm 19.08}$ & $\underline{0.00 \pm 2.24}$ \\
&Median Extra Edges\tozero & $463.00 \pm 20.23$ & $\underline{460.00 \pm 20.39}$ & $1101.00 \pm 21.94$ & $462.00 \pm 21.05$ & $967.00 \pm 17.34$ & $462.00 \pm 20.31$ & $\mathbf{13.00 \pm 7.77}$ \\
&Median Missing Edges\tozero & $\mathbf{0.00 \pm 0.00}$ & $\mathbf{0.00 \pm 0.00}$ & $\mathbf{0.00 \pm 0.00}$ & $\mathbf{0.00 \pm 0.00}$ & $\mathbf{0.00 \pm 0.00}$ & $\mathbf{0.00 \pm 0.00}$ & $\underline{0.00 \pm 1.38}$ \\

\bottomrule
\bottomrule

\end{tabular}}

\end{table} 

\begin{table}[H] 
    \centering
    \caption{Comprehensive evaluation results—including edge, node, and structural metrics—for all benchmark models on \textbf{TGBL-Wiki}. \tozero indicates that values closer to 0 are better. \higherbetter denotes that higher values are better, while \lowerbetter denotes that lower values are better. \textbf{Bold} indicates best, the second-best is \underline{underlined}.}
    \label{tab:tgbl_wiki_results}
    \resizebox{1.00\linewidth}{!}{\begin{tabular}{l | c | c c c c c c | c}
\toprule
\toprule
& \textbf{Metric} & \textbf{HTGN} & \textbf{ROLAND} & \textbf{VGRNN} & \textbf{TGCN} & \textbf{GCLSTM} & \textbf{EvolveGCN} & \textbf{TopoGED} \\ \midrule
\multirow{11}{*}{\rotatebox{90}{\textbf{Edge Evaluation}}}&oo-bank Precision\higherbetter & $\underline{0.39 \pm 0.23}$ & $0.24 \pm 0.15$ & $0.00 \pm 0.00$ & $\mathbf{0.65 \pm 0.40}$ & $0.00 \pm 0.00$ & $0.00 \pm 0.00$ & $0.27 \pm 0.16$ \\
&oo-bank Recall\higherbetter & $0.03 \pm 0.02$ & $0.01 \pm 0.01$ & $0.00 \pm 0.00$ & $0.01 \pm 0.01$ & $0.00 \pm 0.00$ & $0.00 \pm 0.00$ & $\mathbf{0.34 \pm 0.20}$ \\
&oo-bank F1\higherbetter & $\underline{0.05 \pm 0.03}$ & $0.02 \pm 0.01$ & $0.00 \pm 0.00$ & $0.02 \pm 0.01$ & $0.00 \pm 0.00$ & $0.00 \pm 0.00$ & $\mathbf{0.30 \pm 0.18}$ \\
&oo-nobank Precision\higherbetter & $0.00 \pm 0.00$ & $0.00 \pm 0.00$ & $0.00 \pm 0.00$ & $0.00 \pm 0.00$ & $0.00 \pm 0.00$ & $0.00 \pm 0.00$ & $\mathbf{0.07 \pm 0.07}$ \\
&oo-nobank Recall\higherbetter & $\underline{0.01 \pm 0.01}$ & $\underline{0.01 \pm 0.00}$ & $0.00 \pm 0.00$ & $0.00 \pm 0.00$ & $0.00 \pm 0.00$ & $0.00 \pm 0.00$ & $\mathbf{0.07 \pm 0.08}$ \\
&oo-nobank F1\higherbetter & $0.00 \pm 0.00$ & $0.00 \pm 0.00$ & $0.00 \pm 0.00$ & $0.00 \pm 0.00$ & $0.00 \pm 0.00$ & $0.00 \pm 0.00$ & $\mathbf{0.07 \pm 0.08}$ \\
&Edge Precision\higherbetter & $\underline{0.01 \pm 0.00}$ & $0.00 \pm 0.00$ & $0.00 \pm 0.00$ & $0.00 \pm 0.00$ & $0.00 \pm 0.00$ & $0.00 \pm 0.00$ & $\mathbf{0.17 \pm 0.01}$ \\
&Edge Recall\higherbetter & $\underline{0.01 \pm 0.00}$ & $\underline{0.01 \pm 0.00}$ & $0.00 \pm 0.00$ & $\underline{0.01 \pm 0.00}$ & $0.00 \pm 0.00$ & $0.00 \pm 0.00$ & $\mathbf{0.18 \pm 0.01}$ \\
&Edge F1\higherbetter & $\underline{0.01 \pm 0.00}$ & $0.00 \pm 0.00$ & $0.00 \pm 0.00$ & $0.00 \pm 0.00$ & $0.00 \pm 0.00$ & $0.00 \pm 0.00$ & $\mathbf{0.18 \pm 0.01}$ \\
&Num o-n Predicted\tozero & $\underline{-0.89 \pm 0.03}$ & $-1.00 \pm 0.00$ & $-1.00 \pm 0.00$ & $-0.94 \pm 0.10$ & $0.90 \pm 0.24$ & $-1.00 \pm 0.00$ & $\mathbf{0.14 \pm 0.18}$ \\
&Num n-n Predicted\tozero & $\mathbf{0.05 \pm 0.94}$ & $\underline{-0.60 \pm 0.55}$ & $\underline{-0.60 \pm 0.55}$ & $\underline{-0.60 \pm 0.55}$ & $14.76 \pm 12.79$ & $\underline{-0.60 \pm 0.55}$ & $3.09 \pm 5.06$ \\

\midrule
\multirow{7}{*}{\rotatebox{90}{\textbf{Node Evaluation}}}&Precision Nodes\higherbetter & $\underline{0.57 \pm 0.03}$ & $0.33 \pm 0.05$ & $0.11 \pm 0.01$ & $0.34 \pm 0.03$ & $0.53 \pm 0.09$ & $0.20 \pm 0.03$ & $\mathbf{0.63 \pm 0.03}$ \\
&Recall Nodes\higherbetter & $0.25 \pm 0.01$ & $0.18 \pm 0.01$ & $\underline{0.48 \pm 0.03}$ & $0.13 \pm 0.08$ & $0.08 \pm 0.01$ & $0.18 \pm 0.08$ & $\mathbf{0.67 \pm 0.02}$ \\
&F1 Nodes\higherbetter & $\underline{0.35 \pm 0.01}$ & $0.24 \pm 0.02$ & $0.17 \pm 0.01$ & $0.17 \pm 0.07$ & $0.13 \pm 0.02$ & $0.17 \pm 0.05$ & $\mathbf{0.65 \pm 0.01}$ \\
&Precision Old Nodes\higherbetter & $\underline{0.57 \pm 0.03}$ & $0.33 \pm 0.05$ & $0.11 \pm 0.01$ & $0.34 \pm 0.02$ & $\underline{0.57 \pm 0.06}$ & $0.20 \pm 0.03$ & $\mathbf{0.60 \pm 0.02}$ \\
&Recall Old Nodes\higherbetter & $0.28 \pm 0.01$ & $0.21 \pm 0.02$ & $\underline{0.55 \pm 0.04}$ & $0.15 \pm 0.09$ & $0.08 \pm 0.01$ & $0.20 \pm 0.10$ & $\mathbf{0.63 \pm 0.02}$ \\
&F1 Old Nodes\higherbetter & $\underline{0.38 \pm 0.01}$ & $0.25 \pm 0.02$ & $0.18 \pm 0.01$ & $0.19 \pm 0.08$ & $0.14 \pm 0.01$ & $0.19 \pm 0.05$ & $\mathbf{0.62 \pm 0.01}$ \\
&New Nodes Predicted\higherbetter & $-0.98 \pm 0.00$ & $-1.00 \pm 0.00$ & $-1.00 \pm 0.00$ & $-0.99 \pm 0.01$ & $\underline{-0.86 \pm 0.02}$ & $-1.00 \pm 0.00$ & $\mathbf{0.14 \pm 0.18}$ \\

\midrule
\multirow{11}{*}{\rotatebox{90}{\textbf{Structure Evaluation}}}&Avg Node Degree\tozero & $5.13 \pm 0.40$ & $4.02 \pm 0.63$ & $\underline{0.14 \pm 0.02}$ & $8.12 \pm 4.85$ & $17.51 \pm 1.16$ & $2.59 \pm 2.12$ & $\mathbf{0.01 \pm 0.02}$ \\
&Unique Degree Count\tozero & $1.42 \pm 0.20$ & $2.24 \pm 0.32$ & $\underline{-0.90 \pm 0.01}$ & $2.35 \pm 0.40$ & $2.92 \pm 0.23$ & $1.51 \pm 0.55$ & $\mathbf{0.26 \pm 0.19}$ \\
&Degree Centrality\tozero & $13.11 \pm 1.64$ & $8.38 \pm 2.43$ & $\underline{-0.75 \pm 0.02}$ & $38.96 \pm 33.94$ & $130.76 \pm 13.01$ & $5.27 \pm 7.07$ & $\mathbf{-0.05 \pm 0.06}$ \\
&Assortativity Coefficient\tozero & $-6.90 \pm 1.44$ & $\underline{3.51 \pm 1.32}$ & $8.97 \pm 2.27$ & $3.91 \pm 0.48$ & $4.27 \pm 1.21$ & $3.80 \pm 1.48$ & $\mathbf{1.58 \pm 0.35}$ \\
&Clustering Coefficient\tozero & $0.36 \pm 0.01$ & $\mathbf{0.00 \pm 0.00}$ & $\mathbf{0.00 \pm 0.00}$ & $0.56 \pm 0.20$ & $0.79 \pm 0.01$ & $0.41 \pm 0.28$ & $\underline{0.02 \pm 0.01}$ \\
&Density\tozero & $13.11 \pm 1.64$ & $8.38 \pm 2.43$ & $\underline{-0.75 \pm 0.02}$ & $38.96 \pm 33.94$ & $130.76 \pm 13.01$ & $5.27 \pm 7.07$ & $\mathbf{-0.05 \pm 0.06}$ \\
&Num Triangles\tozero & $11045.00 \pm 1653.77$ & $\mathbf{0.00 \pm 0.00}$ & $\mathbf{0.00 \pm 0.00}$ & $30045.00 \pm 13822.67$ & $40188.20 \pm 6563.66$ & $11052.40 \pm 4986.92$ & $\underline{99.60 \pm 39.51}$ \\ 
&Descriptor Norm\tozero & $3447.24 \pm 307.45$ & $2997.41 \pm 242.59$ & $17861.99 \pm 1923.61$ & $3205.85 \pm 759.27$ & $3762.26 \pm 296.79$ & $\underline{2652.77 \pm 461.59}$ & $\mathbf{1034.67 \pm 135.03}$ \\
&Median Extra Nodes\tozero & $\mathbf{0.00 \pm 0.00}$ & $\mathbf{0.00 \pm 0.00}$ & $5104.00 \pm 414.76$ & $\mathbf{0.00 \pm 0.00}$ & $\mathbf{0.00 \pm 0.00}$ & $\underline{0.00 \pm 203.32}$ & $99.00 \pm 51.77$ \\
&Median Missing Nodes\tozero & $753.00 \pm 72.77$ & $646.00 \pm 77.90$ & $\mathbf{0.00 \pm 0.00}$ & $1008.00 \pm 294.69$ & $1154.00 \pm 66.14$ & $77.00 \pm 257.93$ & $\underline{0.00 \pm 4.34}$ \\
&Median Extra Edges\tozero & $2123.00 \pm 176.71$ & $2257.00 \pm 117.44$ & $5292.00 \pm 418.19$ & $\underline{2070.00 \pm 185.88}$ & $\underline{2070.00 \pm 188.06}$ & $\underline{2070.00 \pm 185.93}$ & $\mathbf{87.00 \pm 60.57}$ \\
&Median Missing Edges\tozero & $\mathbf{0.00 \pm 0.00}$ & $\mathbf{0.00 \pm 0.00}$ & $\mathbf{0.00 \pm 0.00}$ & $\mathbf{0.00 \pm 0.00}$ & $\mathbf{0.00 \pm 0.00}$ & $\mathbf{0.00 \pm 0.00}$ & $\underline{0.00 \pm 1.68}$ \\
\bottomrule
\bottomrule

\end{tabular}}

\end{table} 

\clearpage

\end{document}